\documentclass[letterpaper]{article} 
\usepackage{aaai2026}  
\usepackage{times}  
\usepackage{helvet}  
\usepackage{courier}  
\usepackage[hyphens]{url}  
\usepackage{graphicx} 
\usepackage{natbib}  
\usepackage{caption} 
\usepackage{algpseudocode}

\usepackage[ruled,vlined]{algorithm2e}
\usepackage{amssymb}
\usepackage{pifont}
\usepackage{multirow}
\usepackage[table]{xcolor}
\usepackage{subcaption}

\usepackage{booktabs} 
\usepackage{tabularx}
\usepackage{amsmath} 
\usepackage{array} 
\usepackage{newfloat}
\usepackage{listings}
\DeclareCaptionStyle{ruled}{labelfont=normalfont,labelsep=colon,strut=off} 
\title{
Cross-Sample Augmented Test-Time Adaptation for Personalized\\ 
Intraoperative Hypotension Prediction
}
\author {
    Kanxue Li\textsuperscript{\rm 1},
    Yibing Zhan\textsuperscript{\rm 1\thanks{Corresponding authors: Yibing Zhan and Hua Jin.}},
    Hua Jin\textsuperscript{\rm 2*},
    Chongchong Qi\textsuperscript{\rm 3},
    Xu Lin\textsuperscript{\rm 3},
    Baosheng Yu\textsuperscript{\rm 4}
}
\affiliations {
    \textsuperscript{\rm 1}School of Computer Science, Wuhan University\\
    \textsuperscript{\rm 2}First People's Hospital of Yunnan Province\\
    \textsuperscript{\rm 3}Yunnan United Vision Technology Company Limited\\
    \textsuperscript{\rm 4}Nanyang Technological University\\
    
    likanxue@whu.edu.cn
}
\usepackage{bibentry}

\begin{document}

\maketitle

\begin{abstract}
Intraoperative hypotension (IOH) poses significant surgical risks, but accurate prediction remains challenging due to patient-specific variability. While test-time adaptation (TTA) offers a promising approach for personalized prediction, the rarity of IOH events often leads to unreliable test-time training. To address this, we propose CSA-TTA, a novel \textbf{C}ross-\textbf{S}ample \textbf{A}ugmented \textbf{T}est-\textbf{T}ime \textbf{A}daptation framework that enhances training by incorporating hypotension events from other individuals. Specifically, we first construct a cross-sample bank by segmenting historical data into hypotensive and non-hypotensive samples. Then, we introduce a coarse-to-fine retrieval strategy for building test-time training data: we initially apply K-Shape clustering to identify representative cluster centers and subsequently retrieve the top-K semantically similar samples based on the current patient signal. Additionally, we integrate both self-supervised masked reconstruction and retrospective sequence forecasting signals during training to enhance model adaptability to rapid and subtle intraoperative dynamics. We evaluate the proposed CSA-TTA on both the VitalDB dataset and a real-world in-hospital dataset by integrating it with state-of-the-art time series forecasting models, including TimesFM and UniTS. CSA-TTA consistently enhances performance across settings—for instance, on VitalDB, it improves Recall and F1 scores by +1.33\% and +1.13\%, respectively, under fine-tuning, and by +7.46\% and +5.07\% in zero-shot scenarios—demonstrating strong robustness and generalization.
\end{abstract}

\begin{links}
    \link{Code}{https://github.com/kanxueli/CSA-TTA}
\end{links}

\section{Introduction}

\begin{figure}[!t]
    \centering
    \includegraphics[height=0.82\linewidth, width=0.99\linewidth]
    {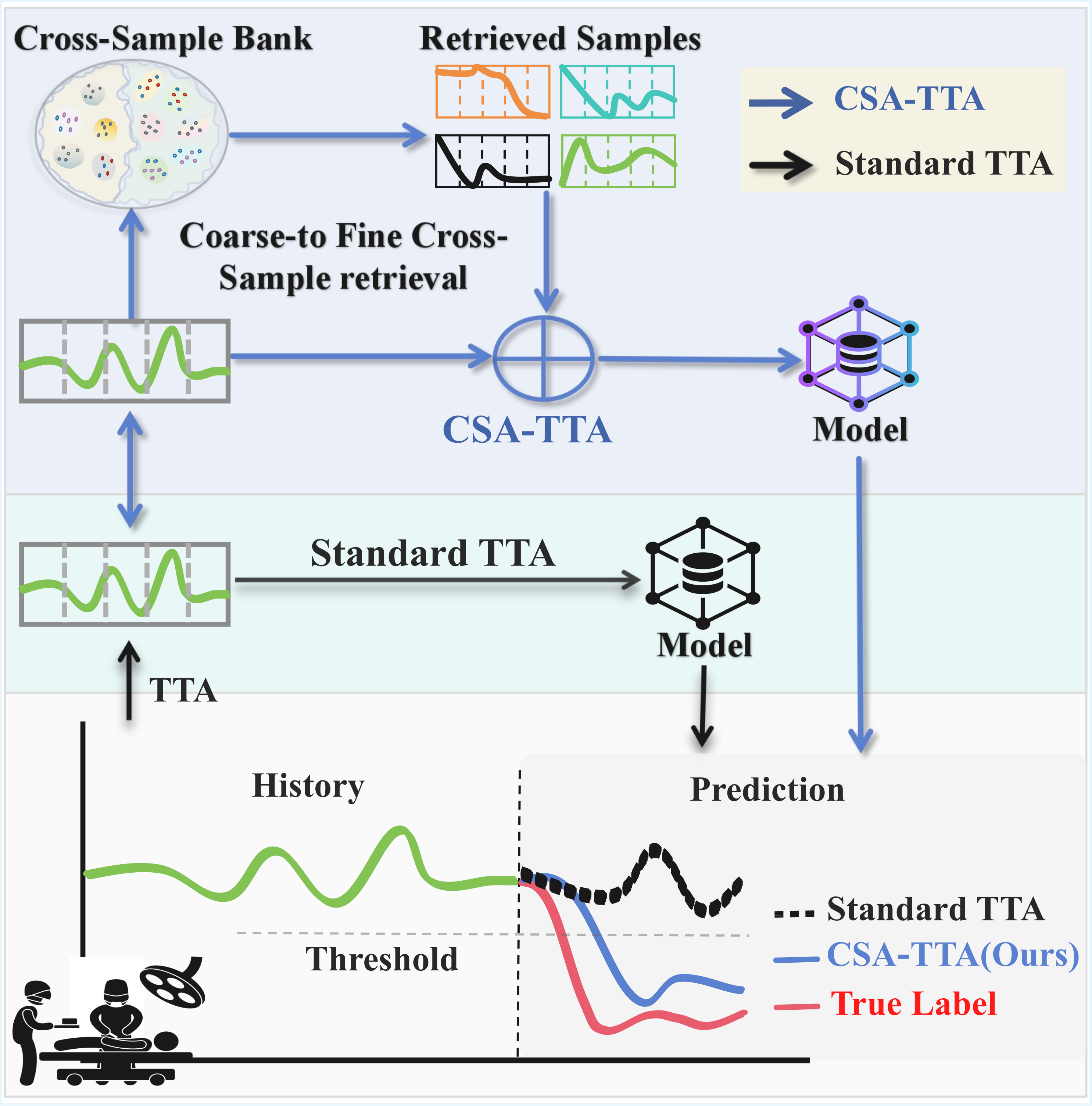}
    \caption{An illustrative comparison. Standard TTA, relying on recent stable history, often produces overly smooth predictions and misses sudden changes. CSA-TTA leverages a cross-sample augmented dataset to capture diverse temporal patterns, enabling personalized IOH prediction.
    }
    \label{fig:compare_tta}
\end{figure}

\begin{figure*}[ht!]
    \centering
    \includegraphics[width=\linewidth]{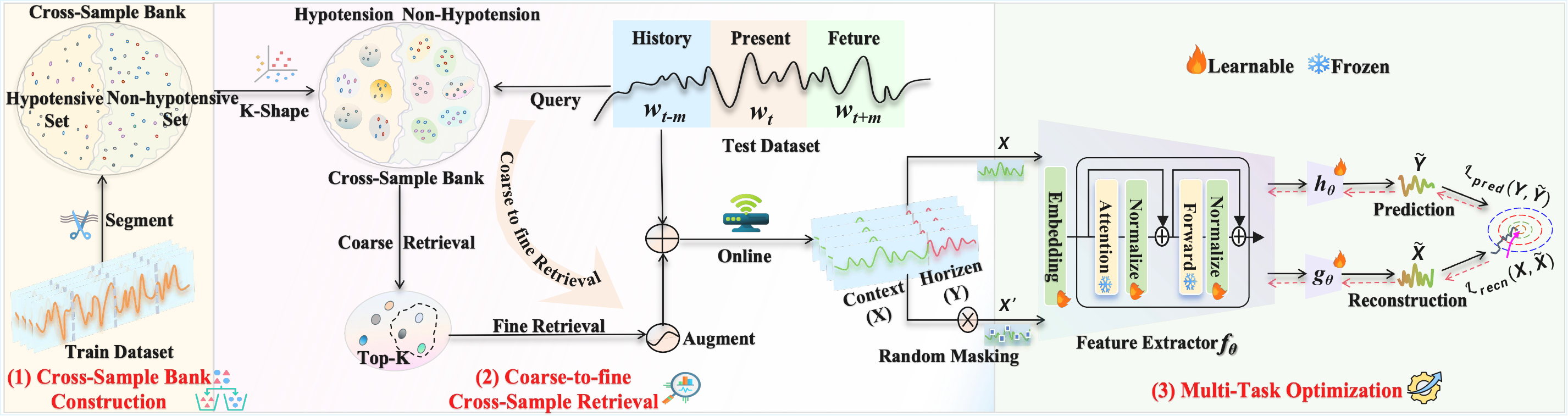}
    \caption{The main proposed CSA-TTA framework. It comprises three key steps: (1) Cross-sample bank construction, (2) Coarse-to-fine retrieval, and (3) Multi-task optimization.}
    \label{fig:overview_pipeline}
\end{figure*}

Intraoperative hypotension (IOH) — typically defined as blood pressure falling below a critical threshold for a sustained period~\cite{ioh_review1, ioh_define2}—is a common but serious complication during surgery. It is strongly associated with adverse outcomes such as acute kidney injury, myocardial infarction, stroke, and even mortality~\cite{ioh_class11, ioh_method111}. Accurate and timely prediction of IOH is critical for enabling early interventions before blood pressure drops to dangerous levels~\cite{ioh_class121,ioh_class12}, thereby reducing both the incidence and severity of these adverse events~\cite{ioh_method1}. However, due to the complex, dynamic, and highly patient-specific nature of physiological responses during surgery, reliable IOH prediction remains a significant challenge despite advances in intraoperative monitoring and machine learning.

Recent studies have explored various methods to improve IOH prediction~\cite{ioh_class21, ioh_review333, vitaldb}. For instance, CMA~\cite{cma} employed attention mechanisms to capture temporal and feature-level dependencies, while HMF~\cite{hmf} integrated contextual, physiological, and temporal features. However, the ability of these models to generalize remains limited by individual differences in patients' physiology and the influence of clinical interventions (e.g., anesthesia or drug administration)~\cite{ioh_review2,ioh_review444, ioh_review}. These factors introduce implicit distribution shifts in real-time signals that are typically poorly captured by population-level models.

TTA offers a promising paradigm to tackle such distribution shifts by adapting models using test data during inference~\cite{ttt_survey1}. Techniques such as TTT~\cite{ttt} and TTT++~\cite{ttt++} leverage self-supervised auxiliary tasks to refine models at inference time and have shown effectiveness in fields like computer vision~\cite{tta_cv,ttn} and NLP~\cite{tta_nlp}. In the context of IOH, TTA holds potential for personalizing predictions by utilizing recent patient history to dynamically adjust to individual physiological patterns. Despite its promise, standard TTA struggles due to the rarity of hypotensive events. For example, in the VitalDB dataset~\cite{vitaldb}, hypotension accounts for just 12.6\% of all samples, with the majority of patients experiencing it in less than 10\% of the surgical timeline. Standard TTA, which typically relies on single-sample adaptation, often fails to detect sudden blood pressure drops, especially when such transitions are not present in the current patient’s history. As shown in Figure~\ref{fig:compare_tta}, these methods yield overly smooth predictions and underperform during abrupt physiological changes caused by clinical interventions.

To overcome these limitations, we propose cross-sample augmented test-time adaptation (CSA-TTA) for personalized IOH prediction. As depicted in Figure~\ref{fig:overview_pipeline}, CSA-TTA first builds a cross-sample bank by partitioning historical data into hypotensive and non-hypotensive segments, increasing the availability and diversity of critical training signals. To retrieve relevant samples efficiently, we design a coarse-to-fine strategy: K-Shape clustering~\cite{kshape} is used to identify representative cluster centers in the coarse stage, followed by fine-grained retrieval of the top-K semantically similar samples based on the patient’s current data. These retrieved samples—optionally augmented with perturbations—are combined with the patient’s signal to form a balanced and representative adaptation set. Finally, CSA-TTA employs a multi-task optimization strategy, incorporating both self-supervised masked reconstruction and retrospective forecasting to improve model adaptability to subtle and rapid intraoperative dynamics.

Our main contributions are twofold:
\textbf{1)} To the best of our knowledge, this is the first attempt to apply test-time adaptation for personalized IOH prediction.
\textbf{2)} We propose CSA-TTA, which introduces a cross-sample bank, a coarse-to-fine retrieval strategy, and multi-task optimization. Together, these components enable robust test-time adaptation for personalized IOH prediction, despite the extreme scarcity of individual IOH events during surgery.

\section{Related Work}

\paragraph{Intraoperative Hypotension Prediction}
IOH prediction is critical for timely intervention and reducing postoperative risks~\cite{ioh_review1,ioh_2,ioh_review3}.
With the growing availability of continuous physiological data, many deep learning models have been developed, following~\cite{ioh_method1,ioh_review444}. The first treats IOH prediction as a classification task based on time series inputs, but suffers from limited interpretability and sensitivity to varying clinical definitions~\cite{ioh_class11,ioh_class12}. The second forecasts continuous blood pressure trajectories, followed by post hoc IOH detection, offering improved flexibility but still struggling with patient-specific variability and intraoperative distribution shifts—factors that undermine robustness in real-world deployment~\cite{hmf}. 

\paragraph{Test-time Adaptation}
TTA is a paradigm for mitigating distribution shifts between training (source) and test (target) data, enabling models to adjust to new data distributions during inference~\cite{ttt_survey1}. The foundational approach to TTA is TTT~\cite{ttt}, which incorporates self-supervised auxiliary tasks during inference to update model parameters and enhance generalization under distribution shift. Building on this, TTT++~\cite{ttt++} systematically analyzes the scenarios where self-supervised test-time training is effective or limited, and proposes strategies to improve its stability and robustness. These methods have shown success in various fields, such as vision~\cite{tta_cv} and language~\cite{ttt_survey2,tta_nlp}. Despite its potential, standard test-time adaptation struggles in IOH prediction due to rapidly shifting patient-specific physiological signals and rare 
IOH events. 

\section{Method}

In this section, we detail CSA-TTA for personalized IOH prediction. As illustrated in Figure~\ref{fig:overview_pipeline}, CSA-TTA comprises three key components: cross-sample bank construction, coarse-to-fine retrieval, and multi-task optimization. 

\subsection{Problem Formulation} 
Let $X \in \mathbb{R}^{L \times C}$ represent a multivariate time series of a patient’s vital signs, where $L$ is the lookback window length and $C$ is the number of variables (e.g., mean arterial pressure, body temperature, heart rate, ECG). The objective of IOH prediction is to forecast the blood pressure sequence $Y \in \mathbb{R}^{1 \times H}$ over the next H time steps using a model $F_\theta$: $\hat{Y} = F_\theta(X)$. Traditional models aim to learn the conditional distribution $P(Y|X)$, assuming that the observed features $X$ fully capture the relevant physiological dynamics. However, unmeasured patient-specific factors and subtle signal shifts—denoted as $X_{\text{latent}}$—can significantly affect blood pressure. The true target distribution is thus better represented as $P(Y|X, X_{\text{latent}})$.

Since $X_{\text{latent}}$ is not observable at inference, models trained only on X often suffer from distributional mismatch, leading to poor generalization—especially in the presence of individualized IOH patterns. Test-time adaptation (TTA) addresses this by leveraging a patient’s recent history $D_{\text{hist}}$ to infer latent physiological context, implicitly aligning predictions with:
\begin{equation}
  P(Y | X) \cdot P(X, X_{\text{latent}}| D_{\text{hist}}) \approx P(Y | X, X_{\text{latent}}).
\end{equation}
Here, $P(X, X_{\text{latent}} \mid D_{\text{hist}})$ serves as an empirical posterior over latent dynamics, estimated from the patient's trajectory. However, due to the episodic and sparse nature of hypotension, short-term patient data often lacks sufficient hypotensive information. To address this, we posit that patients with similar physiological profiles exhibit similar intraoperative responses. Thus, we extend the adaptation context using cross-sample retrieval:
\begin{equation}
P(X, X_{\mathrm{latent}} | D_{\mathrm{hist}}) \approx \hat{P}(X, X_{\mathrm{latent}} | {R}\!\left(D_{\text{hist}}\right)),
\end{equation}
where $\mathcal{R}\!\left(D_{\text{hist}}\right)$ is a set of retrieved cross-sample segments that are temporally and semantically aligned with the current patient trajectory.  This strategy enriches the adaptation signal with personalized priors from similar patients, enabling more robust estimation of $P(Y | X, X_{\text{latent}})$, even in the absence of local hypotensive events.

\begin{figure}[!t]
    \centering
    \includegraphics[width=\linewidth]
    {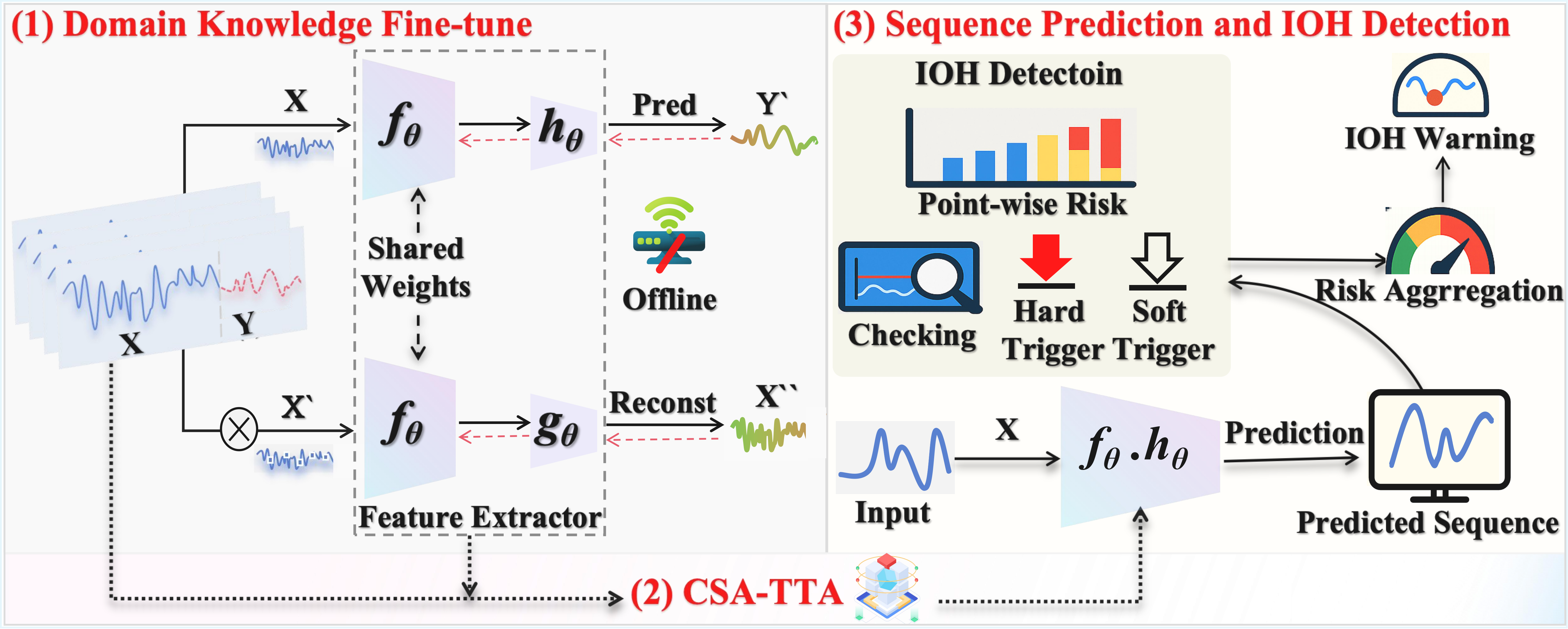}
    \caption{Illustration of the key steps in applying CSA-TTA for personalized IOH prediction.  (1) Domain knowledge adaptation, (2) CSA-TTA, and (3) Sequence prediction and IOH detection.}
    \label{fig:ioh_prediction}
\end{figure}

\subsection{Cross-Sample Bank Construction}

CSA-TTA introduces a cross-sample bank to enrich the diversity of temporal patterns. Specifically, it first segments physiological time series data from all patients in the historical dataset into fixed-length fragments, forming a cross-sample bank \(B\).
The original intraoperative hypotension dataset is denoted as:
\begin{equation}
    \mathcal{D} = \{(X_n, Y_n)\}_{n=1}^N,
\end{equation}
where $N$ is the total number of samples, $X_n$ represents the $n$-th time series sample, and $Y_n$ denotes the corresponding label. Each sample derived from $D$ is represented as $s=(X_n, Y_n)$, where $x$ is a contiguous input subsequence and $y$ is its label/future sequence.
The cross-sample collection $\mathcal{B}$ is then partitioned into two subsets based on the occurrence of hypotensive events, i.e.,
\begin{equation}
\mathcal{B} = \mathcal{B}_{\text{hypo}} \cup \mathcal{B}_{\text{non-hypo}}.
\end{equation}

A hypotensive event is defined as a mean arterial pressure below 65 mmHg lasting at least one minute~\cite{pekingioh,ioh_define2}. During adaptation, CSA uses the target patient’s historical data to retrieve semantically relevant segments from a cross-sample bank, effectively combining personalized patterns with population-level insights. This strategy improves the model’s ability to handle sudden state changes and enhances generalization.

\subsection{Coarse-to-Fine Retrieval}
Direct retrieval of relevant samples from the cross-sample bank can be computationally expensive for large datasets. To alleviate this, we propose a coarse-to-fine retrieval strategy that enhances effectiveness while preserving relevance. It employs an adaptive context window to process the patient's streaming vital data. 
At each time step $t$, a present window of length $m$ is defined as the test window, while 
the previous window of the same length serves as the history window, denoted as:
\begin{equation}
\mathcal W^{\text{hist}}_{t-m:t} = [x_{t-m}, \ldots, x_{t-1}],
\end{equation}
where $x_t \in \mathbb{R}^C$ represents the $C$-dimensional vital signals collected at time $t$, and each window contains multiple time series samples. 

\textbf{Coarse-grained Retrieval}. To efficiently narrow the search space, CSA-TTA conducts clustering separately on the hypotensive subset $\mathcal{B}_{\text{hypo}}$ and the non-hypotensive subset $\mathcal{B}_{\text{non-hypo}}$.
Specifically, the K-Shape clustering\cite{kshape} is employed independently on each subset to identify representative cluster centroids. 
K-Shape is particularly well-suited for physiological time series data due to its shape-based clustering approach, which aligns temporal dynamics of sequences without requiring explicit time warping or amplitude normalization\cite{kshape_study1,kshape_study}. When a query sample \(s\) from the adaptive context window \(W^{\text{hist}}_{t-m:t}\) is available, CSA-TTA first determines its category and then computes its similarity to the cluster centroids of the corresponding subset. The query is then assigned to the cluster with the most semantically relevant centroid, thereby localizing its nearest temporal neighborhood within the large cross-sample bank. 

\textbf{Fine-grained Retrieval}. CSA-TTA refines the candidate selection by computing the semantic similarity between the query sample \(s\) and all samples within the cluster identified during the coarse retrieval phase from the cross-sample bank \(\mathcal{B}\). We employ Dynamic Time Warping (DTW) ~\cite{dtw,dtw1} as similarity metric for cross-sample retrieval. DTW measures the optimal alignment between two time series, allowing it to capture temporal shifts and variations, making it effective for identifying similarities in physiological signals~\cite{dtw2}. 
The top-\(K\) samples with the highest similarity scores are then retrieved to form a refined candidate set. Formally, the refined candidate set \(\mathcal{D}_{\mathrm{retrieval}}\) is given by:
\begin{equation}
\small
    \mathcal{D}_{\mathrm{retrieval}} = \left\{ r \in \mathcal{B},s \in W^{\mathrm{hist}}_{t-m:t} \mid r \in \operatorname{TopK}\left( \mathcal{B}, \mathrm{DTW}\left(r, s\right) \right) \right\},
\end{equation}
where \(\operatorname{TopK}(\mathcal{B}, \mathrm{DTW}(\cdot, \cdot))\) denotes the subset of \(K\) samples from \(\mathcal{B}\) having the highest semantic similarity scores with the query $s$, and \(\mathrm{DTW}(\cdot, \cdot)\) is a semantic similarity function between time series samples.
The retrieved samples are further augmented with perturbations (e.g., Gaussian noise, temporal scaling)
to increase variability and better simulate potential patient-specific variations. At each adaptation step, we construct a cross-sample augmented dataset by combining the patient's own history window with these augmented reference samples:
\begin{equation}
    \mathcal{D}^{\text{CSA-TTA}}_{t} = W^{\text{hist}}_{t-m:t} \cup \text{Aug} (\mathcal{D}_{\mathrm{retrieval}}).
\end{equation}

\subsection{Multi-task Optimization}

We adopt a multi-task optimization framework, i.e., the model is refined through two learning objectives: self-supervised masked reconstruction and retrospective sequence forecasting. As shown in Figure~\ref{fig:overview_pipeline}-(3), the architecture includes a shared feature encoder $f_\theta$, and two task-specific branches: $h_\theta$ for the primary prediction task and $g_\theta$ for the auxiliary self-supervised task.

The complete model is denoted as $F_\theta = (f_\theta, h_\theta, g_\theta)$. For self-supervision, we employ a masked reconstruction objective, a lightweight yet effective strategy for enhancing time-series representation learning~\cite{patchTST, moirai}. The model is trained to minimize a combined loss:
\begin{equation}
    \min_{f_\theta, h_\theta, g_\theta} \frac{1}{N} \sum_{n=1}^{N} \mathcal{L}_{\text{Pred}}(X_n, Y_n; f_\theta, h_\theta) + \mathcal{L}_{\text{Recon}}(X_n; f_\theta, g_\theta).
    \label{eq:fun_loss}
\end{equation}
The total loss $\mathcal{L}_{\text{CSA-TTA}}$ is computed over the cross-sample adaptation dataset, and model parameters are updated via gradient descent:
\begin{equation}
    \theta \leftarrow \theta - \eta \nabla_{\theta} \mathcal{L}_{\text{CSA-TTA}},
\end{equation}
where $\eta$ is the learning rate. To preserve generalization while enabling personalization, only a subset of parameters—typically in the input, output, and normalization layers—are updated during adaptation~\cite{membn}.

CSA-TTA supports two modes: fine-tuning and zero-shot. In the fine-tuning setting (Figure~\ref{fig:ioh_prediction}-(1)), both the main and auxiliary tasks are jointly fine-tuned offline before test-time adaptation, following~\cite{ttt}. In contrast, the zero-shot mode eliminates the need for offline fine-tuning or architectural changes; test-time adaptation is performed directly via retrospective regression. This allows CSA-TTA to remain effective even when auxiliary task training is impractical or historical data is unavailable.

To handle uncertainty and signal fluctuations during prediction, we implement a hybrid mechanism (Figure~\ref{fig:ioh_prediction}-(3)). Predicted blood pressure sequences are converted into point-wise risk scores representing the probability of MAP falling below a clinical threshold. We then apply two triggers: a hard trigger to detect sustained hypotensive periods and a soft trigger that evaluates average risk in sliding windows. These are combined to produce a final probabilistic estimate, balancing clinical reliability with model flexibility. 

\section{Experiments}

\subsection{Experimental Setups}

\paragraph{Dataset.} We conduct experiments using the VitalDB dataset, a real-world clinical database collected at Seoul National University Hospital. 
It contains vital sign data from 6,388 patients who underwent noncardiac surgeries between June 2016 and August 2017~\cite{vitaldb, pekingioh}.
From the raw records, we extract temporal features such as mean arterial pressure, body temperature, and heart rate, consistent with prior studies~\cite{cma}. Rigorous quality control is applied to exclude cases with over 20\% missing or abnormal values, as well as surgeries lasting less than one hour. This results in a curated dataset of 2,150 patient cases. To assess the impact of temporal resolution, the data are sampled at both 2-second and 30-second intervals.
The dataset is split at the patient level into training (70\%), validation (20\%), and test (10\%) subsets. The training and validation sets are primarily used for offline fine-tuning. The training set also serves to construct the cross-sample bank required for test-time adaptation. 
The test set is reserved exclusively for evaluating model performance under test-time adaptation using CSA-TTA.
In addition, we utilize a real-world clinical dataset from a collaborating hospital. The in-hospital test data are fully de-identified, and all experiments are conducted in compliance with the institution’s ethical review and data governance protocols. These data were collected from standard clinical devices—including blood pressure monitors, ventilators, and infusion pumps—and processed using the same quality control procedures as VitalDB. 
This dataset is sampled at one-minute intervals and comprises 130 cases for evaluation, with a additional 910 cases used exclusively to construct the cross-sample retrieval bank for test-time adaptation.

\newcommand{\up}{\textcolor{red}{$\blacktriangle$}}
\newcommand{\down}{\textcolor{green}{$\blacktriangledown$}}
\definecolor{lightgray}{gray}{0.9}
\setlength{\tabcolsep}{0.5mm}
\renewcommand{\arraystretch}{1.1}

\begin{table*}[t]
\centering
\small
\begin{subtable}[t]{\textwidth}
\centering
\caption*{(a) Zero-shot Setting}
\setlength{\tabcolsep}{0.5mm}
\begin{tabular}{ll|rrrrrr|rrrrrr}
\toprule
\multicolumn{2}{c|}{\textbf{Model}} & \multicolumn{6}{c|}{\textbf{VitalDB (2S)}} & \multicolumn{6}{c}{\textbf{VitalDB (30S)}} \\
\cmidrule(lr){3-8} \cmidrule(lr){9-14}
&& \textbf{F1$\uparrow$} & \textbf{Rec$\uparrow$} & \textbf{Prec$\uparrow$} & \textbf{Acc$\uparrow$} & \textbf{MAE$\downarrow$} & \textbf{MSE$\downarrow$}
   & \textbf{F1$\uparrow$} & \textbf{Rec$\uparrow$} & \textbf{Prec$\uparrow$} & \textbf{Acc$\uparrow$} & \textbf{MAE$\downarrow$} & \textbf{MSE$\downarrow$} \\
\midrule
\multirow{3}{*}{TimesFM} & Test\cite{timesfm} & 62.40 & \underline{64.97} & 60.53 & 88.07 & 6.19 & 83.61 & \underline{64.17} & \underline{58.87} & \underline{70.87} & \underline{87.60} & 6.49 & 92.77 \\
& \cellcolor{lightgray}CSA-TTA (Ours) & \cellcolor{lightgray}\textbf{64.10} & \cellcolor{lightgray}\textbf{66.09} & \cellcolor{lightgray}\textbf{62.88} & \cellcolor{lightgray}\textbf{88.40} & \cellcolor{lightgray}\textbf{6.07} & \cellcolor{lightgray}\textbf{80.55} & \cellcolor{lightgray}\textbf{64.90} & \cellcolor{lightgray}\textbf{59.27} & \cellcolor{lightgray}\textbf{72.20} & \cellcolor{lightgray}\textbf{87.93} & \cellcolor{lightgray}\textbf{6.28} & \cellcolor{lightgray}\textbf{85.28} \\
& \cellcolor{lightgray}$~~~~~~~~~~\Delta(\%)$ & \cellcolor{lightgray}\up1.70 & \cellcolor{lightgray}\up1.13 & \cellcolor{lightgray}\up2.35 & \cellcolor{lightgray}\up0.33 & \cellcolor{lightgray}\up1.94 & \cellcolor{lightgray}\up3.66 & \cellcolor{lightgray}\up0.73 & \cellcolor{lightgray}\up0.40 & \cellcolor{lightgray}\up1.33 & \cellcolor{lightgray}\up0.33 & \cellcolor{lightgray}\up3.24 & \cellcolor{lightgray}\up8.07 \\
\midrule
\multirow{3}{*}{Units} & Test\cite{units} & 49.67 & \underline{44.83} & \underline{58.44} & 87.53 & 7.36 & 98.88 & \underline{52.23} & \underline{43.24} & \underline{71.14} & 85.47 & 7.32 & 99.96 \\
& \cellcolor{lightgray}CSA-TTA (Ours) & \cellcolor{lightgray}\textbf{54.53} & \cellcolor{lightgray}\textbf{52.50} & \cellcolor{lightgray}\textbf{59.32} & \cellcolor{lightgray}\textbf{88.03} & \cellcolor{lightgray}\textbf{7.22} & \cellcolor{lightgray}\textbf{96.62} & \cellcolor{lightgray}\textbf{57.30} & \cellcolor{lightgray}\textbf{50.70} & \cellcolor{lightgray}66.56 & \cellcolor{lightgray}\textbf{85.83} & \cellcolor{lightgray}\textbf{7.19} & \cellcolor{lightgray}\textbf{95.84} \\
& \cellcolor{lightgray}$~~~~~~~~~~\Delta(\%)$ & \cellcolor{lightgray}\up4.87 & \cellcolor{lightgray}\up7.67 & \cellcolor{lightgray}\up0.88 & \cellcolor{lightgray}\up0.50 & \cellcolor{lightgray}\up1.90 & \cellcolor{lightgray}\up2.29 & \cellcolor{lightgray}\up5.07 & \cellcolor{lightgray}\up7.46 & \cellcolor{lightgray}\down4.58 & \cellcolor{lightgray}\up0.36 & \cellcolor{lightgray}\up1.78 & \cellcolor{lightgray}\up4.12 \\
\bottomrule
\end{tabular}
\end{subtable}

\begin{subtable}[t]{\textwidth}
\centering
\caption*{(b) Fine-tuned Setting}
\setlength{\tabcolsep}{0.5mm}
\begin{tabular}{ll|rrrrrr|rrrrrr}
\toprule
\multicolumn{2}{c|}{\textbf{Model}} & \multicolumn{6}{c|}{\textbf{VitalDB (2S)}} & \multicolumn{6}{c}{\textbf{VitalDB (30S)}} \\
\cmidrule(lr){3-8} \cmidrule(lr){9-14}
&& \textbf{F1$\uparrow$} & \textbf{Rec$\uparrow$} & \textbf{Prec$\uparrow$} & \textbf{Acc$\uparrow$} & \textbf{MAE$\downarrow$} & \textbf{MSE$\downarrow$}
   & \textbf{F1$\uparrow$} & \textbf{Rec$\uparrow$} & \textbf{Prec$\uparrow$} & \textbf{Acc$\uparrow$} & \textbf{MAE$\downarrow$} & \textbf{MSE$\downarrow$} \\
\midrule
\multirow{5}{*}{TimesFM} & Test\cite{timesfm} & 64.20 & 64.93 & 65.13 & 89.13 & 6.03 & 77.87 & \underline{65.80} & \textbf{62.67} & 69.97 & 87.77 & 5.94 & 76.27 \\
& TTT\cite{ttt} & 64.00 & 64.77 & 64.79 & 89.07 & 6.02 & 77.70 & 65.77 & \underline{62.63} & 69.93 & 87.77 & 5.94 & 76.28 \\
& TTT++\cite{ttt++} & 64.10 & 64.80 & 65.13 & 89.13 & 6.02 & 77.68 & \underline{65.80} & 62.60 & 70.00 & \underline{87.80} & 5.93 & 76.19 \\
& \cellcolor{lightgray}CSA-TTA (Ours) & \cellcolor{lightgray}\textbf{64.83} & \cellcolor{lightgray}\textbf{65.99} & \cellcolor{lightgray}\underline{65.42} & \cellcolor{lightgray}\textbf{89.40} & \cellcolor{lightgray}\textbf{5.94} & \cellcolor{lightgray}\textbf{76.19} & \cellcolor{lightgray}\textbf{66.07} & \cellcolor{lightgray}62.33 & \cellcolor{lightgray}\textbf{70.97} & \cellcolor{lightgray}\textbf{88.03} & \cellcolor{lightgray}\textbf{5.82} & \cellcolor{lightgray}\textbf{72.93} \\
& \cellcolor{lightgray}$~~~~~~~~~~\Delta(\%)$ & \cellcolor{lightgray}\up0.63 & \cellcolor{lightgray}\up1.06 & \cellcolor{lightgray}\up0.28 & \cellcolor{lightgray}\up0.27 & \cellcolor{lightgray}\up1.49 & \cellcolor{lightgray}\up2.16 & \cellcolor{lightgray}\up0.27 & \cellcolor{lightgray}\down0.34 & \cellcolor{lightgray}\up1.00 & \cellcolor{lightgray}\up0.26 & \cellcolor{lightgray}\up2.02 & \cellcolor{lightgray}\up4.38 \\
\midrule
\multirow{5}{*}{Units} & Test\cite{units} & 64.60 & \underline{63.93} & 67.13 & 89.50 & 5.80 & 71.65 & \underline{63.83} & \underline{59.07} & 70.00 & 87.97 & 6.37 & 81.77 \\
& TTT\cite{ttt} & 64.64 & 63.67 & 67.48 & \underline{89.57} & 5.82 & 71.58 & 63.47 & 55.23 & \underline{76.09} & 88.07 & 6.33 & 81.20 \\
& TTT++\cite{ttt++} & \underline{64.71} & 63.90 & 67.39 & 89.53 & 5.81 & \underline{71.45} & 63.17 & 56.66 & 73.23 & 88.00 & \underline{6.30} & 81.20 \\
& \cellcolor{lightgray}CSA-TTA (Ours) & \cellcolor{lightgray}\textbf{65.73} & \cellcolor{lightgray}\textbf{65.27} & \cellcolor{lightgray}\textbf{68.13} & \cellcolor{lightgray}\textbf{89.70} & \cellcolor{lightgray}\textbf{5.72} & \cellcolor{lightgray}\textbf{70.07} & \cellcolor{lightgray}\textbf{65.57} & \cellcolor{lightgray}\textbf{59.93} & \cellcolor{lightgray}72.88 & \cellcolor{lightgray}\textbf{88.47} & \cellcolor{lightgray}\textbf{6.23} & \cellcolor{lightgray}\textbf{79.31} \\
& \cellcolor{lightgray}$~~~~~~~~~~\Delta(\%)$ & \cellcolor{lightgray}\up1.13 & \cellcolor{lightgray}\up1.33 & \cellcolor{lightgray}\up0.99 & \cellcolor{lightgray}\up0.20& \cellcolor{lightgray}\up1.38 & \cellcolor{lightgray}\up2.21 & \cellcolor{lightgray}\up1.74 & \cellcolor{lightgray}\up0.86 & \cellcolor{lightgray}\up2.88 & \cellcolor{lightgray}\up0.50 & \cellcolor{lightgray}\up2.20 & \cellcolor{lightgray}\up3.01 \\
\bottomrule
\end{tabular}
\end{subtable}
\caption{Performance comparison under different adaptation settings: (a) zero-shot setting; (b) fine-tuning setting. All results are averaged over three predict horizons (5, 10, and 15 minutes), using a fixed lookback length of 15 minutes. The best performance is highlighted in \textbf{bold}, and the second-best is \underline{underlined}. Metrics with the $\uparrow$ symbol (e.g., F1, Recall) indicate higher is better, while those with $\downarrow$ (e.g., MAE, MSE) indicate lower is better. Performance changes are marked with \up~for improvement and \down~for degradation. 
}
\label{tab:main_results_combined}
\end{table*}

\paragraph{Baselines.}
We use two state-of-the-art pre-trained time series models as backbones: TimesFM~\cite{timesfm}, a decoder-only foundation model trained on a large corpus of real and synthetic time series, and UniTS~\cite{units}, a unified multi-task model designed for diverse time series applications. We compare against two test-time training baselines: TTT~\cite{ttt}, which adapts the model during inference via a simple self-supervised task (e.g., rotation prediction) to mitigate distribution shifts, and TTT++~\cite{ttt++}, which enhances this by incorporating feature alignment to improve robustness and reduce catastrophic forgetting during adaptation.


\definecolor{lightgray}{gray}{0.9}
\setlength{\tabcolsep}{0.5mm}
\renewcommand{\arraystretch}{1.1}

\begin{table*}[ht]
\centering
\begin{tabular}{ll|rrrrrr|rrrrrr}
\toprule
\multicolumn{2}{c|}{\textbf{Model}} & \multicolumn{6}{c|}{\textbf{TimesFM\cite{timesfm}}} & \multicolumn{6}{c}{\textbf{UniTS\cite{units}}} \\
\cmidrule(lr){3-8} \cmidrule(lr){9-14}
&& \textbf{F1$\uparrow$} & \textbf{Rec$\uparrow$} & \textbf{Prec$\uparrow$} & \textbf{Acc$\uparrow$} & \textbf{MAE$\downarrow$} & \textbf{MSE$\downarrow$}
   & \textbf{F1$\uparrow$} & \textbf{Rec$\uparrow$} & \textbf{Prec$\uparrow$} & \textbf{Acc$\uparrow$} & \textbf{MAE$\downarrow$} & \textbf{MSE$\downarrow$} \\
\midrule
\multirow{4}{*}{Zero-shot} 
& Test & 70.23 & \underline{60.77} & 83.28 & \underline{86.63} & 6.00 & 88.02 
& 56.10 & \underline{43.77} & 80.09 & \underline{82.93} & 6.45 & 91.97 \\
& Random samples & \underline{70.30} & 60.70 & \underline{83.53} & 86.57 & \underline{5.93} & \underline{86.40} 
& \underline{56.17} & 42.03 & \textbf{85.43} & \underline{83.10} & \underline{6.40} & \underline{90.12} \\
& \cellcolor{lightgray}CSA-TTA (Ours) & \cellcolor{lightgray}\textbf{71.60} & \cellcolor{lightgray}\textbf{62.44} & \cellcolor{lightgray}\textbf{84.03} & \cellcolor{lightgray}\textbf{87.20} & \cellcolor{lightgray}\textbf{5.75} & \cellcolor{lightgray}\textbf{83.37} & \cellcolor{lightgray}\textbf{63.80} & \cellcolor{lightgray}\textbf{53.33} & \cellcolor{lightgray}\underline{80.43} & \cellcolor{lightgray}\textbf{84.47} & \cellcolor{lightgray}\textbf{6.30} & \cellcolor{lightgray}\textbf{88.69} \\
& \cellcolor{lightgray}~~~~~~~~$\Delta(\%)$ & \cellcolor{lightgray}\up1.37 & \cellcolor{lightgray}\up1.67 & \cellcolor{lightgray}\up0.75 & \cellcolor{lightgray}\up0.57 & \cellcolor{lightgray}\up4.17 & \cellcolor{lightgray}\up5.28 & \cellcolor{lightgray}\up7.70 & \cellcolor{lightgray}\up9.56 & \cellcolor{lightgray}\up0.34 & \cellcolor{lightgray}\up1.54 & \cellcolor{lightgray}\up2.33 & \cellcolor{lightgray}\up3.57 \\
\bottomrule
\end{tabular}
\caption{Performance comparison on the in-hospital test set. All results are averaged over prediction horizons (5, 10, and 15 minutes), using a fixed 15-minute lookback window.}
\label{tab:zero_shot_results}
\end{table*}

\paragraph{Implementation Details.}
We evaluate model performance using both regression and classification metrics. For regression, we use Mean Absolute Error (MAE) and Mean Squared Error (MSE). For classification, we report Accuracy, Recall, Precision, and F1-score.
For all experiments, we set the lookback window length \( L \) to 15 minutes and vary the forecast horizon \(H\) among 5, 10, and 15 minutes.  
During the fine-tuning stage, models are trained on training data for 10 epochs with a learning rate of \(1 \times 10^{-4}\), batch size of 64, and dropout rate of 0.01.
During test-time adaptation, models are updated for one epoch in the fine-tuning setting and three epochs in the zero-shot setting. We apply a partial fine-tuning strategy, where only the parameters in the input layer, output layer, and layer normalization layers of the backbone are updated, to balance adaptability with generalization. To ensure fairness, the same architecture and hyperparameter configurations are applied across all datasets. 
All experiments were accelerated by four NVIDIA A100 GPUs.

\subsection{Performance Comparison}

\paragraph{Zero-Shot Setting.}
In the zero-shot setting, where no fine-tuning is performed before test-time adaptation, CSA-TTA achieves significant improvements. 
As shown in Table~\ref{tab:main_results_combined}(a),  TimesFM~+~CSA-TTA improves Recall by 1.13\% and F1 by 1.70\% on VitalDB(2S). Units~+~CSA-TTA shows even greater performance, with Recall increasing by 7.46\% and F1 improving by 5.07\% on VitalDB(30S). 
Similar improvements are observed on the in-hospital test set, Units~+~CSA-TTA improves Recall by 9.56\% (43.77\% → 53.33\%) and F1 by 7.70\% (56.10\% → 63.80\%), as shown in Table~\ref{tab:zero_shot_results}. This substantial improvement in Recall and F1 is clinically significant for intraoperative hypotension prediction, allowing it to more effectively identify hypotension events and reduce missed diagnosis risks. 
Additionally, CSA-TTA demonstrates notable regression performance, with reductions in MAE and MSE of 3.24\% (6.49 → 6.28) and 8.07\% (92.77 → 85.28) on VitalDB, and 4.17\% (6.00 → 5.75) and 5.28\% (88.02 → 83.37) on the in-hospital test set, respectively.
These improvements underscore CSA-TTA's ability to effectively address the challenges of personalized IOH dynamics, enhancing predictive accuracy and reducing regression errors by adapting to rapid signal shifts and rare hypotensive events.

\definecolor{checkmark}{rgb}{0.1,0.8,0.1}
\definecolor{crossmark}{rgb}{1,0.1,0.1}
\definecolor{lightgray}{gray}{0.9}
\begin{table}[t]
\centering
\small
\begin{tabular}{l|ll|rrrrrr}
\toprule
\multicolumn{3}{c|}{\textbf{Config}} & \multicolumn{6}{c}{\textbf{VitalDB Dataset}} \\ 
\cmidrule(lr){1-3} \cmidrule(lr){4-9}
\textbf{Horizon} & \textbf{Pred} & \textbf{Recon} & \textbf{F1$\uparrow$} & \textbf{Rec$\uparrow$} & \textbf{Prec$\uparrow$} & \textbf{Acc$\uparrow$} & \textbf{MAE$\downarrow$} & \textbf{MSE$\downarrow$} \\
\midrule
\multirow{3}{*}{5min} 
 & \textcolor{crossmark}{\ding{55}} & \textcolor{checkmark}{\ding{51}} & 70.00 & \textbf{71.90} & 68.20 & 91.50 & 4.82 & 55.81 \\
 & \textcolor{checkmark}{\ding{51}} & \textcolor{crossmark}{\ding{55}} & \textbf{70.60} & 71.40 & \textbf{69.80} & \textbf{91.80} & \underline{4.79} & \underline{54.08} \\
 & \cellcolor{lightgray}\textcolor{checkmark}{\ding{51}} & \cellcolor{lightgray}\textcolor{checkmark}{\ding{51}} & \cellcolor{lightgray}\textbf{70.60} & \cellcolor{lightgray}\underline{71.60} & \cellcolor{lightgray}\underline{69.70} & \cellcolor{lightgray}\textbf{91.80} & \cellcolor{lightgray}\textbf{4.77} & \cellcolor{lightgray}\textbf{53.17} \\
\hline
\multirow{3}{*}{10min} 
 & \textcolor{crossmark}{\ding{55}} & \textcolor{checkmark}{\ding{51}} & \underline{64.60} & \underline{60.50} & \underline{69.30} & 87.30 & 6.20 & 82.14 \\
 & \textcolor{checkmark}{\ding{51}} & \textcolor{crossmark}{\ding{55}} & 64.40 & 59.50 & \textbf{70.20} & \underline{87.40} & \underline{6.15} & \underline{80.49} \\
 & \cellcolor{lightgray}\textcolor{checkmark}{\ding{51}} & \cellcolor{lightgray}\textcolor{checkmark}{\ding{51}} & \cellcolor{lightgray}\textbf{64.70} & \cellcolor{lightgray}\textbf{60.80} & \cellcolor{lightgray}69.20 & \cellcolor{lightgray}\textbf{87.50} & \cellcolor{lightgray}\textbf{6.05} & \cellcolor{lightgray}\textbf{77.60} \\
\hline
\multirow{3}{*}{15min} 
 & \textcolor{crossmark}{\ding{55}} & \textcolor{checkmark}{\ding{51}} & 62.80& \textbf{55.30} & 72.70 & 84.60 & 6.70 & 90.02 \\
 & \textcolor{checkmark}{\ding{51}} & \textcolor{crossmark}{\ding{55}} & 62.80 & 54.59 & \textbf{74.00} & \textbf{84.80} & \underline{6.69} & \underline{89.29} \\
 & \cellcolor{lightgray}\textcolor{checkmark}{\ding{51}} & \cellcolor{lightgray}\textcolor{checkmark}{\ding{51}} & \cellcolor{lightgray}\textbf{62.90} & \cellcolor{lightgray}\underline{54.60} & \cellcolor{lightgray}\textbf{74.00} & \cellcolor{lightgray}\textbf{84.80} & \cellcolor{lightgray}\textbf{6.64} & \cellcolor{lightgray}\textbf{88.02} \\ 
\bottomrule
\end{tabular}
\caption{Ablation study on multi-task optimization. Performance of TimesFM in the fine-tuning setting using different optimization strategies: Pred (supervised prediction) and Recon (self-supervised masked reconstruction). }
\label{tab:multitask_ablation}
\end{table}

\paragraph{Fine-Tuning Setting.}
CSA-TTA shows consistent improvements across all performance metrics. 
As shown in Table~\ref{tab:main_results_combined}(b),  TimesFM~+~CSA-TTA improves Recall by 1.06\% and F1 by 0.63\% on VitalDB (2S), surpassing all other test-time adaptation methods. Units~+~CSA-TTA achieves even greater gains, with Recall increasing by 1.33\%, F1 by 1.13\%, Precision by 0.99\%, and Accuracy by 0.20\%, demonstrating consistent improvements across multiple metrics. Similar trends appear on VitalDB (30S), where Units~+~CSA-TTA raises Recall by 0.86\% and F1 by 1.74\%.
Regarding regression metrics, CSA-TTA further improves performance, with the TimesFM-based model reducing MAE and MSE by 2.02\% and 4.38\%, respectively, on VitalDB (30S).
These improvements, particularly in Recall, are clinically significant. In intraoperative hypotension prediction, higher Recall reduces false negatives—missed hypotensive events that can lead to myocardial injury or death. Even modest Recall gains can improve patient outcomes. CSA-TTA enhances Recall while maintaining stable Precision, helping to reduce unnecessary interventions.

\begin{figure}[!t]
    \centering
    \begin{subfigure}{\linewidth}
        \centering
        \includegraphics[width=0.95\linewidth]{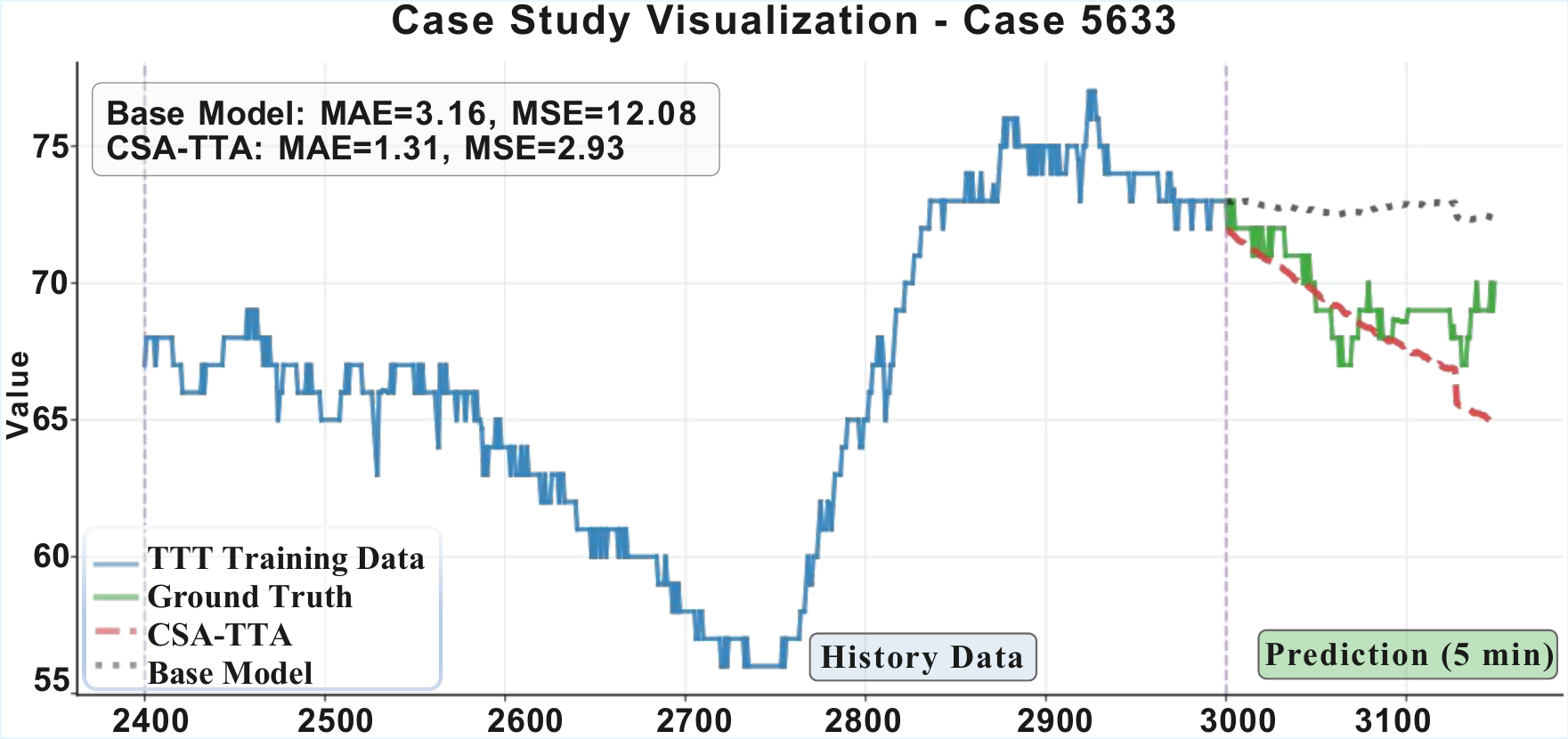}
        \caption{5-min horizon}
        \label{fig:case_study_a}
    \end{subfigure}

    \vspace{0.3em} 

    \begin{subfigure}{\linewidth}
        \centering
        \includegraphics[width=0.95\linewidth]{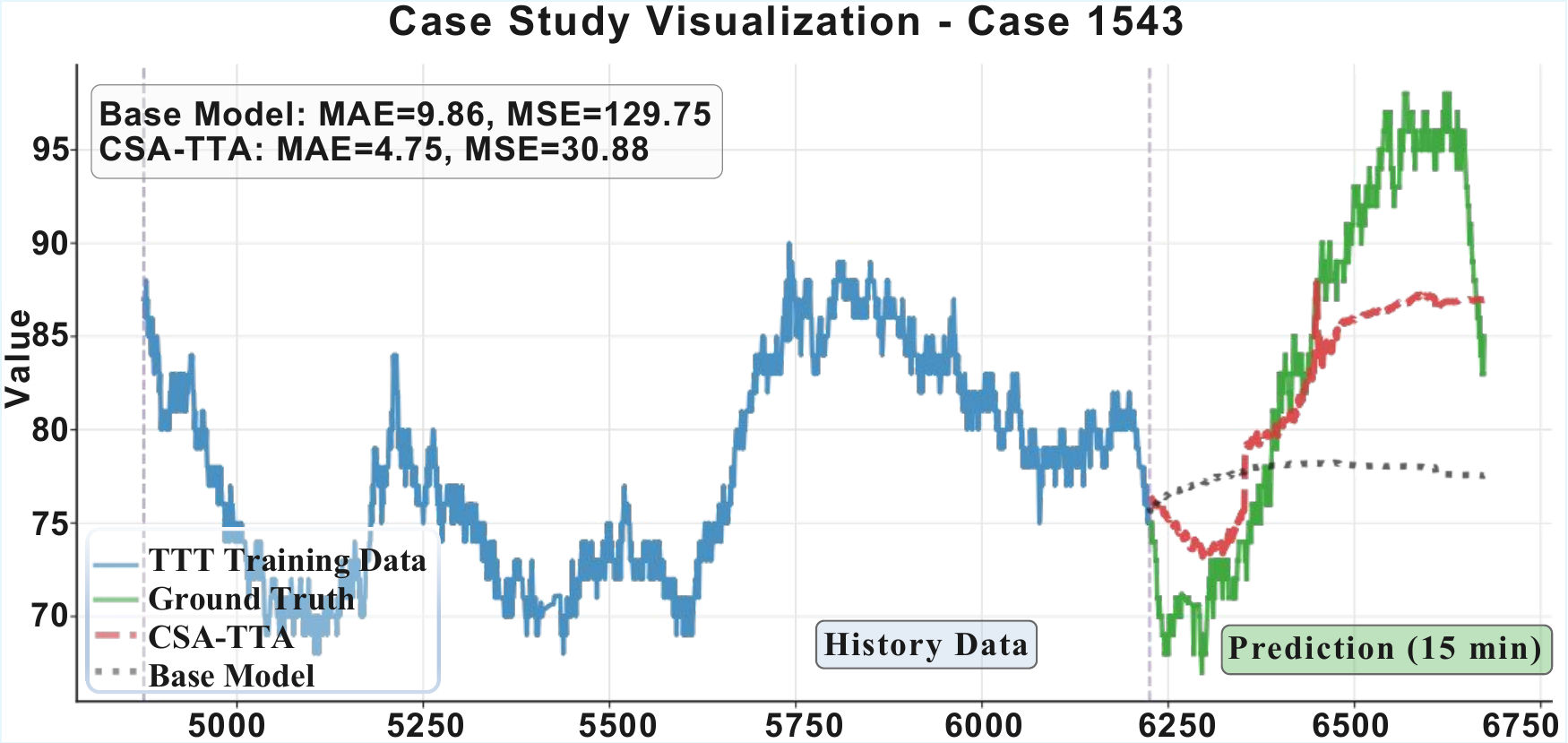}
        \caption{15-min horizon}
        \label{fig:case_study_b}
    \end{subfigure}

\caption{Case study visualizations on VitalDB (2-second sampling). TimesFM~+~CSA-TTA predicts 5-minute (top) and 15-minute (bottom) horizons.
}
\label{fig:case_study}
\end{figure}

\paragraph{Case Study.}
To compare CSA-TTA with the baseline, we evaluate vanilla TimesFM and TimesFM~+~CSA-TTA on two VitalDB (2S) cases. At the 5-minute horizon (Figure~\ref{fig:case_study_a}), the static model smooths the blood pressure drop, while CSA-TTA—guided by cross-sample retrieval and multi-task adaptation—closely follows the ground truth. This advantage continues at the 15-minute horizon (Figure~\ref{fig:case_study_b}), where CSA-TTA accurately captures the sharp decline and rebound, reducing errors by more than half (MAE: 9.86 → 4.75; MSE: 129.75 → 30.88). These examples show how incorporating patient-specific context at inference allows TimesFM to model abrupt, individualized IOH dynamics missed by static models. 


\begin{table}[t]
\centering
\begin{tabular}{l|l|rrrrrr}
\toprule
\multicolumn{2}{c|}{\textbf{Config}} & \multicolumn{6}{c}{\textbf{Zero-Shot Setting}}\\ 
\cmidrule(lr){1-2} \cmidrule(lr){3-8}
\textbf{Model} & \textbf{Top-K} & \textbf{F1$\uparrow$} & \textbf{Rec$\uparrow$} & \textbf{Prec$\uparrow$} & \textbf{Acc$\uparrow$} & \textbf{MAE$\downarrow$} & \textbf{MSE$\downarrow$} \\
\midrule
\multirow{3}{*}{TimesFM} 
 & 1 & 64.57 & 58.83 & 71.97 & \underline{87.83} & 6.32 & 86.96\\
 & 2 & \underline{64.73} & \underline{59.03} & \underline{72.13} & 87.80 & \textbf{6.27} & \underline{86.28} \\
 & \cellcolor{lightgray}3 & \cellcolor{lightgray}\textbf{64.90} & \cellcolor{lightgray}\textbf{59.27} & \cellcolor{lightgray}\textbf{72.20} & \cellcolor{lightgray}\textbf{87.93} & \cellcolor{lightgray}\underline{6.28} & \cellcolor{lightgray}\textbf{85.28} \\
\bottomrule
\end{tabular}
\caption{Ablation study on Top-K of coarse-to-fine cross-sample retrieval. Performance of the TimesFM backbone under zero-shot settings.}
\label{tab:top_k_performance}
\end{table}

\begin{table}[t]
\centering
\setlength{\tabcolsep}{7.5pt}
\small
\begin{tabular}{lrrrr}
\toprule
Model & \#ToP & \#TuP & \#PuP & \#Time(Avg) \\
\midrule
TimesFM & 477.36M & 5.05M & 1.06\% & 6.587s \\
Units   & 1.01M   & 0.05M & 5.44\% & 1.688s \\
\bottomrule
\end{tabular}
\caption{Computational cost of CSA-TTA per epoch (30-minute history window). ``\#ToP" and ``\#TuP" indicate the total and fine-tuned parameter counts, respectively. ``\#PuP" is the percentage of parameters updated, and ``\#Time" denotes the average wall-clock time per TTA epoch.}
\label{tab:compute_cost}
\end{table}

\subsection{Ablation Study} 

\paragraph{Multi-Task Optimization.} 
Table~\ref{tab:multitask_ablation} compares supervised prediction optimization (``Pred"), self-supervised reconstruction optimization (``Recon"), and their combination on TimesFM under the fine-tuning setting across 5-, 10-, and 15-minute horizons.
Multi-task optimization consistently outperforms the single-task variants, achieving the best F1, MAE, and MSE across all settings. It also yields the highest Recall at the 10-minute horizons (60.80\%).
While single-task setups occasionally show higher Recall—for instance, ``Recon"-only achieves 71.90\% at 5 minutes—they perform worse on other metrics.
These results highlight the effectiveness of multi-task optimization in adapting to personalized intraoperative hypotension dynamics.

\paragraph{Top-\emph{K} of retrieval.}
We ablate the Top-$K$ hyperparameter in the fine-grained retrieval stage of CSA-TTA using a TimesFM backbone on VitalDB under zero-shot settings. As shown in Table~\ref{tab:top_k_performance}, with results averaged over 5/10/15-minute horizons, increasing \emph{K} yields improved or stable F1, Recall, and regression errors. For example, Recall rises from 58.83 (\emph{K}=1) to 59.27 (\emph{K}=3), while MSE drops from 86.96 to 85.28. We therefore adopt \emph{K}=3, which balances relevance and diversity and improves robustness and accuracy.


\begin{table}[t]
\centering
\small
\begin{tabular}{l|ll|rrrrrr}
\toprule
\multicolumn{3}{c|}{\textbf{Config}} & \multicolumn{6}{c}{\textbf{VitalDB Dataset}} \\ 
\cmidrule(lr){1-3} \cmidrule(lr){4-9}
\textbf{Setting} & \textbf{Bank} & \textbf{Aug} & \textbf{F1$\uparrow$} & \textbf{Rec$\uparrow$} & \textbf{Prec$\uparrow$} & \textbf{Acc$\uparrow$} & \textbf{MAE$\downarrow$} & \textbf{MSE$\downarrow$} \\
\midrule
\multirow{4}{*}{Zero-shot} 
 & \textcolor{crossmark}{\ding{55}} & \textcolor{crossmark}{\ding{55}} & 64.47 & 58.67 & 71.90 & 87.80 & 6.33 & 86.70 \\ 
 & \textcolor{crossmark}{\ding{55}} & \textcolor{checkmark}{\ding{51}} & 64.27 & 57.60 & \textbf{72.97} & \underline{87.83} & \underline{6.27} & 85.66 \\
 & \textcolor{checkmark}{\ding{51}} & \textcolor{crossmark}{\ding{55}} & \textbf{64.97} & \textbf{60.13} & 71.13 & 87.77 & \textbf{6.26} & \textbf{85.03} \\
 & \cellcolor{lightgray}\textcolor{checkmark}{\ding{51}} & \cellcolor{lightgray}\textcolor{checkmark}{\ding{51}} & \cellcolor{lightgray}\underline{64.90} & \cellcolor{lightgray}\underline{59.27} & \cellcolor{lightgray}\underline{72.20} & \cellcolor{lightgray}\textbf{87.93} & \cellcolor{lightgray}6.28 & \cellcolor{lightgray}\underline{85.28} \\
\hline
\multirow{4}{*}{fine-tuned} 
 & \textcolor{crossmark}{\ding{55}} & \textcolor{crossmark}{\ding{55}} & 65.90 & 62.17 & 70.82 & 87.90 & 5.87 & 74.79 \\
 & \textcolor{crossmark}{\ding{55}} & \textcolor{checkmark}{\ding{51}} & 65.72 & 61.38 & \textbf{71.43} & 87.97 & \underline{5.84} & \underline{73.95} \\
 & \textcolor{checkmark}{\ding{51}} & \textcolor{crossmark}{\ding{55}} & \underline{66.03} & \underline{62.20} & \underline{71.10} & \underline{88.00} & 5.85 & 74.19 \\
 & \cellcolor{lightgray}\textcolor{checkmark}{\ding{51}} & \cellcolor{lightgray}\textcolor{checkmark}{\ding{51}} & \cellcolor{lightgray}\textbf{66.07} & \cellcolor{lightgray}\textbf{62.33} & \cellcolor{lightgray}70.97 & \cellcolor{lightgray}\textbf{88.03} & \cellcolor{lightgray}\textbf{5.82} & \cellcolor{lightgray}\textbf{72.93} \\ 
\bottomrule
\end{tabular}
\caption{Ablation study on data augmentation strategy. Performance of TimesFM under zero-shot and fine-tuned settings using different augmentation methods: Bank (cross-sample bank) and Aug (perturbation-based augmentation).}
\label{tab:data_augmentation_ablation}
\end{table}

\paragraph{Augmentation Strategy.}
Integrating the cross-sample bank with small perturbations consistently enhances model performance. As shown in Table~\ref{tab:data_augmentation_ablation}, using the cross-sample bank alone improves F1 by 0.5\% (64.47\% → 64.97\%) and Recall by 1.46\% in the zero-shot setting on the VitalDB dataset. Adding small perturbations further boosts performance—under the fine-tuning setting, the combined approach increases Recall by 0.13\% (62.20\% → 62.33\%) and reduces MAE by 1.69\% (74.19 → 72.93).
These results highlight that while each augmentation method offers individual benefits, their combination yields the best overall performance.

\begin{table}[t]
  \centering
  \small
  \setlength{\tabcolsep}{0.42mm}
  \renewcommand{\arraystretch}{1.1}
  \begin{tabular}{ll|rrrrrr}
    \toprule
    \multicolumn{2}{c|}{\textbf{Setting / Data}} &
      \multicolumn{6}{c}{\textbf{VitalDB Dataset}} \\
    \cmidrule(lr){3-8}
    && \textbf{F1$\uparrow$} & \textbf{Recall$\uparrow$} & \textbf{Prec$\uparrow$} & \textbf{Acc$\uparrow$} & \textbf{MAE$\downarrow$} & \textbf{MSE$\downarrow$} \\
    \midrule
    \multirow{2}{*}{Zero-shot}
      & Random & 63.93 & 58.73 & 70.57 & 87.60 & 6.47 & 92.30 \\
      & \cellcolor{lightgray}CSA-TTA & \cellcolor{lightgray}\textbf{64.90} & \cellcolor{lightgray}\textbf{59.27} & \cellcolor{lightgray}\textbf{72.20} & \cellcolor{lightgray}\textbf{87.93} & \cellcolor{lightgray}\textbf{6.28} & \cellcolor{lightgray}\textbf{85.28} \\
    \midrule
    \multirow{2}{*}{Fine-tuned}
      & Random & 65.73 & 62.13 & 70.39 & 87.77 & 5.92 & 75.84 \\
      & \cellcolor{lightgray}CSA-TTA & \cellcolor{lightgray}\textbf{66.07} & \cellcolor{lightgray}\textbf{62.33} & \cellcolor{lightgray}\textbf{70.97} & \cellcolor{lightgray}\textbf{88.03} & \cellcolor{lightgray}\textbf{5.82} & \cellcolor{lightgray}\textbf{72.93} \\
    \bottomrule
  \end{tabular}
  \caption{Ablation study on personalized adaptation. Performance of TimesFM~+~CSA-TTA with random data and personalized cross-sample data.}
  \label{tab:random_samples}
\end{table}

\paragraph{Computational Cost.}
To assess the computational cost, we report adaptation time in Table~\ref{tab:compute_cost}. By using partial fine-tuning (updating only 1.06\%–5.44\% of parameters), it enables efficient adaptation, with per-epoch latency as low as 1.7 seconds for lightweight models and 6.6 seconds for larger ones. This demonstrates its efficiency and practicality for real-world deployment.

\paragraph{Personalized Adaptation.} 


To validate the effectiveness of personalized adaptation, we compare against a “Random” baseline—CSA-TTA without personalized history or augmentation, using the same number of randomly selected sequences from other patients. 
Table~\ref{tab:random_samples} shows that, on VitalDB with TimesFM~+~CSA-TTA, F1 rises from 63.93 to 64.90 with a concurrent MSE drop from 92.30 to 85.28 in the zero-shot setting, while fine-tuning yields smaller yet consistent gains (F1: 65.73 → 66.07; MSE: 75.84 → 72.93). 
These results demonstrate that leveraging personalized history aligns the model more closely with the target distribution than random sampling.



\section{Conclusions}
We present CSA-TTA, the first test-time adaptation framework specifically designed for personalized intraoperative hypotension (IOH) prediction. CSA-TTA addresses the challenges of individual IOH dynamics by leveraging a cross-sample bank with a coarse-to-fine retrieval strategy to provide richer temporal context during adaptation. It further combines supervised and self-supervised objectives in a unified optimization framework to enhance adaptation stability and effectiveness. 
Extensive experiments on two real-world clinical datasets, using TimesFM and UniTS as backbones, demonstrate that CSA-TTA consistently improves prediction performance, offering a robust and efficient solution for real-time, personalized monitoring in surgical settings. 


\section{Limitations and Future Work} 
While CSA-TTA consistently improves both regression and classification metrics across different backbones and datasets under a patient-level split, the current evaluation is still based on a limited number of random partitions. A more systematic assessment of robustness and generalization—e.g., repeated resampling schemes combined with paired statistical tests such as the Wilcoxon signed-rank test across multiple splits—will be valuable in future work. In addition, the effectiveness of CSA-TTA implicitly relies on each test case being reasonably close to at least one cluster represented in the training-derived sample bank; truly out-of-distribution or highly idiosyncratic episodes may therefore be less amenable to retrieval-based adaptation. Explicitly characterizing such OOD scenarios and designing appropriate fallback or uncertainty-aware strategies is an important direction for further study.

Moreover, the current cross-sample bank is constructed from a static cohort and used primarily to demonstrate the feasibility of retrieval-augmented test-time adaptation, without yet modeling how the bank should evolve in a continuously changing clinical environment. Future work could investigate dynamic maintenance of the sample bank, including recency-aware updates, bias and fairness considerations across demographic and surgical subgroups, and more principled control of which patients contribute to adaptation. Finally, while this study focuses on intraoperative hypotension prediction as a clinically important use case, the proposed CSA-TTA framework is not task-specific; extending and validating it on other perioperative or critical-care time-series tasks (e.g., arrhythmia, or postoperative deterioration) is a natural next step toward broader clinical applicability.

\section{Acknowledgments}
This work was partially supported by the Yunnan Provincial Department of Science and Technology-Major Science and Technology Special Project ($202502AS080002$), funding from the National Natural Science Foundation ($81860218$), the Yunling Scholar Talent Program of Yunnan Province under Grant No. ($K264202230207$), and the Deng Cheng Expert Workstation of Yunnan Province ($202305AF150202$).



\bibliography{aaai2026}

\clearpage
\clearpage
\twocolumn[{
    \begin{center}
        \huge{\textbf{Appendix}}
    \end{center}
    \vspace{1.5em}
}]
\section{\textbf{Appendix A}}
\vspace{1em}

\paragraph{Coarse-to-fine cross-sample retrieval procedure.}
To reduce the computational cost of direct retrieval from the large-scale cross-sample bank, CSA-TTA adopts a coarse-to-fine retrieval strategy that balances efficiency and relevance. At each time step $t$, a history window of length $m$ is defined as the retrieval-sample query window, denoted as:
\begin{align}
\small
W^{\text{hist}}_{t-m:t} &= [x_{t-m}, \ldots, x_{t-1}].
\end{align}

When the history window is available, it is segmented into \(h\) temporal fragments of length \(n =|X_n| + |Y_n|\), forming a set of query samples:
\begin{equation}
\small
\mathcal{S}^{\text{query}} = \{s_i\}_{i=1}^h, \quad s_i \in \mathbb{R}^{n \times C}.
\end{equation}

CSA-TTA first performs \textbf{coarse retrieval} by applying K-Shape clustering\cite{kshape} on the hypotensive subset $\mathcal{B}_{\text{hypo}}$ and the non-hypotensive subset $\mathcal{B}_{\text{non-hypo}}$ to obtain representative cluster centroids. Formally, the cluster centroids are given by:
\begin{equation}
\small
\mathcal{C}_{\text{hypo}}^i = \text{KShape}(\mathcal{B}_{\text{hypo}}), \quad \mathcal{C}_{\text{non-hypo}}^i = \text{KShape}(\mathcal{B}_{\text{non-hypo}}),
\end{equation}
where \(\mathcal{C}_{\text{hypo}}^i\) and \(\mathcal{C}_{\text{non-hypo}}^i\) denote the sets of cluster centroids from hypotensive and non-hypotensive sample banks, respectively. 
The query \({S}^{\text{query}}\) is then assigned to the cluster with the most semantically relevant centroid \(\mathcal{C}_{\text{hypo}}\) and \(\mathcal{C}_{\text{non-hypo}}\), thereby localizing its nearest temporal neighborhood within the large cross-sample bank. 

Subsequently, in the \textbf{fine-grained retrieval}, the top-\(K\) samples with the highest similarity scores are then retrieved to form a refined candidate set \(\mathcal{D}_{\mathrm{retrieval}}\):
\begin{equation}
\begin{split}
\mathcal{D}_{\mathrm{retrieval}} = \biggl\{ 
    &\mathcal{D}_{\text{retrieval}}^{\text{hypo}} =\operatorname{TopK} \left( \mathrm{sim}\left(\mathcal{C}_{\text{hypo}}, \mathcal{S}^{\text{query}} \right) \right), \\
    &\mathcal{D}_{\text{retrieval}}^{\text{non-hypo}} =\operatorname{TopK} \left( \mathrm{sim}\left(\mathcal{C}_{\text{non-hypo}}, \mathcal{S}^{\text{query}} \right) \right) 
\biggr\},
\end{split}
\end{equation}
where \(\operatorname{TopK} \left( \mathrm{sim}\left(\cdot, \cdot\right) \right)\) denotes the subset of \(K\) samples from \(\mathcal{C}_{\text{hypo}}\) and \(\mathcal{C}_{\text{non-hypo}}\) having the highest semantic similarity scores with the query \({S}^{\text{query}}\), and \(\mathrm{sim}(\cdot, \cdot)\) is a semantic similarity function between time series samples.
Finally, the retrieved candidates across all \(h\) segments are aggregated and rebalanced according to a predefined hpyo-to-nonhypo ratio (e.g, hypo:non-hypo=3 : 4) to form the retrieval dataset Drctrieval.
\begin{equation}
\small
\mathcal{D}_{\text{retrieval}} = \text{SampleBalance}\left( \mathcal{D}_{\text{retrieval}}^{\text{hypo}}, 
\mathcal{D}_{\text{retrieval}}^{\text{non-hypo}}, P:N \right).
\end{equation}

The final adaptation dataset used for test-time training is constructed by augmenting the retrieval samples and combining them with the segmented history window:
\begin{equation}
\small
\mathcal{D}^{\text{CSA-TTA}}_t = \mathcal{S}^{\text{query}} \cup \text{Aug}(\mathcal{D}_{\text{retrieval}}),
\end{equation}
where \(\text{Aug}(\cdot)\) denotes sample-level augmentation such as Gaussian noise or temporal warping.

\begin{figure}[t!]
    \centering
    \includegraphics[width=\linewidth]{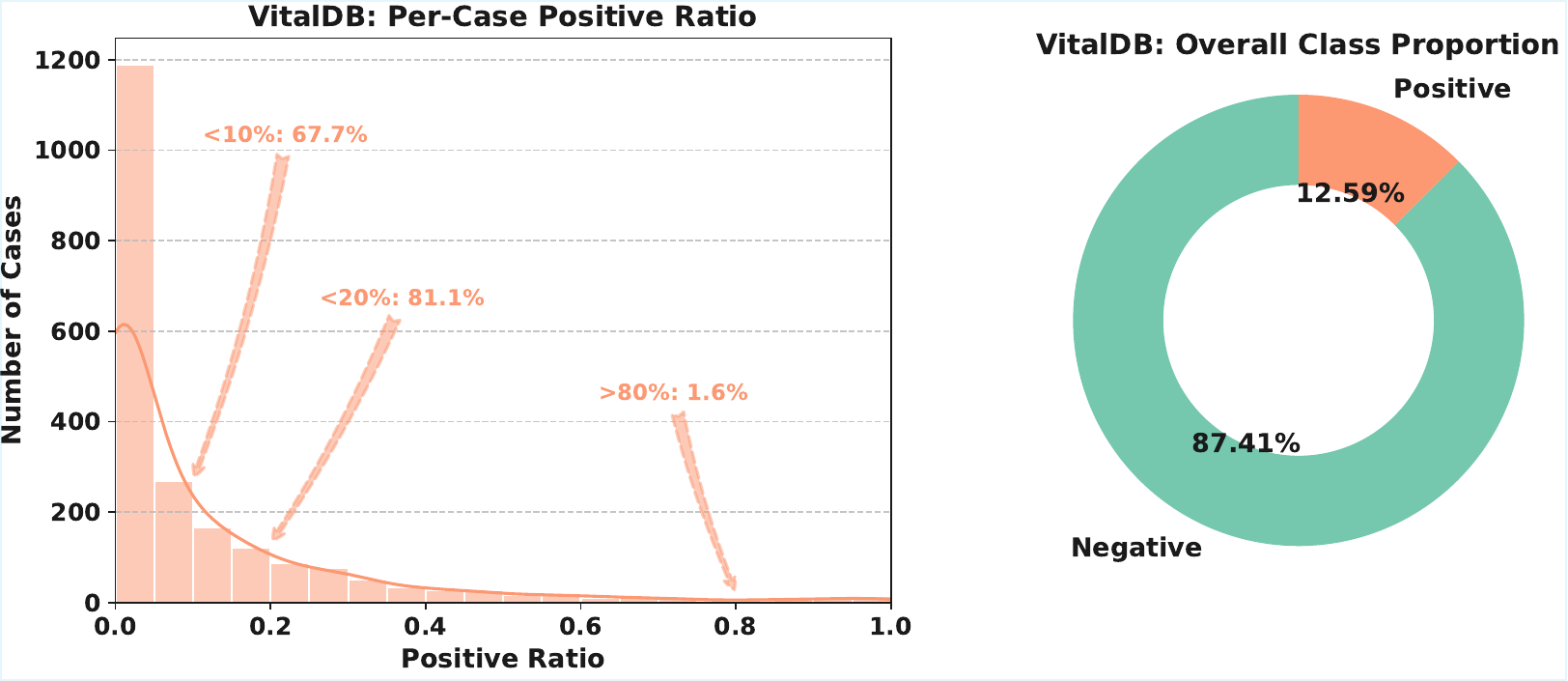}
    \caption{\textbf{Distribution of hypotension event ratios per patient case and overall class proportions in the VitalDB dataset}. The left panel shows the distribution of hypotension event (considered positive sample) ratios for each case, with arrows marking the proportions of cases below 10\%, below 20\%, and above 80\%. The right panel illustrates the global proportion of positive and negative samples for the VitalDB dataset.
    }
    \label{fig:dataset_statistics}
\end{figure}

\begin{figure*}[t!]
    \centering
    \includegraphics[width=\linewidth]{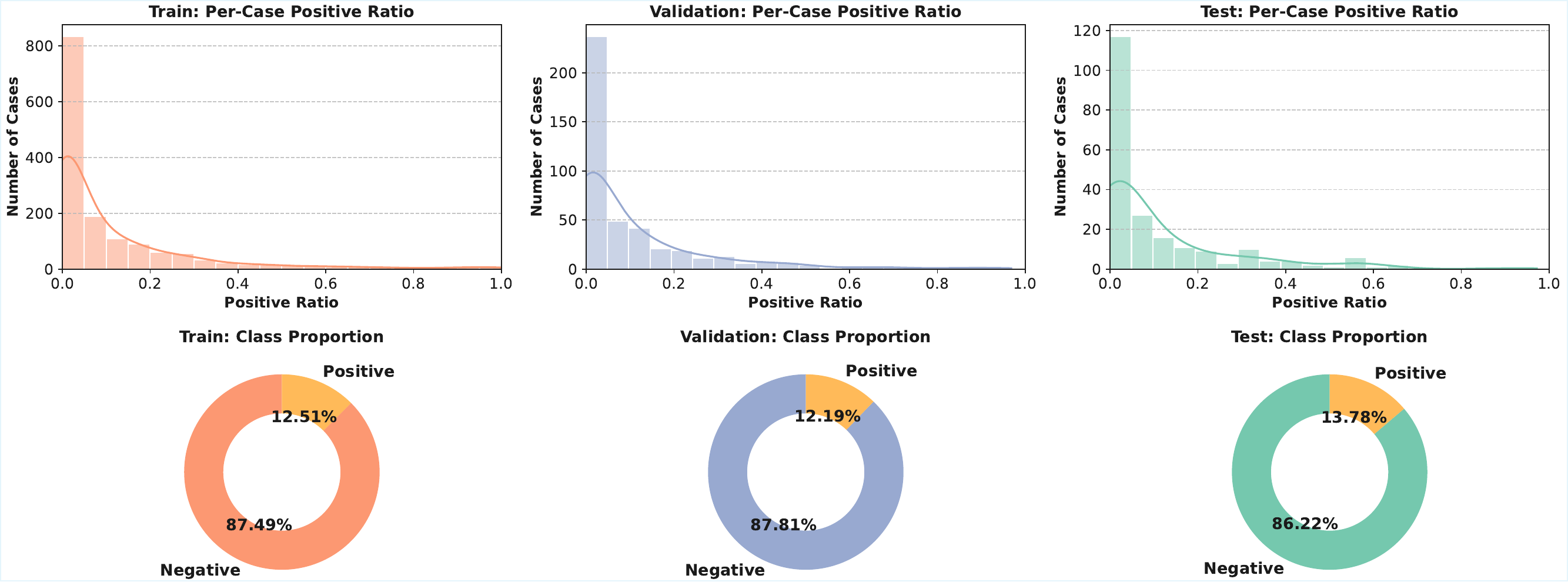}
    \caption{\textbf{Hypotension event distribution in VitalDB subsets.} Top panels: Per-case hypotensive event ratio distributions for training, validation, and test sets. Bottom panels: Corresponding global class proportions for each subset.}
    \label{fig:sub_dataset_statistics}
\end{figure*}

\paragraph{VitalDB dataset statistics.}
In clinical practice, intraoperative hypotension (IOH) events are inherently rare but clinically critical. This characteristic is clearly reflected in the VitalDB dataset~\cite{vitaldb}, where we observe a pronounced rare hypotension events at both individual patient and overall dataset levels. As depicted in Fig.~\ref{fig:dataset_statistics}, where positive samples represent hypotension events, analysis of per-patient hypotension event ratios reveals that a majority of patients experience a very low incidence of such events: specifically, 67.7\% of patients have less than 10\% hypotension events, and 81.1\% have fewer than 20\% hypotension events. This underscores the sparsity of hypotensive episodes within individual surgical timelines. From a global perspective, the overall proportion of hypotension events across the entire VitalDB dataset is only 12.59\%. 
To further understand how these events are distributed across the dataset, we performed a stratified analysis of the training, validation, and test subsets. Fig.~\ref{fig:sub_dataset_statistics} illustrates that all subsets consistently exhibit low ratios of hypotension events—approximately 12.51\%, 12.19\%, and 13.78\% respectively. 
Correspondingly, the distribution of per-case hypotension event ratios within each subset confirms that most cases contain only a small fraction of hypotensive events, reflecting the inherent rarity of hypotension events in real clinical settings.

The scarcity of hypotension events presents intrinsic difficulties for TTA methods, particularly for standard test-time adaptation methods that rely solely on the limited hypotensive patterns present in individual patient histories. 
Our statistical analysis therefore motivates the development of methods that leverage cross-sample information to enrich adaptation contexts and enhance robustness in detecting these rare but clinically significant hypotension episodes. In particular, this study proposes CSA-TTA to address these challenges effectively.

\paragraph{Metric details.}
In this study, we adopt a comprehensive evaluation strategy tailored to the two-stage framework of intraoperative hypotension (IOH) prediction, which involves continuous blood pressure time series forecasting followed by classification of hypotensive events. This dual-task formulation necessitates distinct yet complementary evaluation metrics to thoroughly assess model performance across both regression and classification dimensions.
 
For regression performance, we employ Mean Absolute Error (MAE) and Mean Squared Error (MSE), as detailed in Table~\ref{tab:metrics}. MAE evaluates the average magnitude of errors without considering their direction, offering a direct measure of prediction accuracy that is robust to outliers. In contrast, MSE penalizes larger errors more heavily due to the squaring operation, thereby sensitizing the evaluation to substantial deviations in predicted blood pressure trajectories. 
These metrics provide a balanced view of model fidelity in capturing dynamic, patient-specific blood pressure trends essential for timely clinical intervention and early warning of hypotensive events.

For evaluating the classification of IOH events, we employ Accuracy, Recall, Precision, and F1-score. Among these, Recall is particularly important in the clinical context as it measures the model’s ability to correctly identify hypotensive events (i.e., positive cases), thereby reducing the risks of missed events that could lead to adverse outcomes. Precision complements this by reflecting the reliability of positive predictions, minimizing false alarms and unwarranted interventions. The F1-score balances these two aspects, offering a balanced measure of overall classification effectiveness. Accuracy provides an overall correctness measure but can be less informative in class-imbalanced scenarios typical of IOH. The formal definitions of all metrics used are presented in Table~\ref{tab:metrics}. Together, this set of metrics ensures a holistic and clinically relevant evaluation of predictive performance, capturing both precise blood pressure estimation and reliable detection of hypotensive events.

\begin{table}[t!]
\centering
\setlength{\arrayrulewidth}{1.12pt}
\small
\renewcommand{\arraystretch}{1.7}
\begin{tabular}{p{1.7cm}|p{5.6cm}}
\hline
\textbf{Metrics} & \textbf{Formulation} \\ 
\hline
MAE & \( \displaystyle MAE = \frac{1}{n}\sum_{i=1}^{n} |y_i - \hat{y}_i| \) \\
MSE & \( \displaystyle MSE = \frac{1}{n}\sum_{i=1}^{n} (y_i - \hat{y}_i)^2 \) \\
Accuracy & \( \displaystyle Acc = \frac{TP + TN}{TP + FP + FN + TN} \) \\
Recall & \( \displaystyle Recall = \frac{TP}{TP + FN} \) \\
Precision & \( \displaystyle Precision = \frac{TP}{TP + FP} \) \\
F1-score & \( \displaystyle F_1 = 2 \times \frac{Precision \times Recall}{Precision + Recall} \) \\
\hline
\end{tabular}
\caption{\textbf{Formulation of evaluation metrics.} Where $n$ denotes the total number of samples, $y_i$ and $\hat{y}_i$ represent the ground truth and predicted sequences respectively. For classification metrics: TP (True Positive), TN (True Negative), FP (False Positive), and FN (False Negative) are derived from the confusion matrix.}
\label{tab:metrics}
\end{table}

\paragraph{IOH Event Detection Algorithm.}
Given the predicted blood pressure sequences,  the subsequent task is to determine whether an IOH event occurs within the predicted interval. The IOH event is clinically defined as blood pressure values continuously falling below a certain threshold \( p \) mmHg for a duration of at least \( d \) minutes\cite{ioh_define1,ioh_define2}. A straightforward detection method involves directly examining the predicted sequences for segments strictly meeting this clinical definition. However, due to inherent predictive uncertainty and fluctuations, this binary detection strategy lacks probabilistic representation, potentially resulting in substantial misclassification.

To overcome this,  we design a hybrid method that combines clinical definitions with data-driven probabilistic assessments. Specifically, each predicted blood pressure value is converted into a point-wise risk score representing the probability of MAP falling below a clinical threshold. 
We then apply two triggers: a hard trigger to detect sustained hypotensive periods and a soft trigger that evaluates average risk in sliding windows. 
The hard trigger strictly adheres the clinical definition by identifying sustained periods where blood pressure remains below the threshold within sliding windows, thus detecting persistent hypotensive episodes. 
The soft trigger considers the average risk within the same sliding windows and signals a hypotension risk when this average exceeds a predefined threshold, thereby enhancing robustness in the face of prediction uncertainty and transient fluctuations. These are combined to produce a final probabilistic estimate, balancing clinical reliability with model flexibility. This combined approach balances clinical rigor with the inherent uncertainty of predictive modeling, resulting in more reliable and informed event detection. The detailed procedure is provided in Algorithm~\ref{alg:opt_dual_threshold_ioh}.

\section{\textbf{Appendix B}}
\paragraph{Additional ablation study on Coarse-to-fine cross-sample retrieval.}
To evaluate the contribution of the coarse-to-fine retrieval strategy in CSA-TTA, we perform an ablation study on the TimesFM backbone under zero-shot and fine-tuned settings. There are three retrieval configurations: Random Retrieval (Rand), where samples are randomly selected from both hypotensive and non-hypotensive cross-sample banks; Coarse Retrieval (Coarse), where samples are retrieved based on cluster centroids; and Fine Retrieval (Fine), which refines the retrieval process by selecting the top-K most similar samples from centroids. 
As presented in Table~\ref{tab:coarse_to_fine_ablation}, the full coarse-to-fine retrieval consistently achieves superior performance over the other methods across all prediction horizons. For instance, at the 15-minute horizon in the zero-shot setting, F1-score improves from 60.70 (Rand) and 60.90 (Coarse) to 61.20 with fine-grained retrieval, while MAE and MSE also reach their lowest values. These gains persist across horizons and under fine-tuning, evidencing that hierarchical retrieval not only enhances the relevance of adapted samples but also significantly improves personalized IOH prediction by effectively leveraging temporally and semantically aligned cross-sample information.

\begin{algorithm}[t]
\caption{IOH Event Detection for predicted sequence}
\label{alg:opt_dual_threshold_ioh}
\small
\KwIn{Blood pressure sequence $S = [s_1, s_2, ..., s_n]$, window size $w$, hypotension threshold $\tau_{bp}$, event probability threshold $\tau_{event}$, sigmoid parameters $\beta$, $\delta$, soft threshold $\tau_{soft}$, aggregation parameters $\lambda_{hard}$, $\lambda_{soft}$}

\KwOut{Hypotension event probability $P_{final}$}

\BlankLine

\tcp{Initialize risk containers}
$\mathcal{W}_{hard} \leftarrow \emptyset$, $\mathcal{W}_{soft} \leftarrow \emptyset$\;

\BlankLine

\tcp{Sliding window processing}
\For{$i = 1$ \KwTo $n - w + 1$}{
    $W \leftarrow [s_i, s_{i+1}, ..., s_{i+w-1}]$
    
    \tcp{Hard trigger condition}
    \If{$\forall s_j \in W: s_j < \tau_{bp}$}{
        $R_h \leftarrow \frac{1}{w} \sum \sigma(s_j, \beta, \tau_{bp}, \delta)$ \tcp*{$s_j \in W$}
        $\mathcal{W}_{hard} \leftarrow \mathcal{W}_{hard} \cup \{R_h\}$
    }
    
    \tcp{Soft trigger processing}
    $R_s \leftarrow \frac{1}{w} \sum \sigma(s_j, \beta, \tau_{bp}, 0)$ \tcp*{$s_j \in W$}
    \If{$R_s \geq \tau_{soft}$}{
        $\mathcal{W}_{soft} \leftarrow \mathcal{W}_{soft} \cup \{R_s\}$
    }
}

\BlankLine

\tcp{Aggregate risks}
$P_{hard} \leftarrow \Phi(\mathcal{W}_{hard}, \lambda_{hard})$\;
$P_{soft} \leftarrow \Phi(\mathcal{W}_{soft}, \lambda_{soft})$\;

\BlankLine

\tcp{Determine final probability}
\uIf{$P_{hard} \geq \tau_{event}$}{
    $P_{final} \leftarrow P_{hard}$\;
}
\Else{
    $P_{final} \leftarrow P_{soft}$\;
}

\Return{$P_{final}$}\;

\BlankLine

\tcp{Auxiliary functions}
$\sigma(x, \beta, \tau, \delta) = \frac{1}{1 + \exp(\beta \cdot (x - \tau - \delta))}$ \\
$\Phi(\mathcal{R}, \lambda) = 1 - \exp(-\lambda \sum_{r \in \mathcal{R}} r)$ \\

\end{algorithm}

\paragraph{Additional ablation study on Top-K of Coarse-to-fine retrieval.}
We further investigate the impact of the Top-K parameter in the fine-grained retrieval stage of CSA-TTA by conducting an ablation study on the TimesFM backbone using the VitalDB dataset, under both zero-shot and fine-tuned settings. 
Table~\ref{tab:top_k_performance} compares the model’s performance with Top-K values set to 1, 2, and 3 across three prediction horizons (5, 10, and 15 minutes). 
The results show that increasing Top-K moderately leads to improved or stable performance in key metrics such as F1-score, Recall, and regression errors (MAE and MSE). For example, at the 5-minute horizon in the zero-shot setting, F1-score increases from 69.40 with Top-K=1 to 69.80 with Top-K=3, while MAE decreases from 5.08 to 5.07. Similarly, under the fine-tuned setting, Top-K=3 achieves the best F1-score and the lowest MAE and MSE across most horizons. 
These results suggest that retrieving a moderate number of semantically similar samples improves adaptation by offering diverse yet relevant temporal patterns. Too small a Top-K limits context diversity, while too large may introduce noisy, less relevant samples. We set Top-K to 3 balances relevance and diversity, enhancing adaptation robustness and accuracy.


\begin{table*}[t]
\centering
\begin{tabular}{l|lll|rrrrrr|rrrrrr}
\toprule
\multicolumn{4}{c|}{\textbf{Config}} & \multicolumn{6}{c|}{\textbf{Zero-Shot Setting}} & \multicolumn{6}{c}{\textbf{Fine-tuned Setting}}\\ 
\cmidrule(lr){1-4} \cmidrule(lr){5-10} \cmidrule(lr){11-16}
\textbf{Horizon} & \textbf{Rand} & \textbf{Coarse} & \textbf{Fine} & \textbf{F1$\uparrow$} & \textbf{Rec$\uparrow$} & \textbf{Prec$\uparrow$} & \textbf{Acc$\uparrow$} & \textbf{MAE$\downarrow$} & \textbf{MSE$\downarrow$} & \textbf{F1$\uparrow$} & \textbf{Rec$\uparrow$} & \textbf{Prec$\uparrow$} & \textbf{Acc$\uparrow$} & \textbf{MAE$\downarrow$} & \textbf{MSE$\downarrow$} \\
\midrule
\multirow{3}{*}{5min} 
 & \textcolor{checkmark}{\ding{51}} & \textcolor{crossmark}{\ding{55}} & \textcolor{crossmark}{\ding{55}} & 69.30 & \underline{66.90} & 72.00 & 91.80 & 5.08 & 60.15 & \underline{70.40} & \textbf{71.60} & 69.20 & 91.70 & 4.78 & 54.40 \\ 
 & \textcolor{crossmark}{\ding{55}} & \textcolor{checkmark}{\ding{51}} & \textcolor{crossmark}{\ding{55}} & \underline{69.50} & 66.80 & \underline{72.40} & \underline{91.90} & \textbf{5.05} & \underline{59.38} & 70.20 & 71.30 & 69.20 & 91.70 & \underline{4.76} & \underline{54.28} \\ 
 & \cellcolor{lightgray}\textcolor{crossmark}{\ding{55}} & \cellcolor{lightgray}\textcolor{checkmark}{\ding{51}} & \cellcolor{lightgray}\textcolor{checkmark}{\ding{51}} & \cellcolor{lightgray}\textbf{69.80} & \cellcolor{lightgray}\textbf{67.40} & \cellcolor{lightgray}\textbf{72.50} & \cellcolor{lightgray}\textbf{92.00} & \cellcolor{lightgray}\underline{5.07} & \cellcolor{lightgray}\textbf{59.32} & \cellcolor{lightgray}\textbf{70.60} & \cellcolor{lightgray}\textbf{71.60} & \cellcolor{lightgray}\textbf{69.70} & \cellcolor{lightgray}\textbf{91.80} & \cellcolor{lightgray}\textbf{4.77} & \cellcolor{lightgray}\textbf{53.17} \\
\hline
\multirow{3}{*}{10min} 
 & \textcolor{checkmark}{\ding{51}} & \textcolor{crossmark}{\ding{55}} & \textcolor{crossmark}{\ding{55}} & 63.60 & \underline{57.80} & \underline{70.70} & 87.30 & 6.59 & 93.03 & 64.50 & 59.70 & \underline{70.00} & 87.40 & 6.17 & 80.97 \\
 & \textcolor{crossmark}{\ding{55}} & \textcolor{checkmark}{\ding{51}} & \textcolor{crossmark}{\ding{55}} & \textbf{63.70} & 57.70 & \textbf{71.00} & \underline{87.40} & \underline{6.55} & \underline{91.74} & \textbf{64.70 }& \underline{59.80} & \textbf{70.40 }& \textbf{87.50} & 6.17 & \underline{80.67} \\
 & \cellcolor{lightgray}\textcolor{crossmark}{\ding{55}} & \cellcolor{lightgray}\textcolor{checkmark}{\ding{51}} & \cellcolor{lightgray}\textcolor{checkmark}{\ding{51}} & \cellcolor{lightgray}\textbf{63.70} & \cellcolor{lightgray}\textbf{58.10} & \cellcolor{lightgray}70.50 & \cellcolor{lightgray}\textbf{87.60} & \cellcolor{lightgray}\textbf{6.54} & \cellcolor{lightgray}\textbf{91.01} & \cellcolor{lightgray}\textbf{64.70} & \cellcolor{lightgray}\textbf{60.80} & \cellcolor{lightgray}69.20 & \cellcolor{lightgray}\textbf{87.50} & \cellcolor{lightgray}\textbf{6.05} & \cellcolor{lightgray}\textbf{77.60} \\
 \hline
\multirow{3}{*}{15min} 
 & \textcolor{checkmark}{\ding{51}} & \textcolor{crossmark}{\ding{55}} & \textcolor{crossmark}{\ding{55}} & 60.70 & 52.10 & 72.50 & 84.10 & 7.29 & 108.24 & 62.80 & \textbf{54.80 }& \underline{73.60} & 84.70 & 6.69 & 89.50 \\ 
 & \textcolor{crossmark}{\ding{55}} & \textcolor{checkmark}{\ding{51}} & \textcolor{crossmark}{\ding{55}} & \underline{60.90} & \underline{52.20} & \underline{73.10} & \textbf{84.20} & \underline{7.25} & \underline{106.55} & \textbf{62.90} & 54.60 & 74.20 & \textbf{84.80} & 6.69 & \underline{89.41} \\
 & \cellcolor{lightgray}\textcolor{crossmark}{\ding{55}} & \cellcolor{lightgray}\textcolor{checkmark}{\ding{51}} & \cellcolor{lightgray}\textcolor{checkmark}{\ding{51}} & \cellcolor{lightgray}\textbf{61.20} & \cellcolor{lightgray}\textbf{52.30} & \cellcolor{lightgray}\textbf{73.60} & \cellcolor{lightgray}\textbf{84.20} & \cellcolor{lightgray}\textbf{7.23} & \cellcolor{lightgray}\textbf{105.52} & \cellcolor{lightgray}\textbf{62.90} & \cellcolor{lightgray}54.60 & \cellcolor{lightgray}\textbf{74.00} & \cellcolor{lightgray}\textbf{84.80} & \cellcolor{lightgray}\textbf{6.64} & \cellcolor{lightgray}\textbf{88.02} \\
\bottomrule
\end{tabular}
\caption{\textbf{Ablation study on coarse-to-fine cross-sample retrieval.} Performance of TimesFM under zero-shot and fine-tuned settings using different retrieval methods: Rand (random retrieval from hypotensive and non-hypotensive cross-sample bank), Coarse (random retrieval from hypotensive and non-hypotensive centroid), and Fine (Fine-grained retrieval from hypotensive and non-hypotensive centroid). The best performance is highlighted in \textbf{bold}, and the second-best is \underline{underlined}.}
\label{tab:coarse_to_fine_ablation}
\end{table*}

\begin{table*}[t]
\centering
\begin{tabular}{l|l|rrrrrr|rrrrrr}
\toprule
\multicolumn{2}{c|}{\textbf{Config}} & \multicolumn{6}{c|}{\textbf{Zero-Shot Setting}} & \multicolumn{6}{c}{\textbf{Fine-tuned Setting}}\\ 
\cmidrule(lr){1-2} \cmidrule(lr){3-8} \cmidrule(lr){9-14}
\textbf{Horizon} & \textbf{Top-K} & \textbf{F1$\uparrow$} & \textbf{Rec$\uparrow$} & \textbf{Prec$\uparrow$} & \textbf{Acc$\uparrow$} & \textbf{MAE$\downarrow$} & \textbf{MSE$\downarrow$} & \textbf{F1$\uparrow$} & \textbf{Rec$\uparrow$} & \textbf{Prec$\uparrow$} & \textbf{F1$\uparrow$} & \textbf{MAE$\downarrow$} & \textbf{MSE$\downarrow$} \\
\midrule
\multirow{3}{*}{5min} 
 & 1 & 69.40 & 66.80 & 72.20 & \textbf{92.00} & 5.08 & 59.93 & 70.30 & \underline{71.70} & 68.90 & 91.60 & 4.79 & 54.80 \\
 & 2 & \underline{69.60} & \underline{67.00} & \textbf{72.60} & 91.90 & \textbf{5.07} & \underline{59.89} & \underline{70.40} & 71.50 & \underline{69.30} & \underline{91.70} & \textbf{4.77} & \underline{54.31} \\
 & \cellcolor{lightgray}3 & \cellcolor{lightgray}\textbf{69.80} & \cellcolor{lightgray}\textbf{67.40} & \cellcolor{lightgray}\underline{72.50} & \cellcolor{lightgray}\textbf{92.00} & \cellcolor{lightgray}\textbf{5.07} & \cellcolor{lightgray}\textbf{59.32} & \cellcolor{lightgray}\textbf{70.60} & \cellcolor{lightgray}\textbf{71.60} & \cellcolor{lightgray}\textbf{69.70} & \cellcolor{lightgray}\textbf{91.80} & \cellcolor{lightgray}\textbf{4.77} & \cellcolor{lightgray}\textbf{53.17} \\
\hline
\multirow{3}{*}{10min} 
 & 1 & 63.50 & 57.60 & \underline{70.70} & 87.30 & 6.59 & 93.03 &\textbf{64.80} & \underline{60.20} & \underline{70.10} & 87.40 & 6.17 & 81.01 \\
 & 2 & \underline{63.60} & 57.60 & \textbf{70.90} & 87.30 & \underline{6.57} & \underline{91.96} & 64.60 & 59.70 & \textbf{70.30} & 87.40 & \underline{6.15} & \underline{80.64} \\
 & \cellcolor{lightgray}3 & \cellcolor{lightgray}\textbf{63.70} & \cellcolor{lightgray}\textbf{58.10} & \cellcolor{lightgray}70.50 & \cellcolor{lightgray}\textbf{87.60} & \cellcolor{lightgray}\textbf{6.54} & \cellcolor{lightgray}\textbf{91.01} & \cellcolor{lightgray}\underline{64.70} & \cellcolor{lightgray}\textbf{60.80} & \cellcolor{lightgray}69.20 & \cellcolor{lightgray}\textbf{87.50} & \cellcolor{lightgray}\textbf{6.05} & \cellcolor{lightgray}\textbf{77.60} \\
\hline
\multirow{3}{*}{15min} 
 & 1 & 60.80 & 52.10 & \underline{73.00} & \textbf{84.20} & 7.29 & 107.92 & \underline{62.80} & \textbf{54.60} & 73.90 & 84.70 & 6.69 & 89.47 \\
 & 2 & \underline{61.00} & \textbf{52.50} & 72.90 & \textbf{84.20} & \underline{7.25} & \underline{107.01} & 62.70 & 54.40 & 73.90 & 84.70 & 6.69 & \underline{89.31} \\
 & \cellcolor{lightgray}3 & \cellcolor{lightgray}\textbf{61.20} & \cellcolor{lightgray}\underline{52.30} & \cellcolor{lightgray}\textbf{73.60} & \cellcolor{lightgray}\textbf{84.20} & \cellcolor{lightgray}\textbf{7.23} & \cellcolor{lightgray}\textbf{105.52} & \cellcolor{lightgray}\textbf{62.90} & \cellcolor{lightgray}\textbf{54.60} & \cellcolor{lightgray}\textbf{74.00} & \cellcolor{lightgray}\textbf{84.80} & \cellcolor{lightgray}\textbf{6.64} & \cellcolor{lightgray}\textbf{88.02} \\
\bottomrule
\end{tabular}
\caption{\textbf{Ablation study on Top-K of coarse-to-fine cross-sample retrieval.} Performance of the TimesFM backbone under zero-shot and fine-tuned settings using different retrieval methods: Top-K 1, 2, and 3.}
\label{tab:top_k_performance}
\end{table*}


\paragraph{Historical Window.} Moderately increasing the historical window for TTA provides richer temporal context and leads to gradual performance improvements.
As shown in Table~\ref{tab:window_count_ablation}, within the TimesFM~+~CSA-TTA setting, increasing the window count from 1 to 3 raises zero-shot Recall from 58.67\% to 58.81\% and F1 from 64.27\% to 64.47\%, while MSE decreases from 89.88 to 86.70. A similar pattern holds in the fine-tuning setting. However, as the window length grows, performance gains taper off while inference costs rise. To balance accuracy and efficiency, we cap the window count at 3.

\paragraph{Statistical significance.} In this study, model performance was evaluated using a single random data split. As shown in Table~\ref{tab:satistical_analysis}, we provide a table reporting the mean, standard deviation, and 95\% confidence intervals of key metrics for the Zero-shot setting with the Units on the VitalDB (2S) dataset, based on 5 runs of the random data split. Specifically, CSA-TTA improves Recall from 44.83\% to 52.50\%, with a standard deviation of 0.038, indicating a robust and significant improvement.

\begin{table}[t]
  \centering
  \small
  \setlength{\tabcolsep}{1.2mm}
  \renewcommand{\arraystretch}{1.1}
  \begin{tabular}{ll|rrrrrr}
    \toprule
    \multicolumn{2}{c|}{\textbf{Setting / Window}} &
      \multicolumn{6}{c}{\textbf{VitalDB Dataset}} \\
    \cmidrule(lr){3-8}
    && \textbf{F1$\uparrow$} & \textbf{Recall$\uparrow$} & \textbf{Prec$\uparrow$} & \textbf{Acc$\uparrow$} & \textbf{MAE$\downarrow$} & \textbf{MSE$\downarrow$} \\
    \midrule
    \multirow{3}{*}{Zero-shot}
      & 1 & 64.27 & 58.67 & 71.05 & 87.67 & 6.40 & 89.88 \\
      & 2 & 64.40 & 58.69 & \textbf{71.77} & 87.77 & 6.37 & 88.67 \\
      & \cellcolor{lightgray}3 & \cellcolor{lightgray}\textbf{64.47} & \cellcolor{lightgray}\textbf{58.81} & \cellcolor{lightgray}71.34 & \cellcolor{lightgray}\textbf{87.80} & \cellcolor{lightgray}\textbf{6.33} & \cellcolor{lightgray}\textbf{86.70} \\
    \midrule
    \multirow{3}{*}{Fine-tuning}
      & 1 & 65.87 & 62.17 & \textbf{70.68} & 87.87 & 5.90 & 75.25 \\
      & 2 & 65.83 & 62.13 & \textbf{70.68} & \textbf{87.90} & 5.88 & 75.09 \\
      & \cellcolor{lightgray}3 & \cellcolor{lightgray}\textbf{65.93} & \cellcolor{lightgray}\textbf{62.21} & \cellcolor{lightgray}70.12 & \cellcolor{lightgray}\textbf{87.90} & \cellcolor{lightgray}\textbf{5.87} & \cellcolor{lightgray}\textbf{74.79} \\
    \bottomrule
  \end{tabular}
  \caption{\textbf{Performance comparison of different historical windows.}  Performance of TimesFM~+~CSA-TTA with 1–3 historical windows under zero-shot and fine-tuning settings, evaluated on VitalDB and averaged over three prediction horizons (5/10/15 min).}
  \label{tab:window_count_ablation}
\end{table}

\begin{table}[h]
\centering
\begin{tabular}{@{}lcc@{}}
\toprule
\textbf{Method} & \textbf{Recall (\%)} & \textbf{MSE} \\
\midrule
Units & $44.83 \pm 0.0$ & $98.88 \pm 0.0$ \\
 & $[44.83, 44.83]$ & $[98.88, 98.88]$ \\
\cmidrule{1-3}
Units + CSA-TTA & $\mathbf{52.50 \pm 0.038}$ & $\mathbf{96.63 \pm 0.188}$ \\
 & $[52.46, 52.55]$ & $[96.28, 96.74]$ \\
\bottomrule
\end{tabular}
\caption{\textbf{Statistical significance analysis of zero-shot performance on VitalDB dataset.} Values represent mean ± standard deviation [95\% Confidence Intervals] over 5 random splits.}
\label{tab:satistical_analysis}
\end{table}

\paragraph{Detail Results.}
In this section, we provide detailed performance results for all the backbone models evaluated in this paper. Tables~\ref{tab:timesfm_full_results} and \ref{tab:units_full_results} report the comprehensive metrics of the TimesFM and Units backbones, respectively. Each table presents results under both zero-shot and fine-tuned adaptation settings across three prediction horizons (5, 10, and 15 minutes) using a consistent 15-minute lookback window on the VitalDB dataset sampled at 2-second and 30-second intervals.

As shown in Tables~\ref{tab:timesfm_full_results} and \ref{tab:units_full_results}, we observe that CSA-TTA consistently improves key classification metrics, including F1-score and Recall, as well as regression errors (MAE and MSE) compared to baseline methods across all horizons and data granularities. For example, on the TimesFM backbone in the zero-shot setting at the 5-minute horizon with 2-second sampling, CSA-TTA achieves an F1-score of 68.10 compared to 67.70 from the baseline, alongside a reduction in MAE from 4.94 to 4.82. Similar performance gains are evident on longer horizons and under fine-tuning, reflecting the robustness of CSA-TTA’s personalized adaptation. Analogous trends are demonstrated in the Units backbone results, where improvements in Recall exceed 5.8\% at 15min horizons (Zero-shot setting), accompanied by meaningful decreases in regression errors.

Additionally, Table~\ref{tab:inhospital_results} presents a detailed evaluation of the TimesFM and Units backbones on the in-hospital test set under the zero-shot adaptation setting. This table demonstrates the effectiveness of CSA-TTA in real-world clinical data, showing consistent gains in both classification and regression metrics across all prediction horizons. Notably, CSA-TTA improves Recall by up to 10.5\% and reduces in MAE from 6.13 to 6.01 at the 5-minute horizon with the Units backbone, further demonstrating its practical utility and generalizability.

Overall, these detailed performance analyses confirm CSA-TTA’s robust capability to improve personalized intraoperative hypotension prediction across different backbone architectures, sampling rates, and clinical datasets.

\begin{figure*}[!t]
    \centering
    \begin{subfigure}{0.48\linewidth}
        \includegraphics[width=\linewidth, height=5cm]{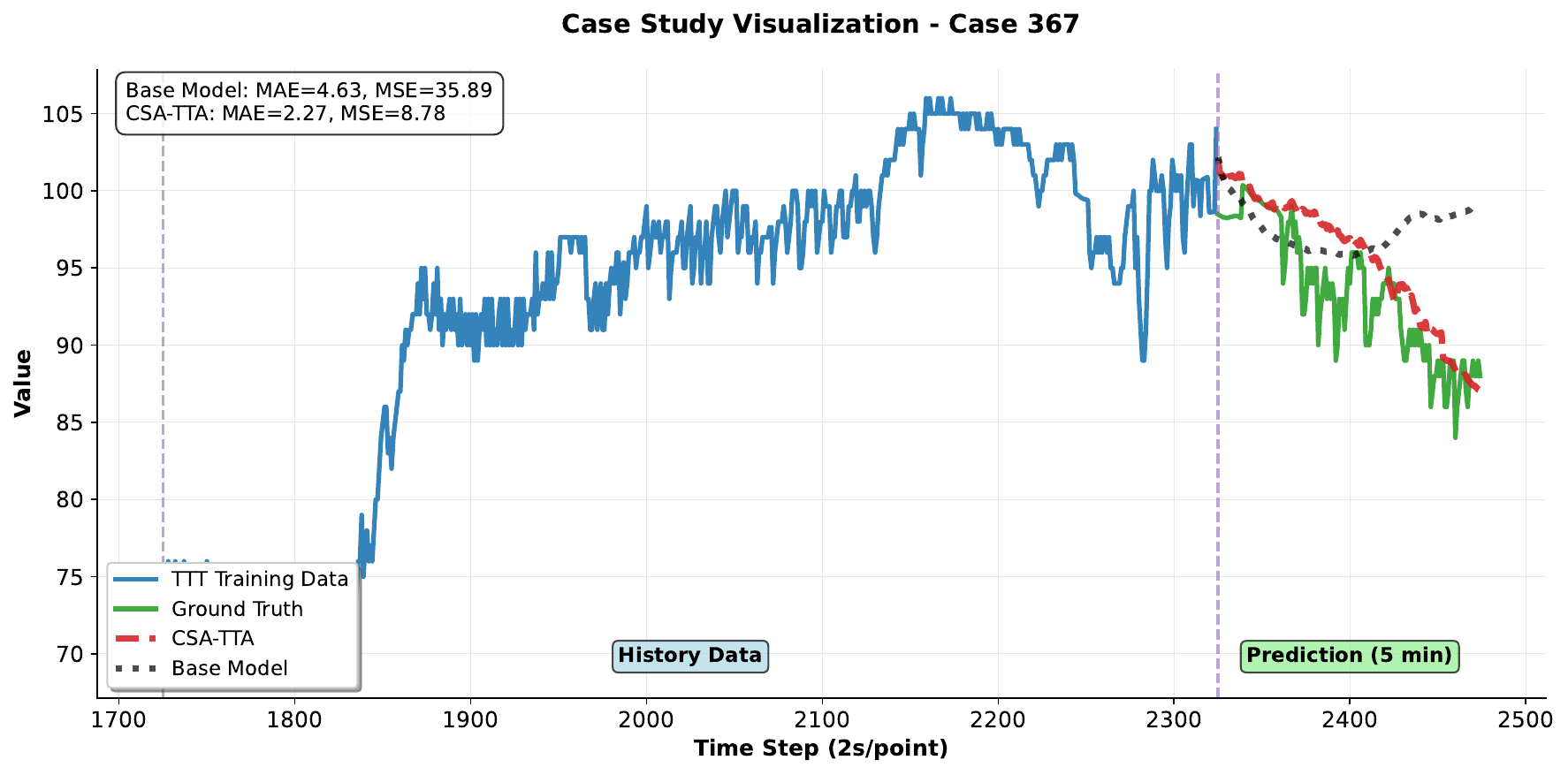}
        \caption{}\label{fig:case_study5_1}
    \end{subfigure}\hfill
    \begin{subfigure}{0.48\linewidth}
        \includegraphics[width=\linewidth, height=5cm]{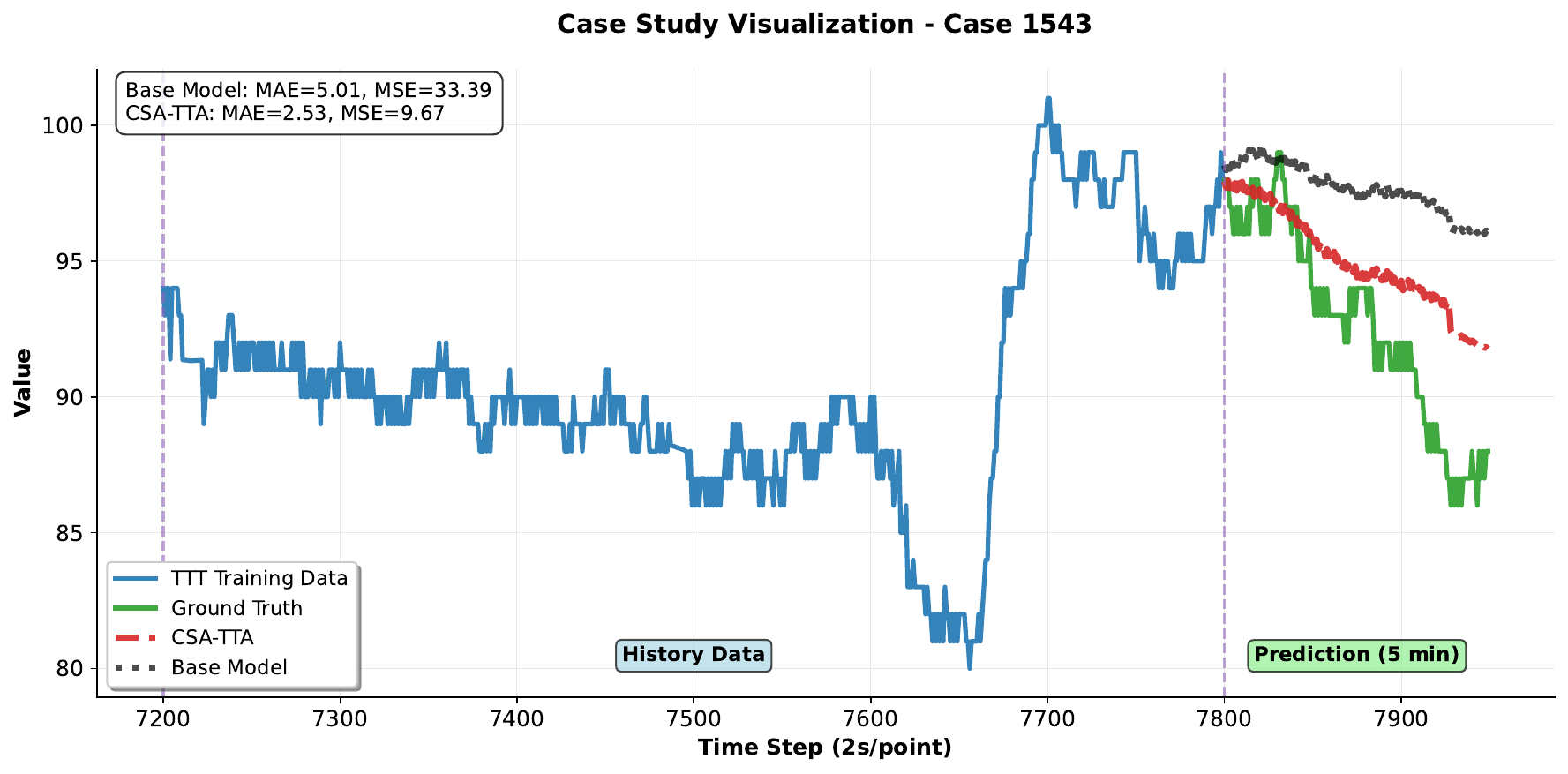}
        \caption{}\label{fig:case_study5_2}
    \end{subfigure}

    \vspace{0.5em} 

    \begin{subfigure}{0.48\linewidth}
        \includegraphics[width=\linewidth, height=5cm]{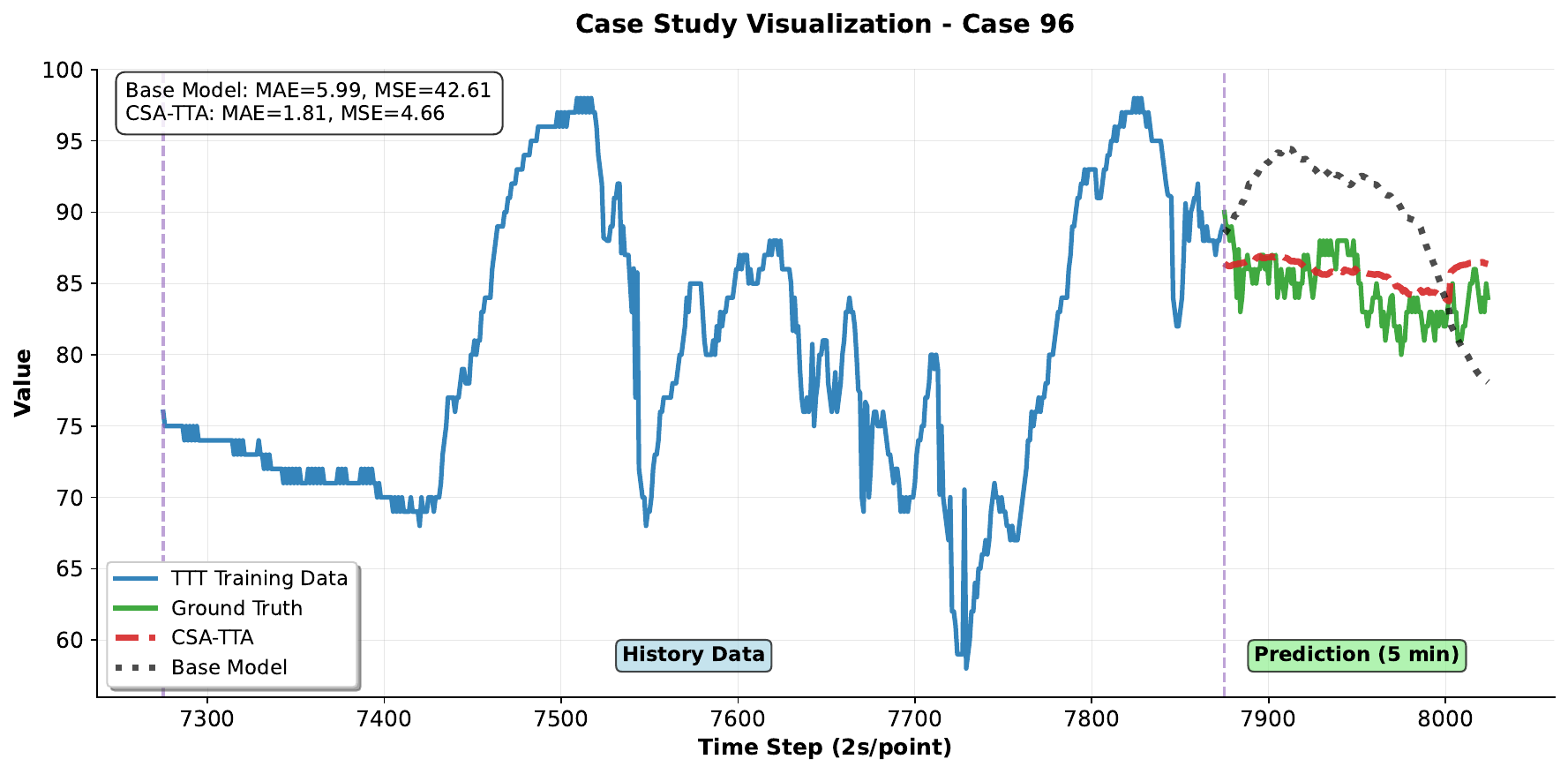}
        \caption{}\label{fig:case_study5_3}
    \end{subfigure}\hfill
    \begin{subfigure}{0.48\linewidth}
        \includegraphics[width=\linewidth, height=5cm]{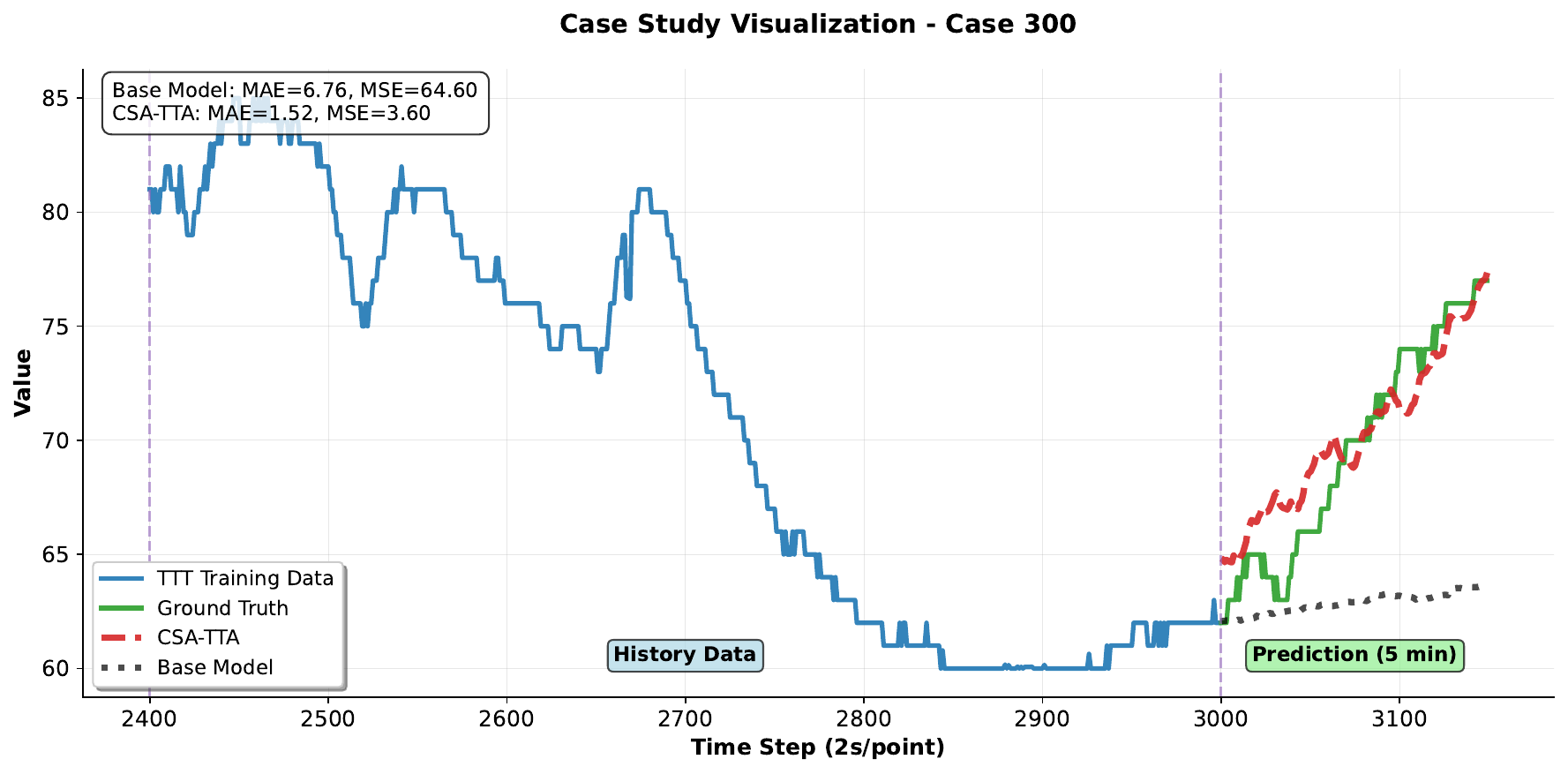}
        \caption{}\label{fig:case_study10_1}
    \end{subfigure}

    \vspace{0.5em}

    \begin{subfigure}{0.48\linewidth}
        \includegraphics[width=\linewidth, height=5cm]{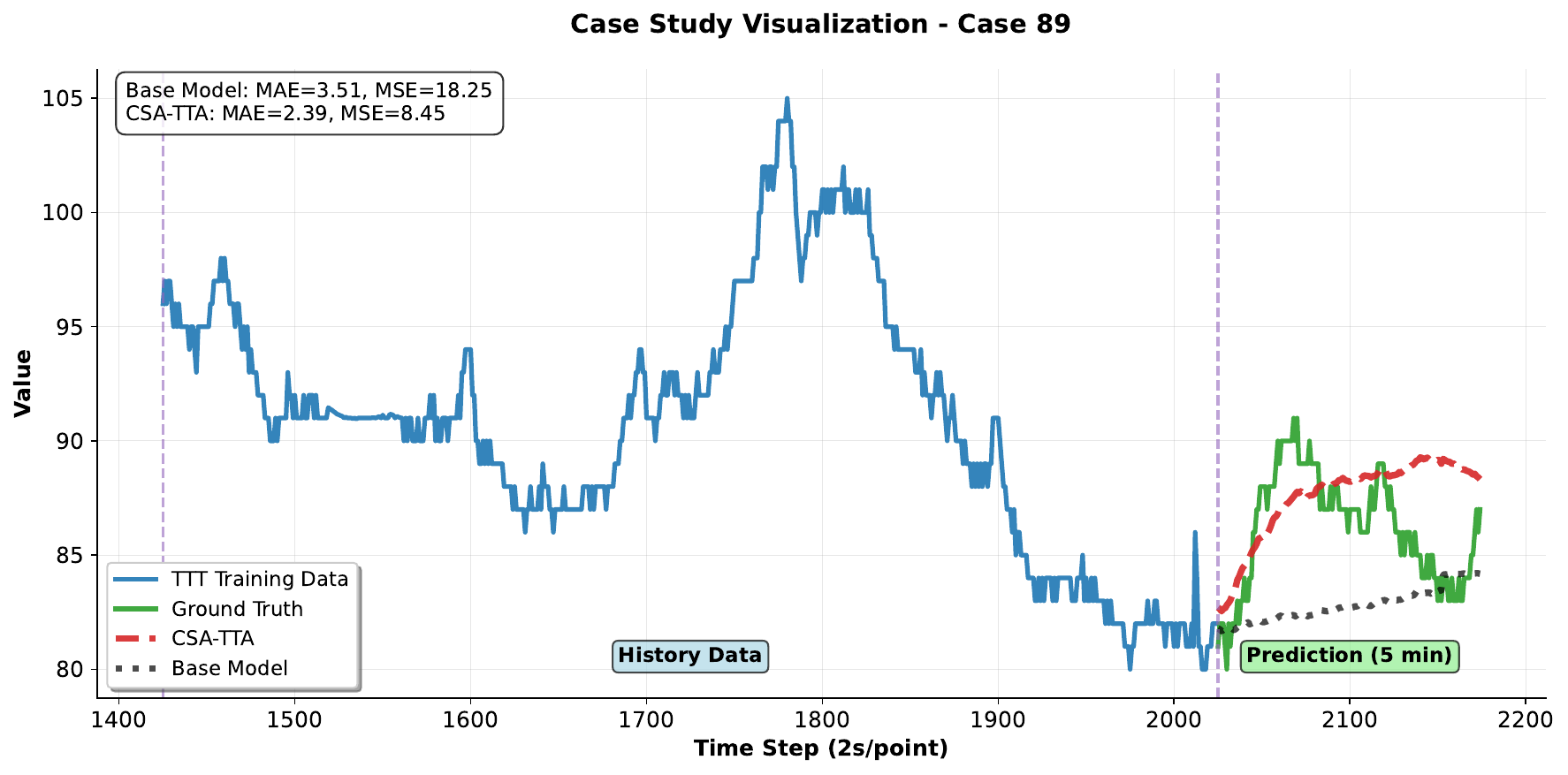}
        \caption{}\label{fig:case_study10_2}
    \end{subfigure}\hfill
    \begin{subfigure}{0.48\linewidth}
        \includegraphics[width=\linewidth, height=5cm]{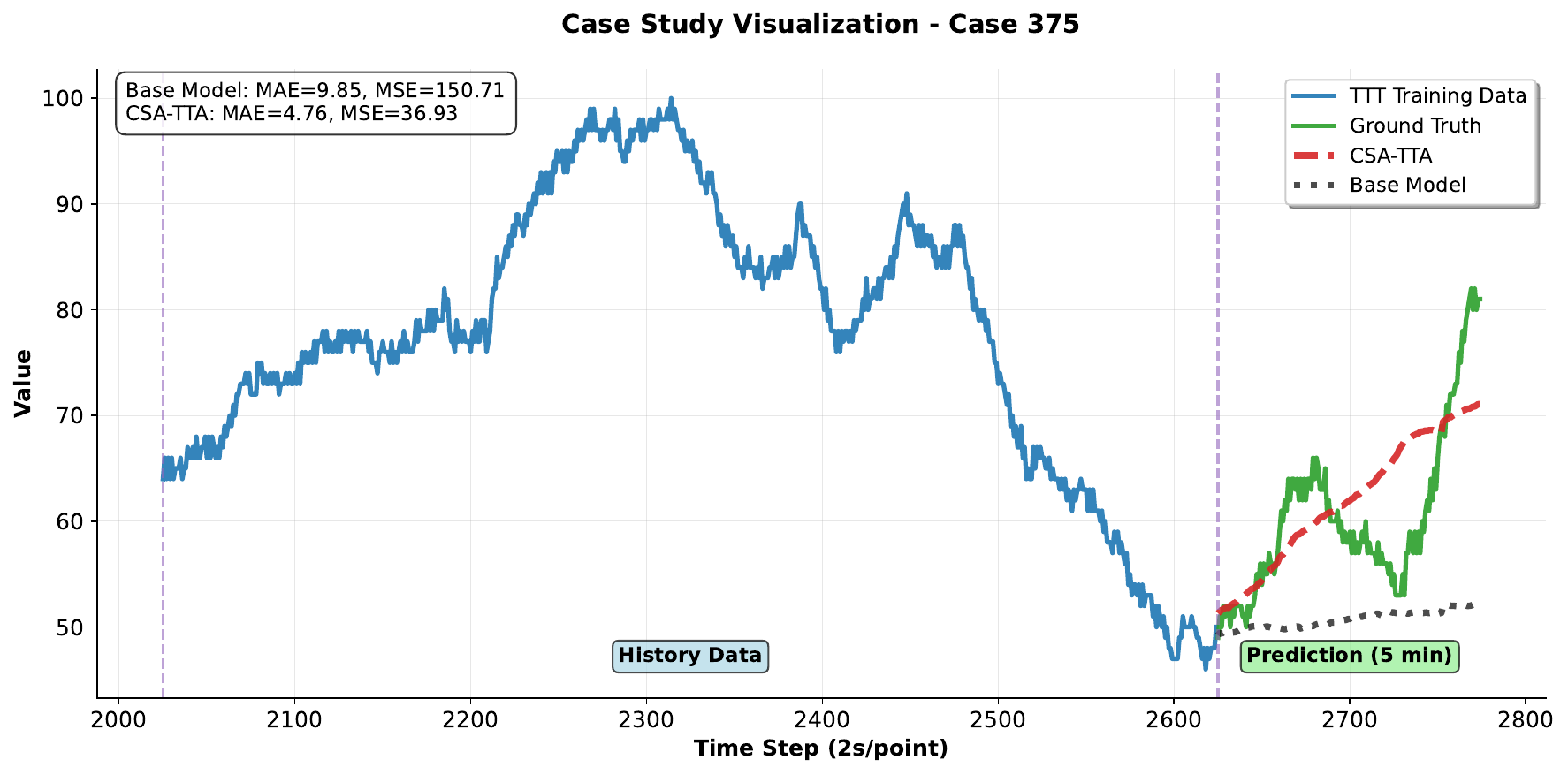}
        \caption{}\label{fig:case_study10_3}
    \end{subfigure}

    \vspace{0.5em}

    \begin{subfigure}{0.48\linewidth}
        \includegraphics[width=\linewidth, height=5cm]{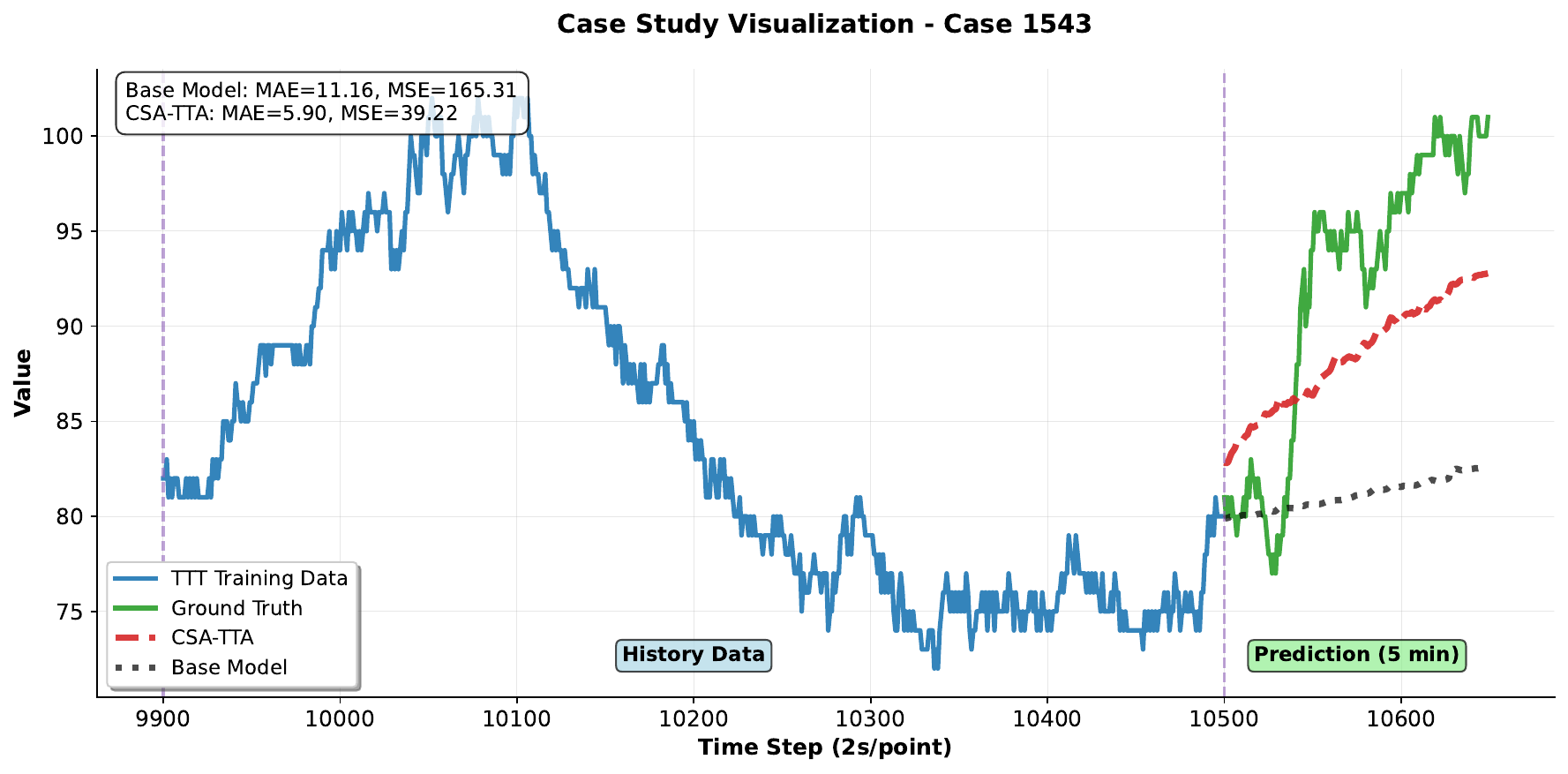}
        \caption{}\label{fig:case_study10_4}
    \end{subfigure}\hfill
    \begin{subfigure}{0.48\linewidth}
        \includegraphics[width=\linewidth, height=5cm]{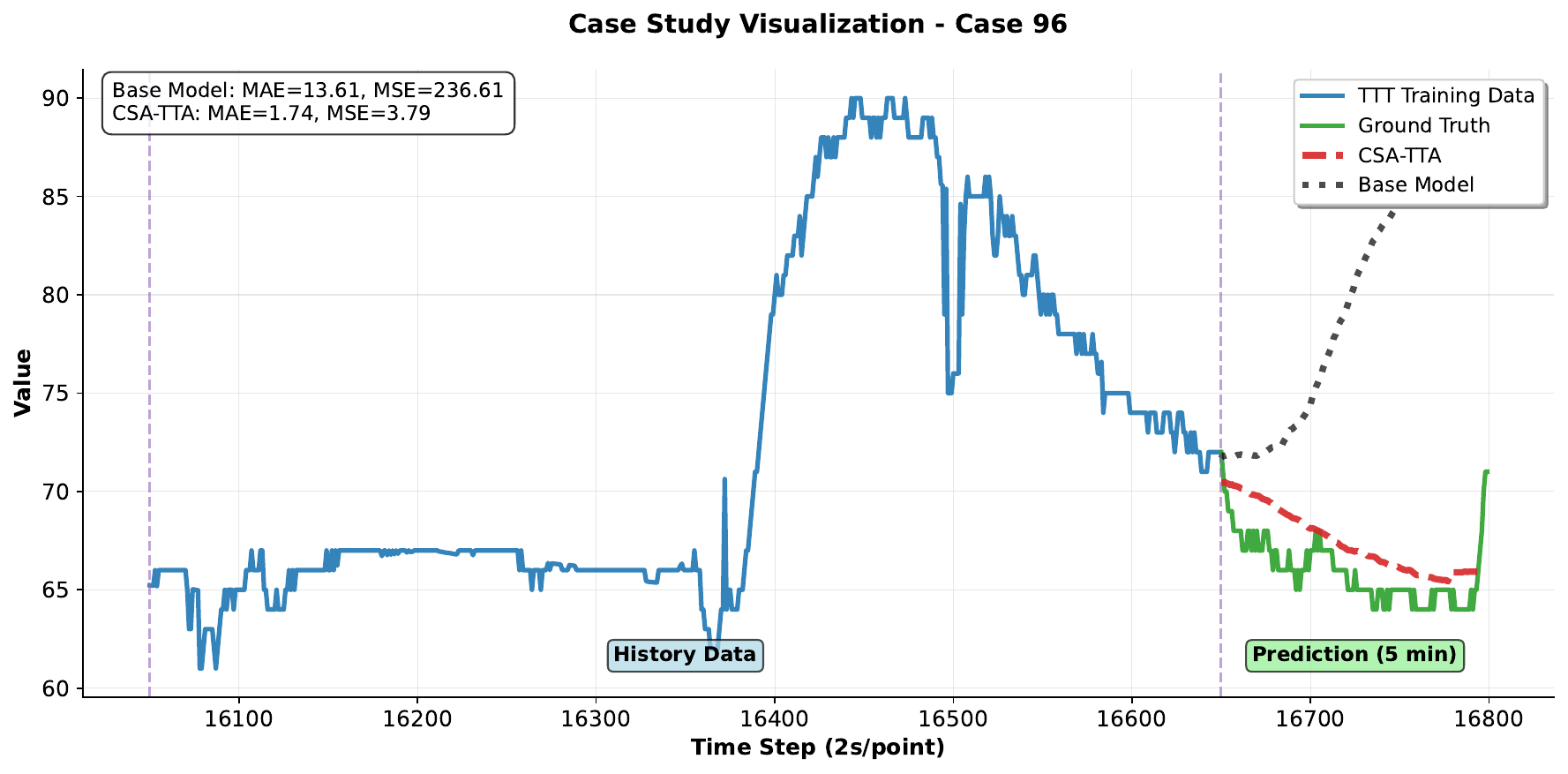}
        \caption{}\label{fig:case_study10_5}
    \end{subfigure}

    \caption{\textbf{Case study visualizations for the 5-minute prediction horizon.} Each subfigure shows an example case from the VitalDB dataset.}
    \label{fig:case_study_5}
\end{figure*}

\begin{figure*}[!t]
    \centering
    \begin{subfigure}{0.48\linewidth}
        \includegraphics[width=\linewidth, height=5cm]{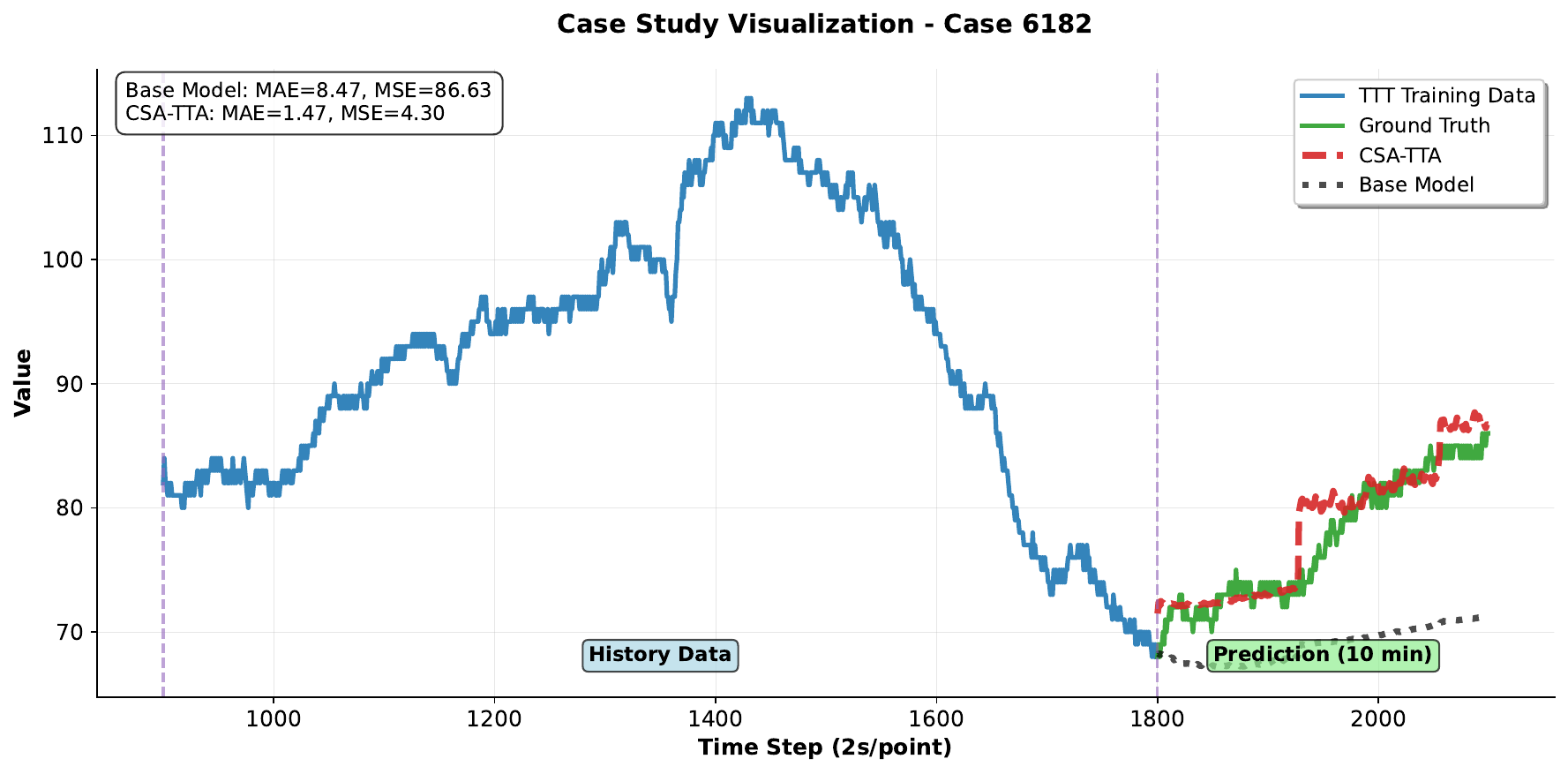}
        \caption{}\label{fig:case_study10_1}
    \end{subfigure}\hfill
    \begin{subfigure}{0.48\linewidth}
        \includegraphics[width=\linewidth, height=5cm]{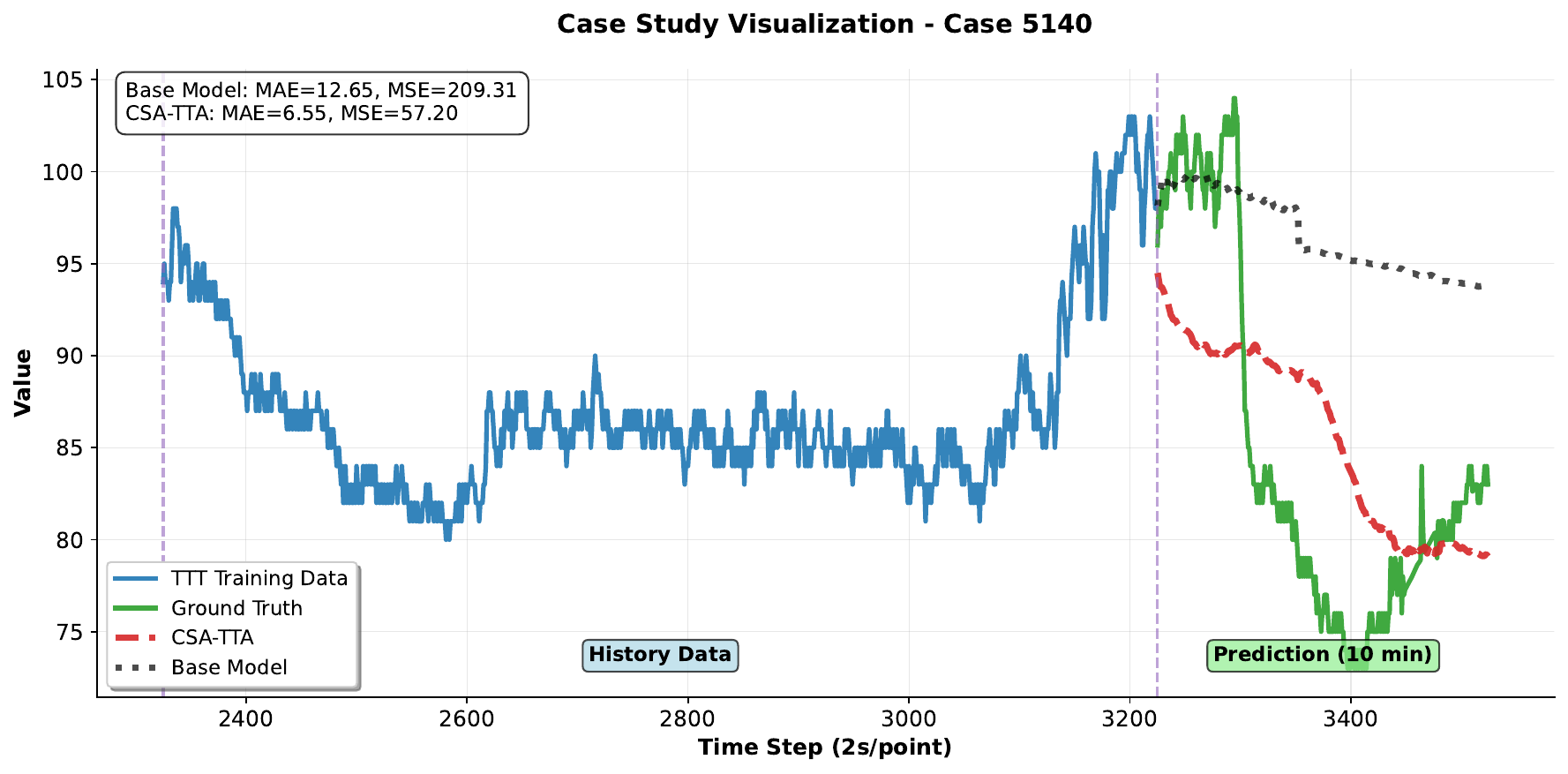}
        \caption{}\label{fig:case_study10_2}
    \end{subfigure}

    \vspace{0.5em}

    \begin{subfigure}{0.48\linewidth}
        \includegraphics[width=\linewidth, height=5cm]{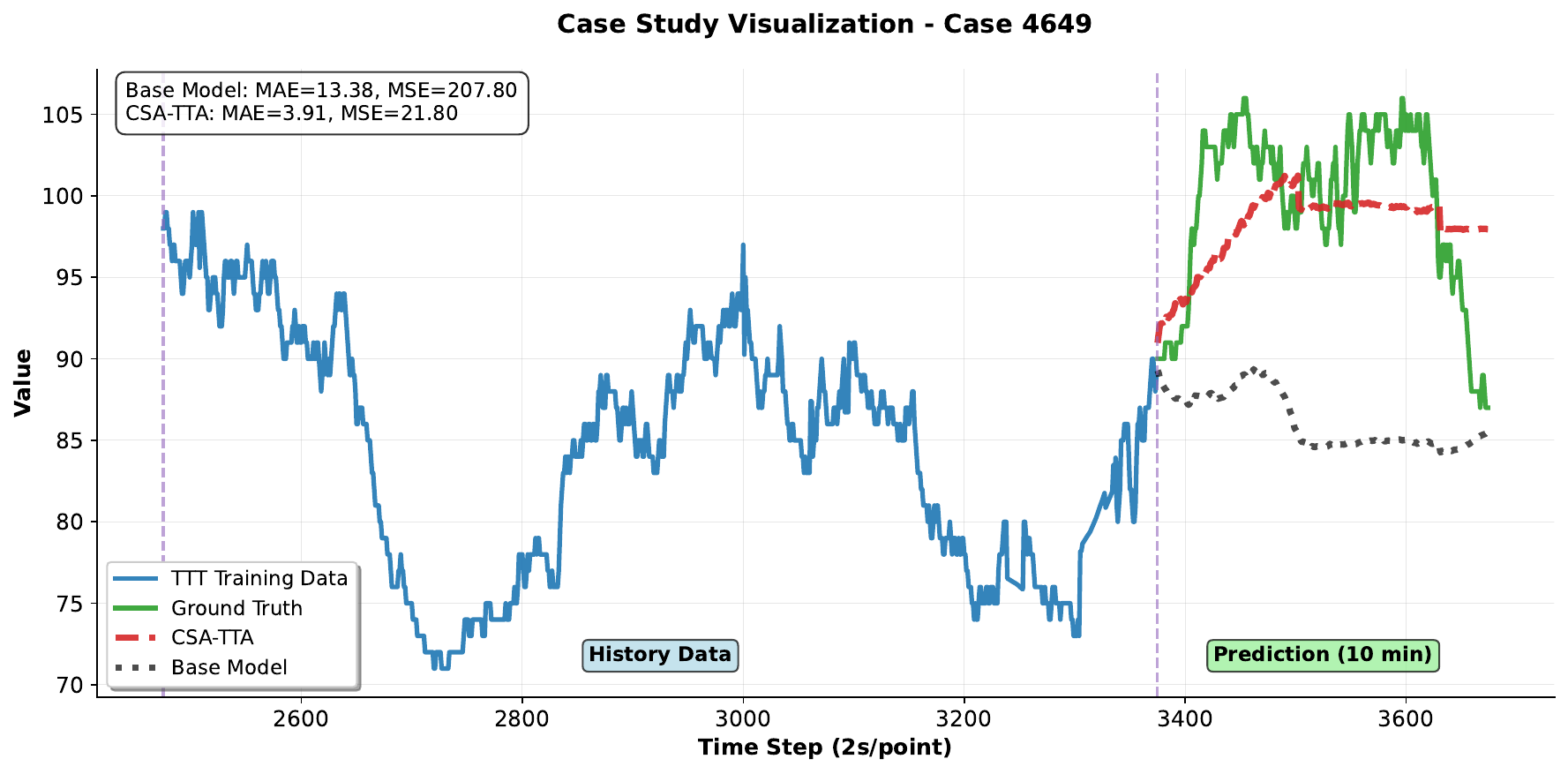}
        \caption{}\label{fig:case_study10_3}
    \end{subfigure}\hfill
    \begin{subfigure}{0.48\linewidth}
        \includegraphics[width=\linewidth, height=5cm]{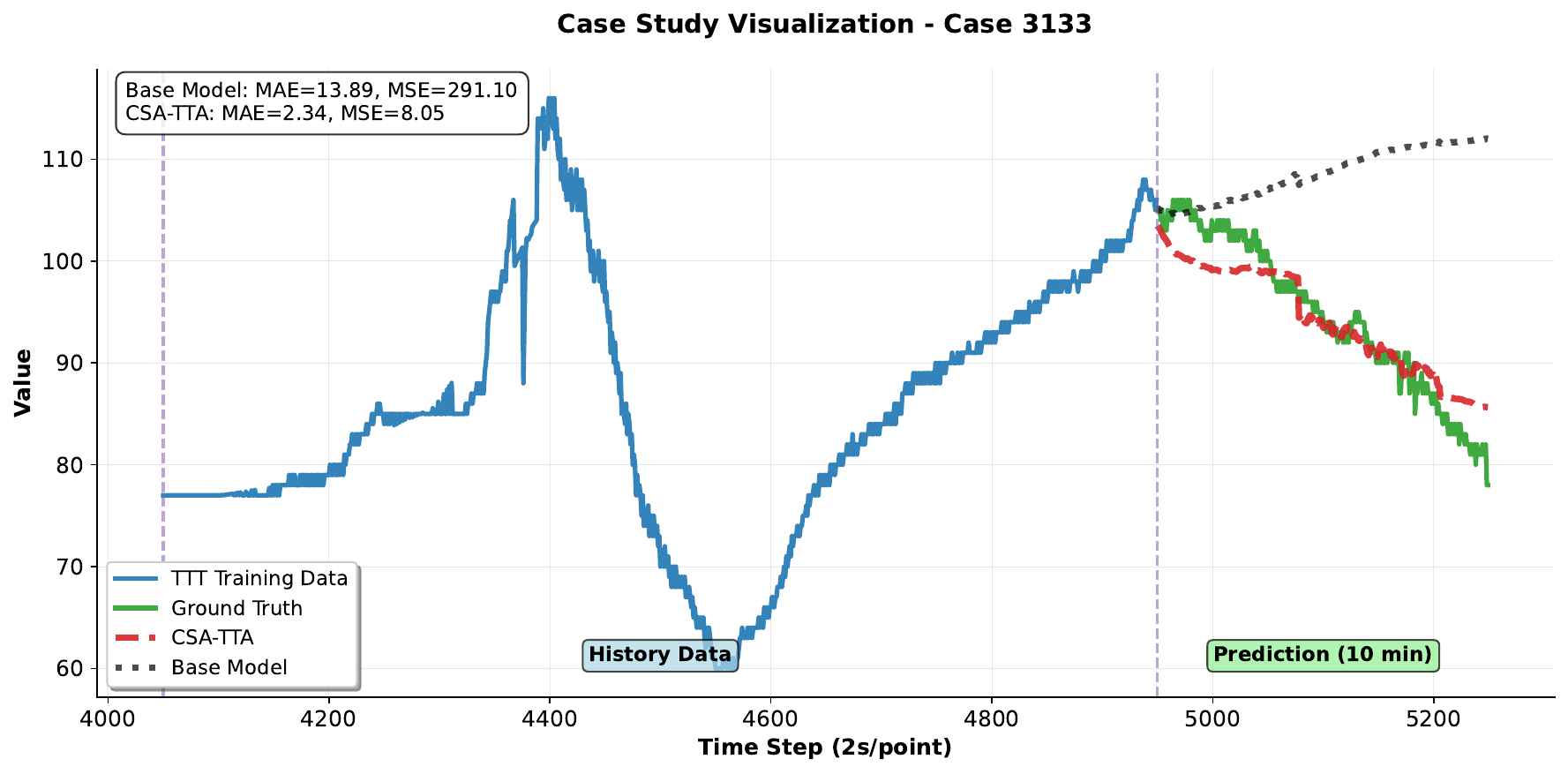}
        \caption{}\label{fig:case_study10_4}
    \end{subfigure}

    \vspace{0.5em}

    \begin{subfigure}{0.48\linewidth}
        \includegraphics[width=\linewidth, height=5cm]{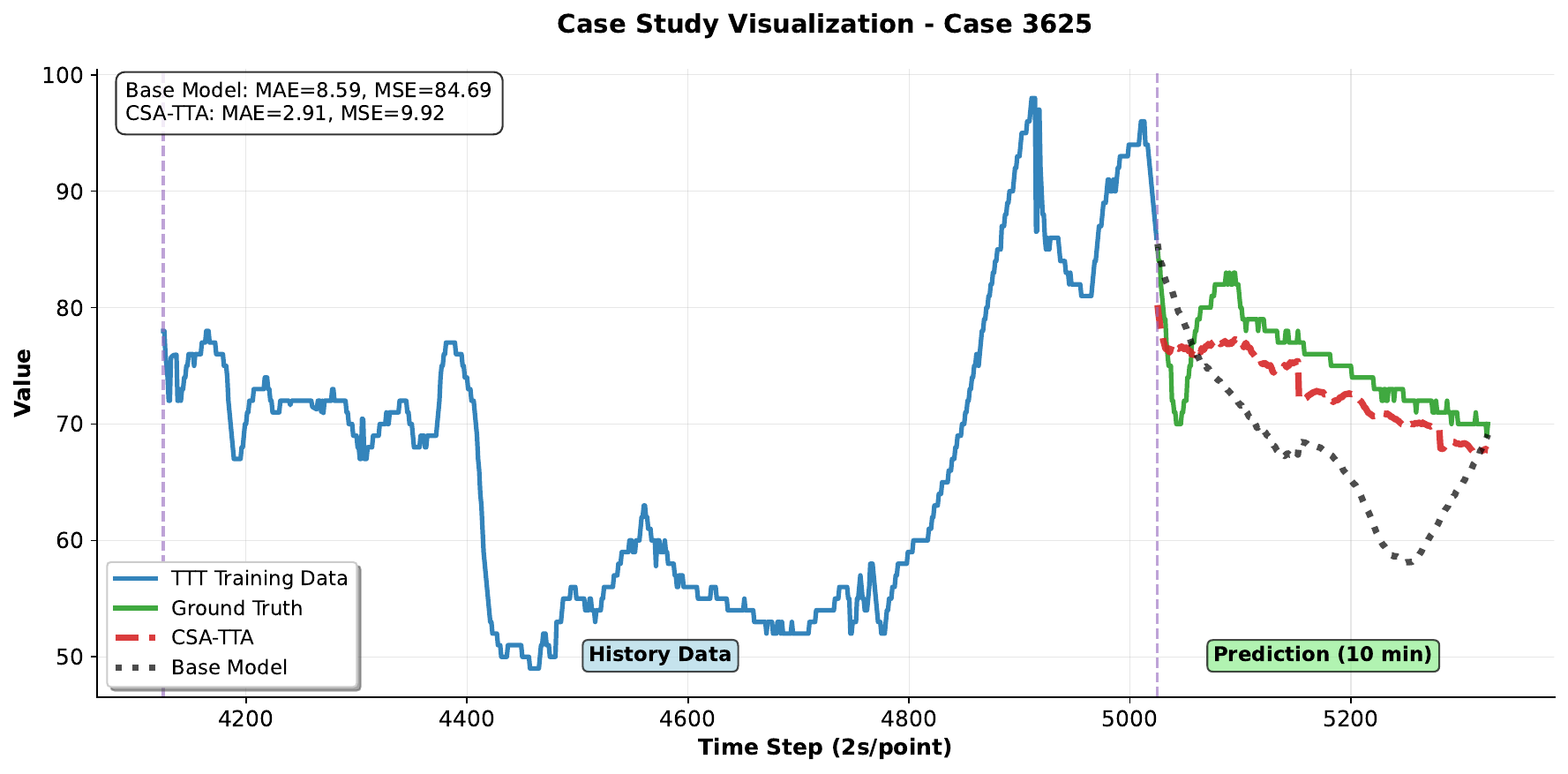}
        \caption{}\label{fig:case_study10_5}
    \end{subfigure}\hfill
    \begin{subfigure}{0.48\linewidth}
        \includegraphics[width=\linewidth, height=5cm]{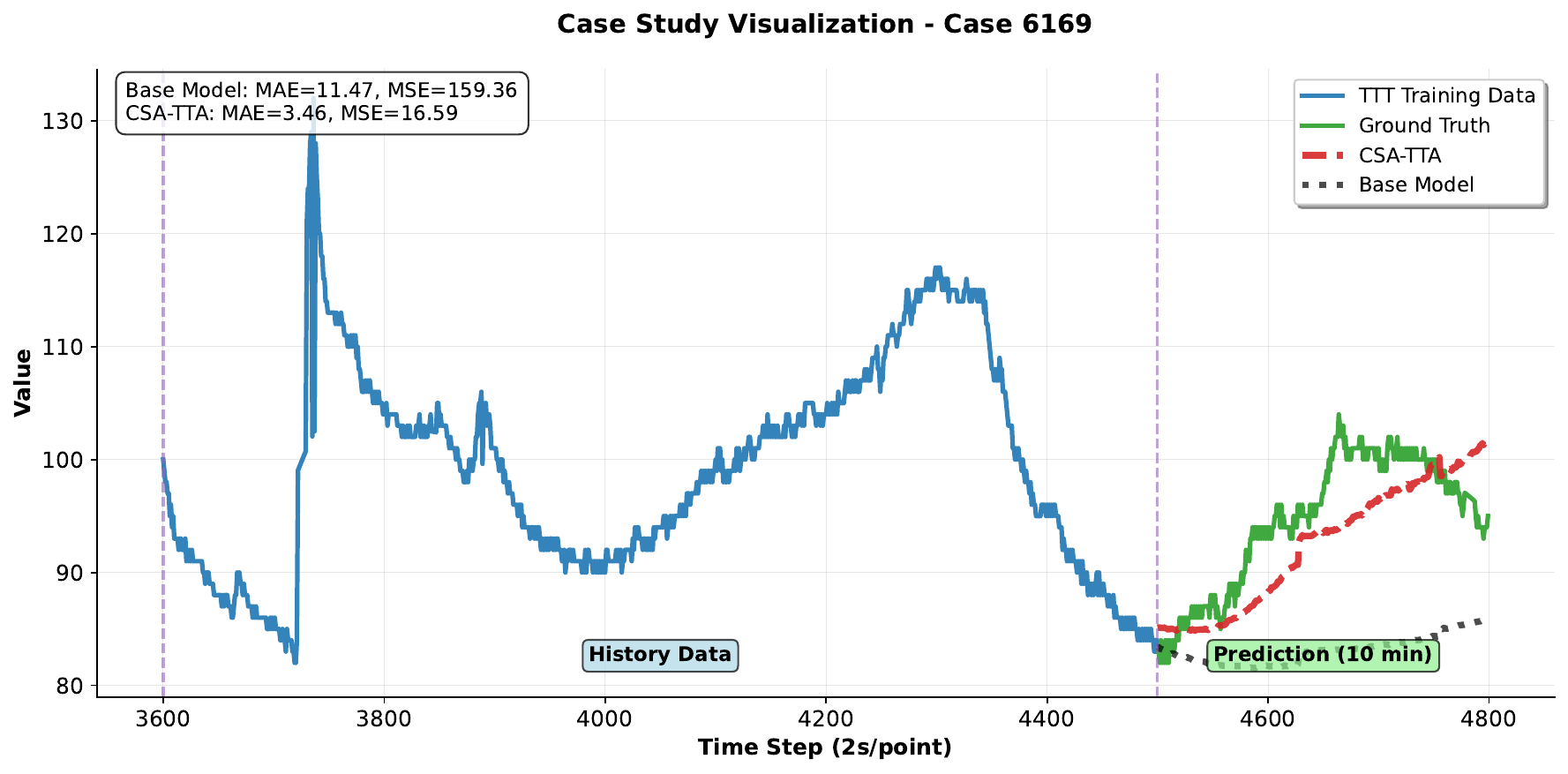}
        \caption{}\label{fig:case_study10_6}
    \end{subfigure}

    \vspace{0.5em}

    \begin{subfigure}{0.48\linewidth}
        \includegraphics[width=\linewidth, height=5cm]{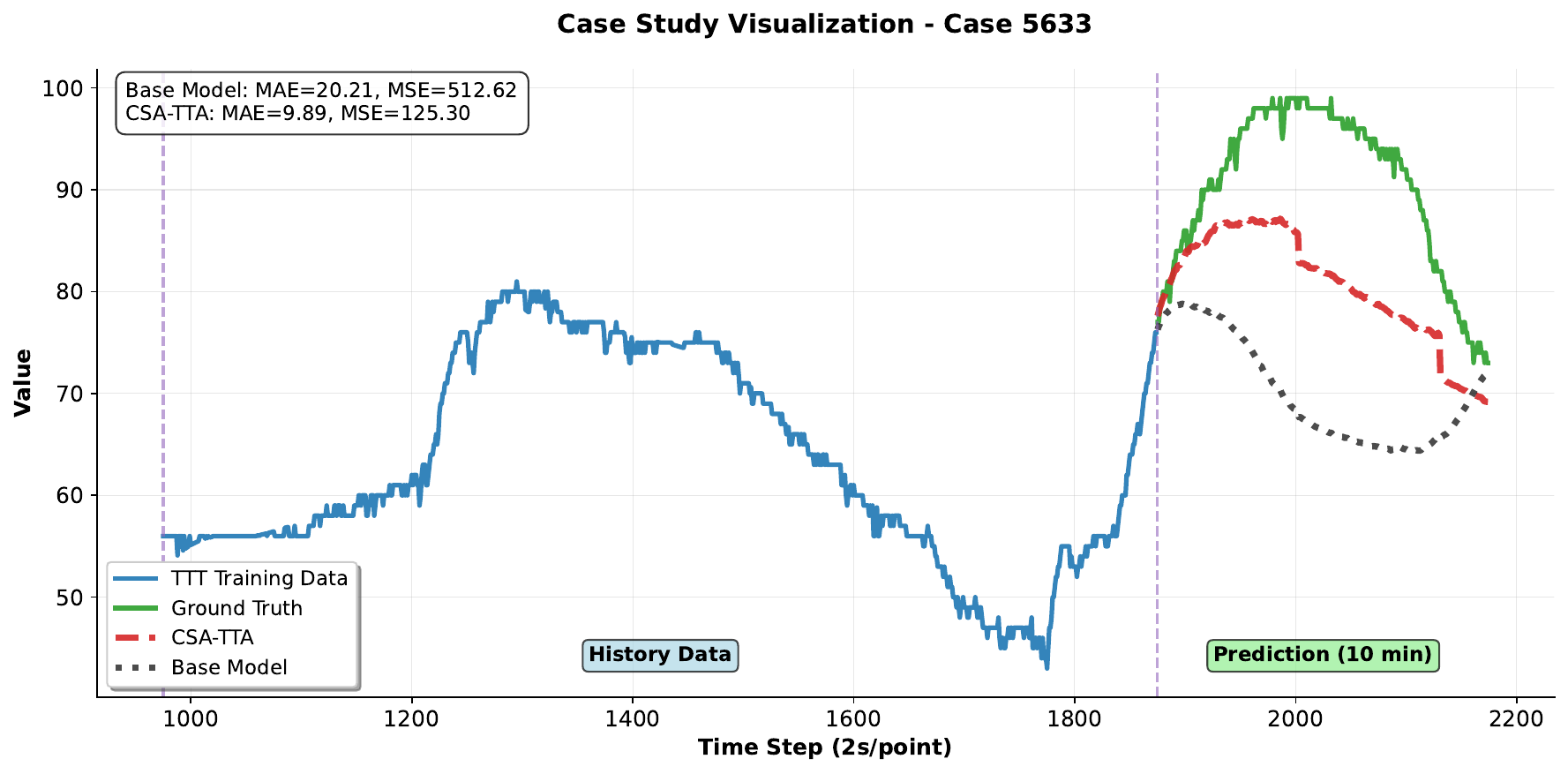}
        \caption{}\label{fig:case_study10_7}
    \end{subfigure}\hfill
    \begin{subfigure}{0.48\linewidth}
        \includegraphics[width=\linewidth, height=5cm]{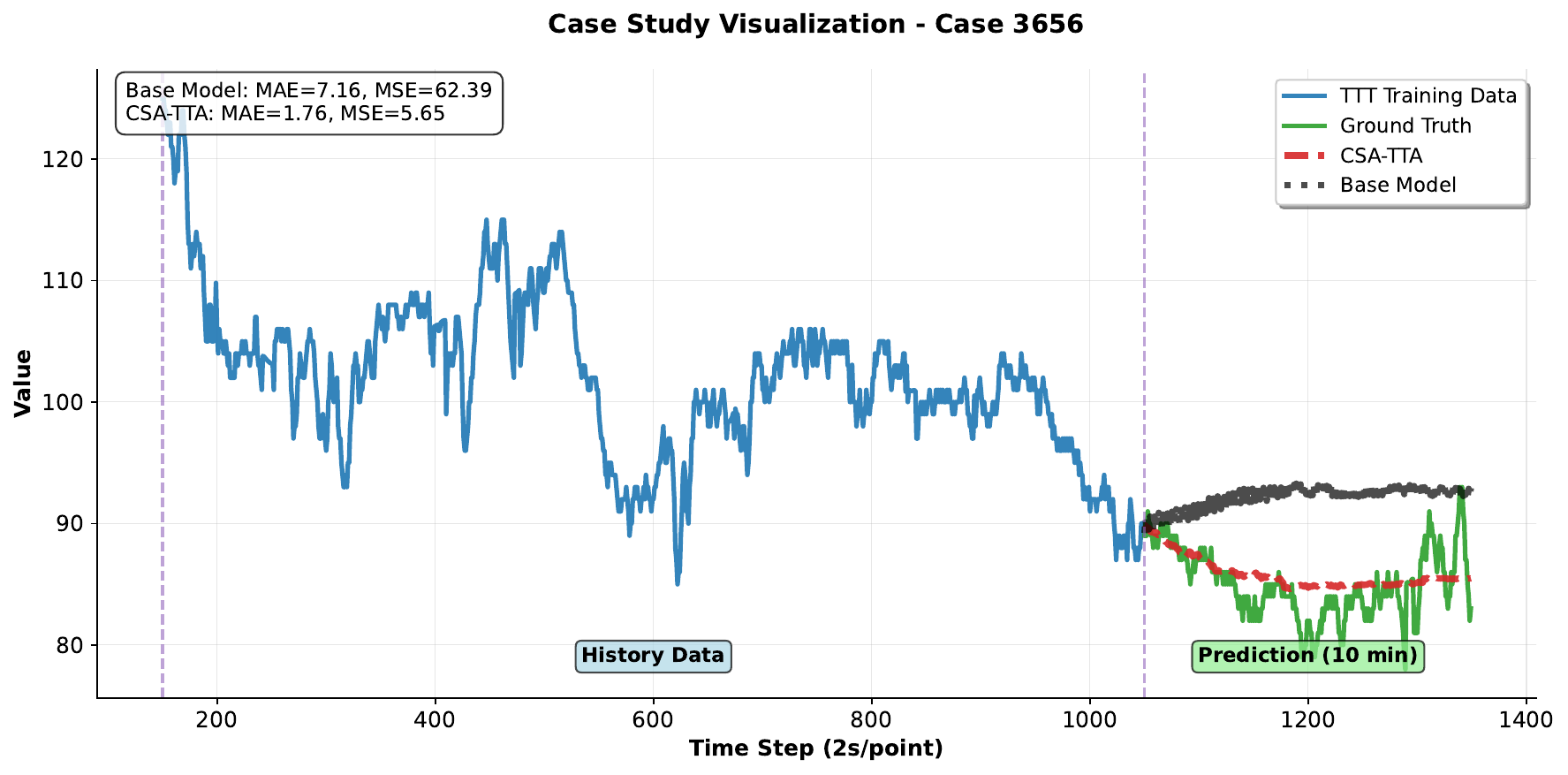}
        \caption{}\label{fig:case_study10_8}
    \end{subfigure}

    \caption{\textbf{Case study visualizations for the 10-minute prediction horizon.} Each subfigure shows an example case from the VitalDB dataset.}
    \label{fig:case_study_10}
\end{figure*}

\begin{figure*}[!t]
    \centering
    \begin{subfigure}{0.48\linewidth}
        \includegraphics[width=\linewidth, height=5cm]{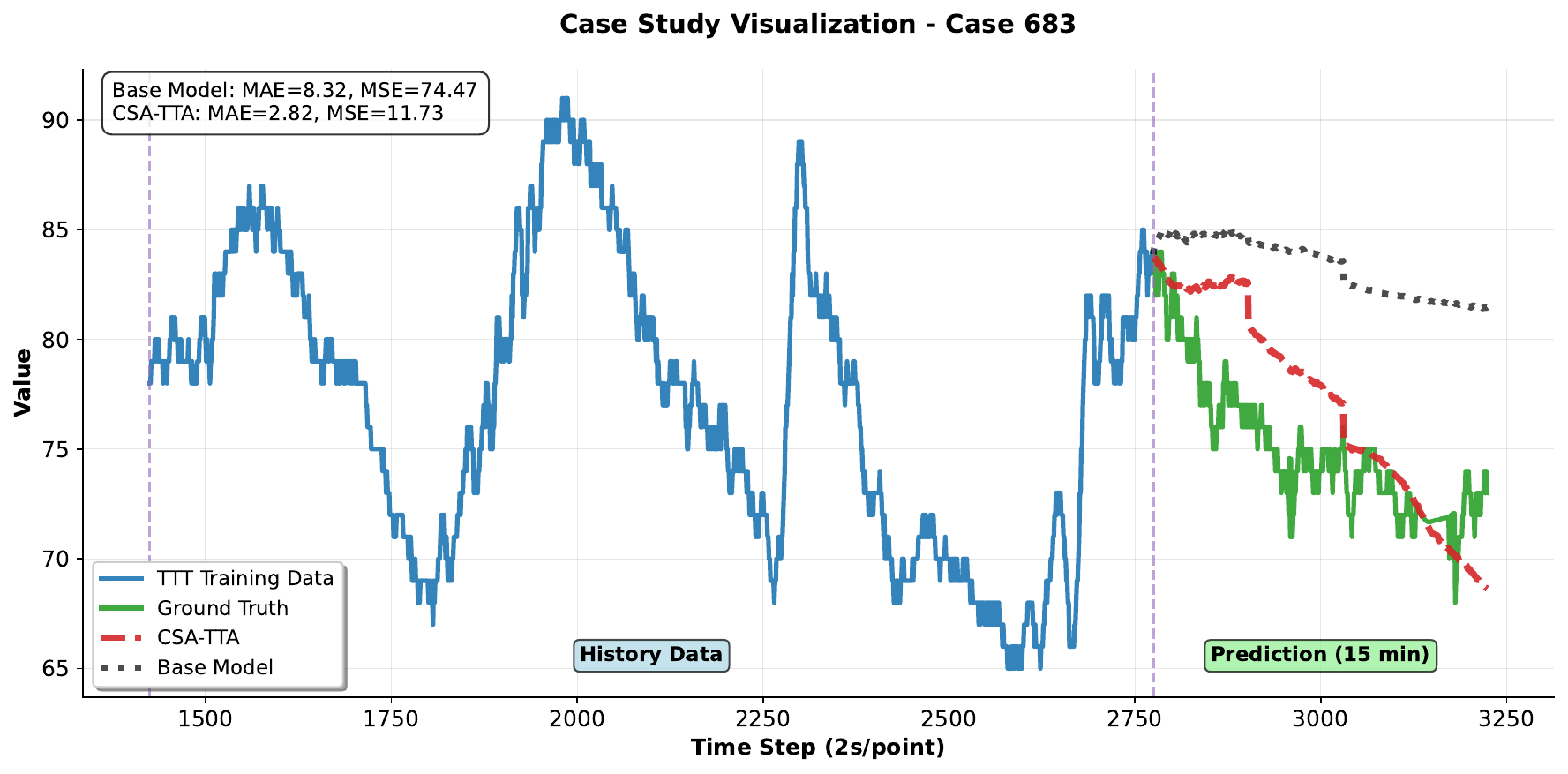}
        \caption{}\label{fig:case_study15_1}
    \end{subfigure}\hfill
    \begin{subfigure}{0.48\linewidth}
        \includegraphics[width=\linewidth, height=5cm]{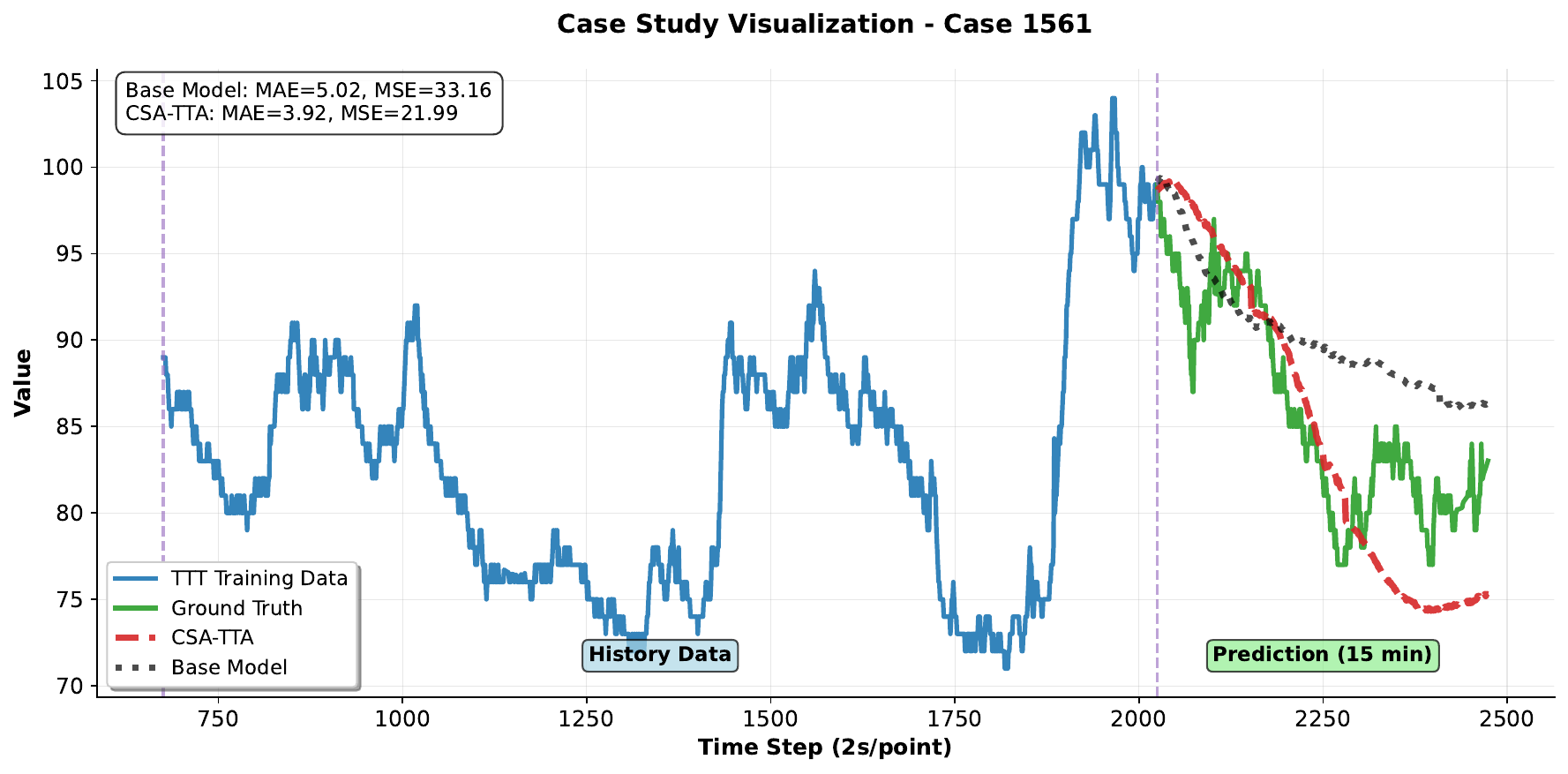}
        \caption{}\label{fig:case_study15_2}
    \end{subfigure}

    \vspace{0.5em}

    \begin{subfigure}{0.48\linewidth}
        \includegraphics[width=\linewidth, height=5cm]{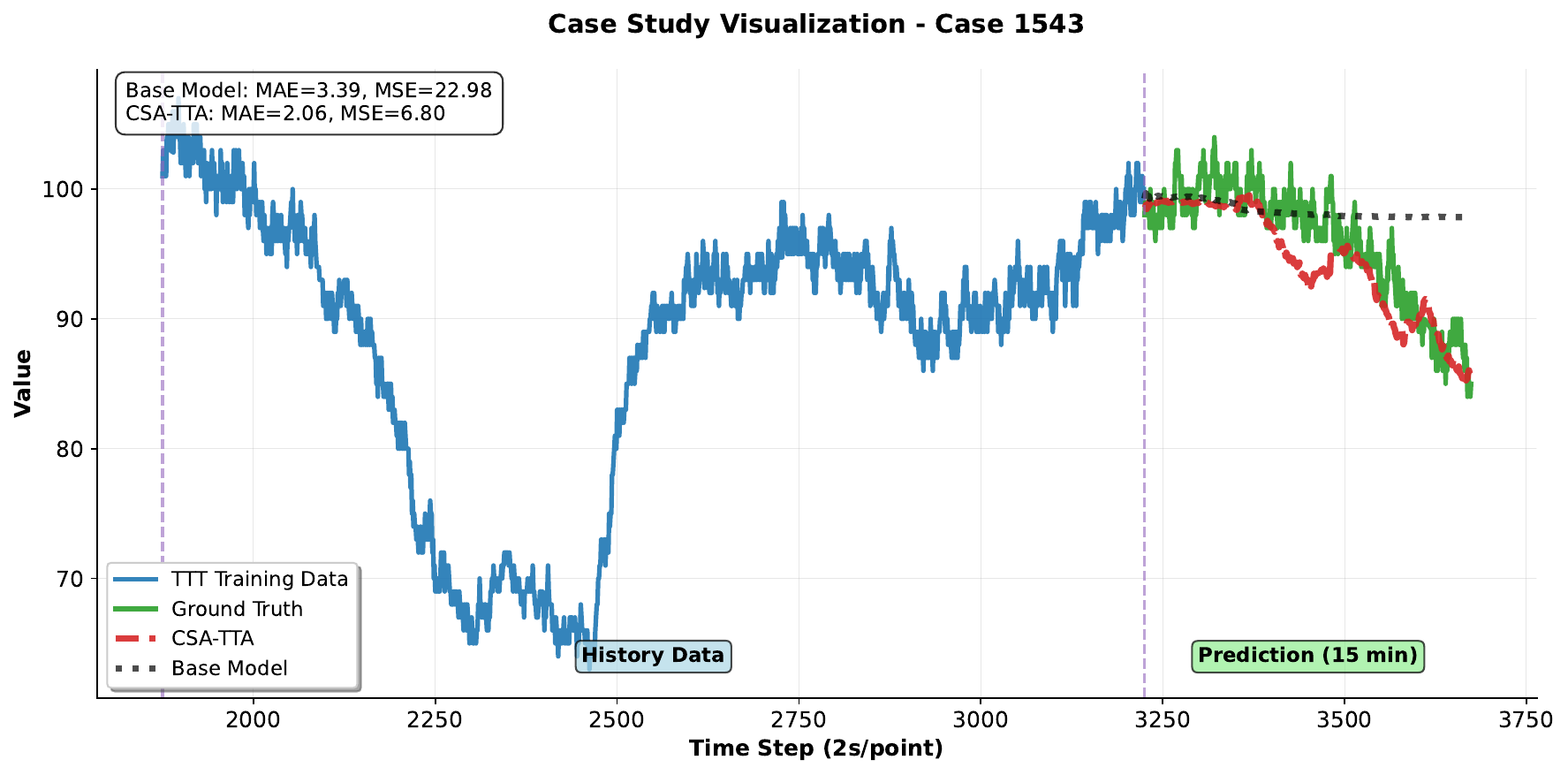}
        \caption{}\label{fig:case_study15_3}
    \end{subfigure}\hfill
    \begin{subfigure}{0.48\linewidth}
        \includegraphics[width=\linewidth, height=5cm]{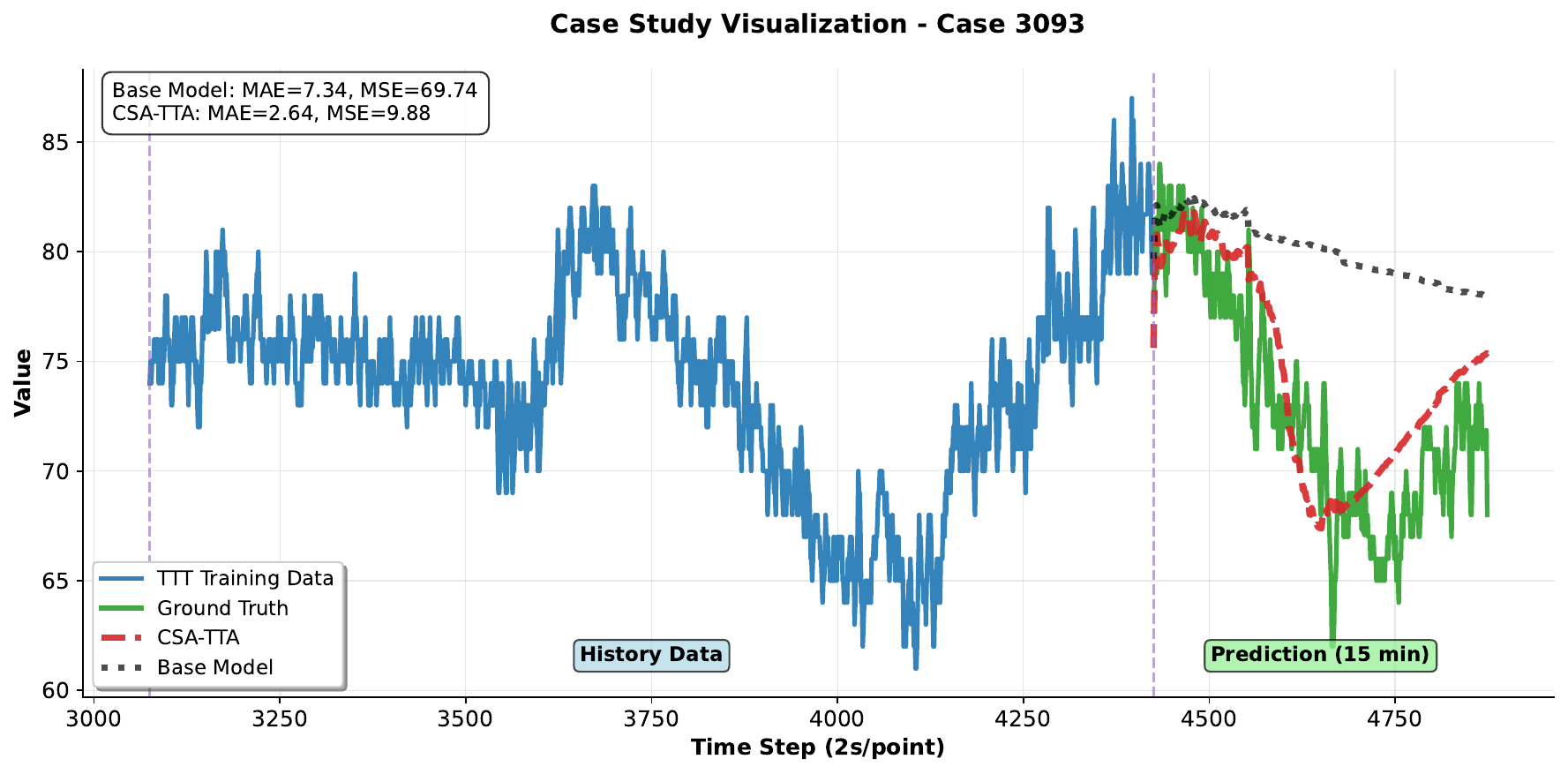}
        \caption{}\label{fig:case_study15_4}
    \end{subfigure}

    \vspace{0.5em}

    \begin{subfigure}{0.48\linewidth}
        \includegraphics[width=\linewidth, height=5cm]{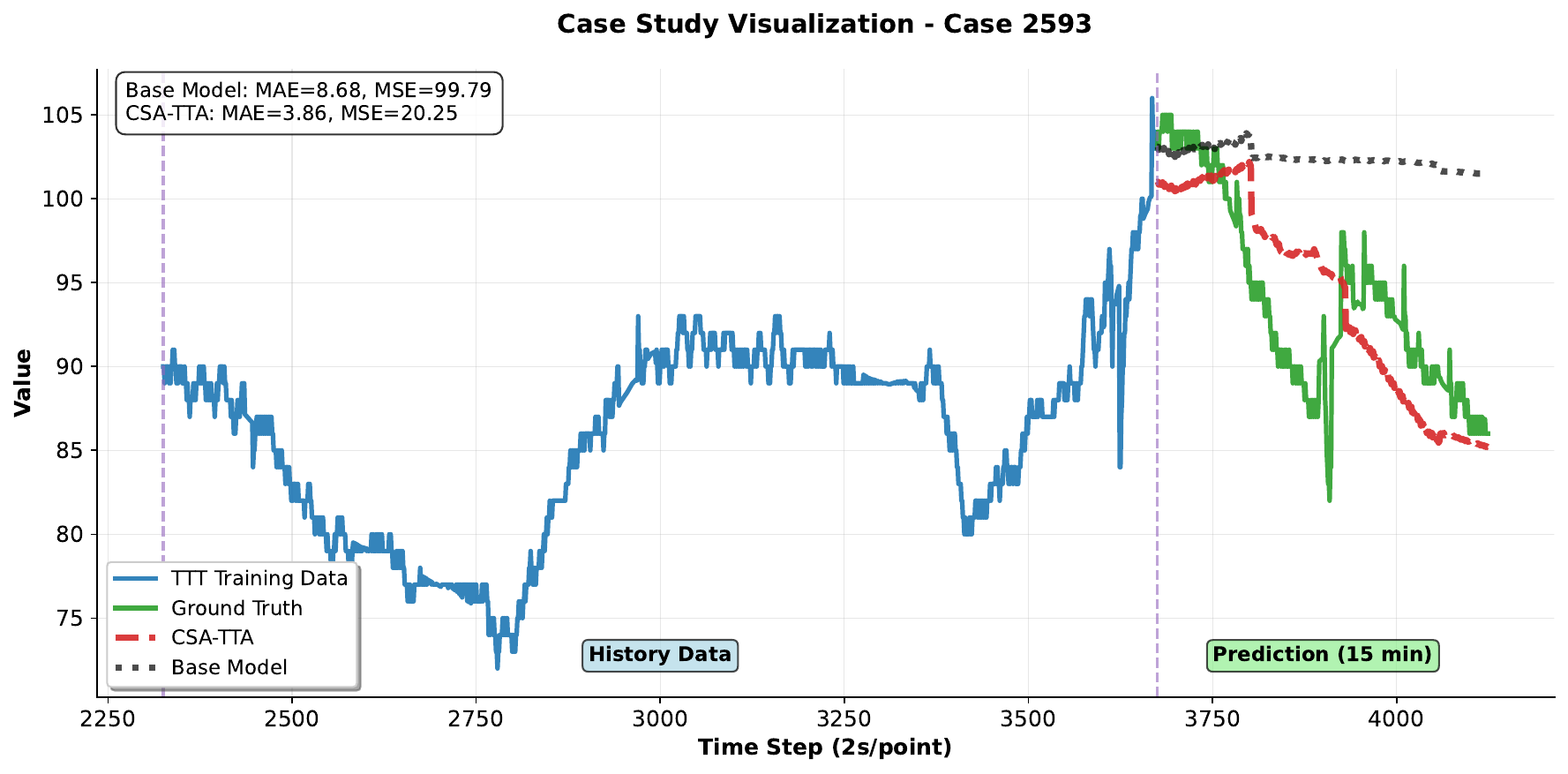}
        \caption{}\label{fig:case_study15_5}
    \end{subfigure}\hfill
    \begin{subfigure}{0.48\linewidth}
        \includegraphics[width=\linewidth, height=5cm]{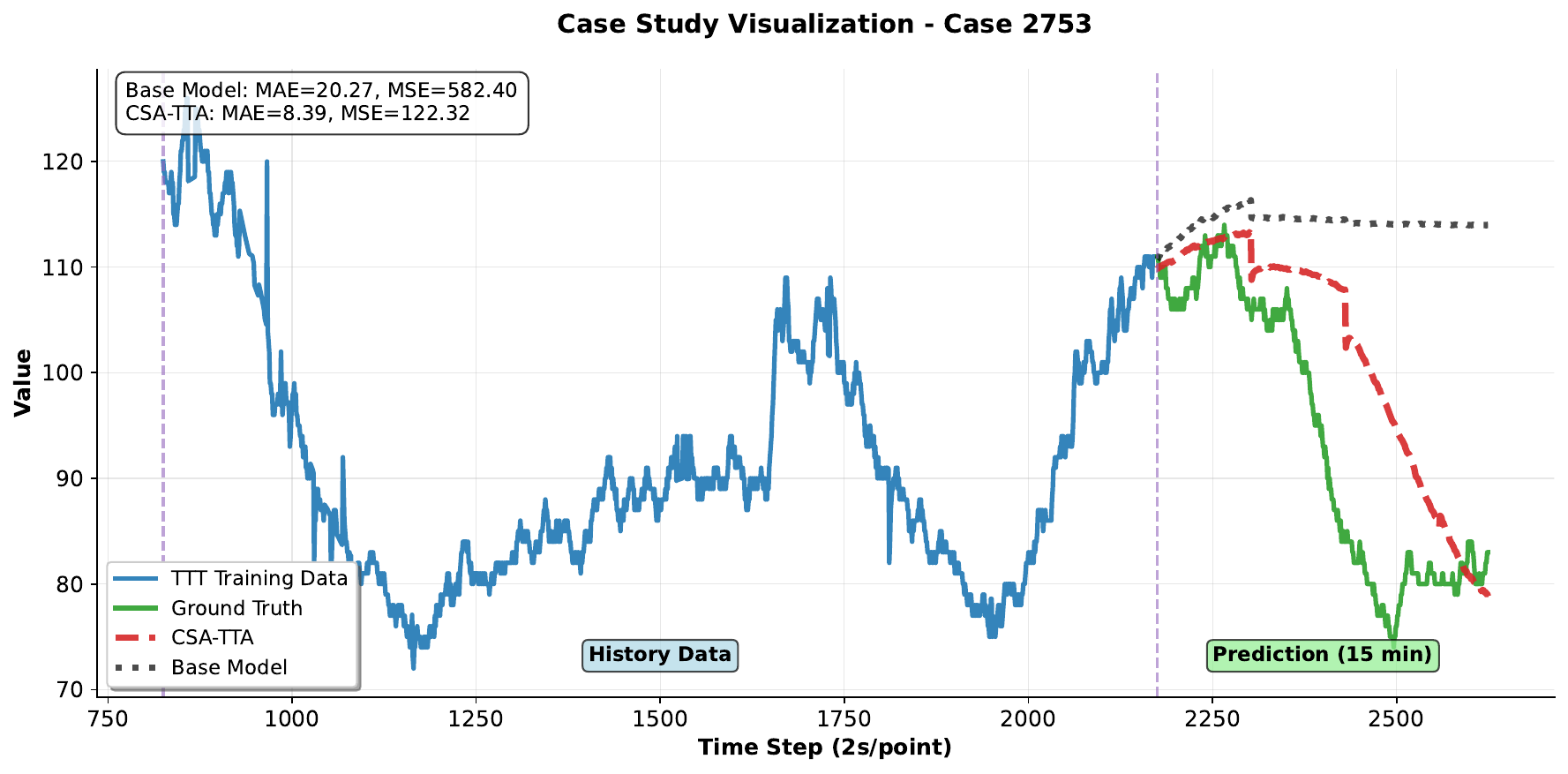}
        \caption{}\label{fig:case_study15_6}
    \end{subfigure}

    \vspace{0.5em}

    \begin{subfigure}{0.48\linewidth}
        \includegraphics[width=\linewidth, height=5cm]{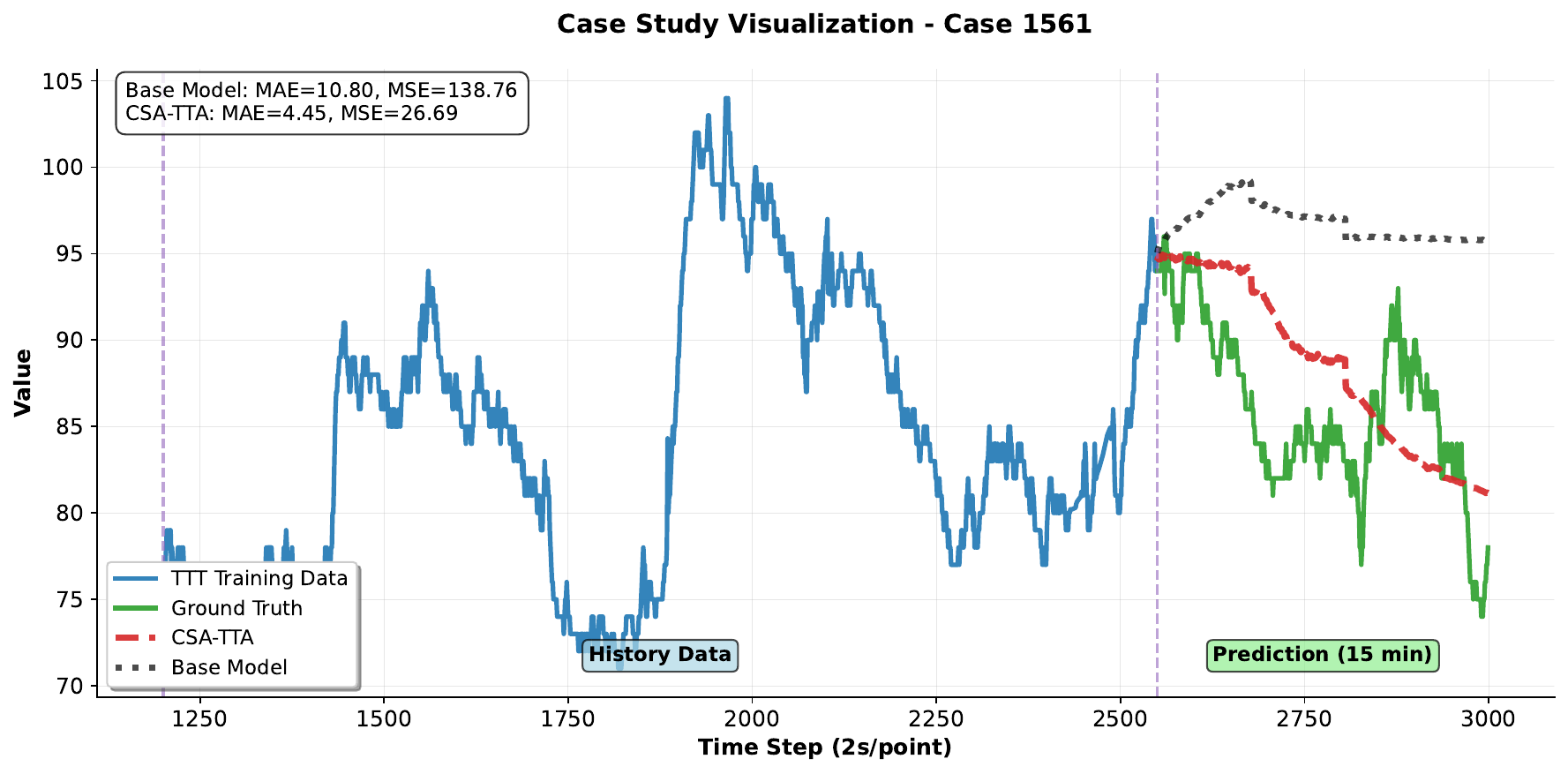}
        \caption{}\label{fig:case_study15_7}
    \end{subfigure}\hfill
    \begin{subfigure}{0.48\linewidth}
        \includegraphics[width=\linewidth, height=5cm]{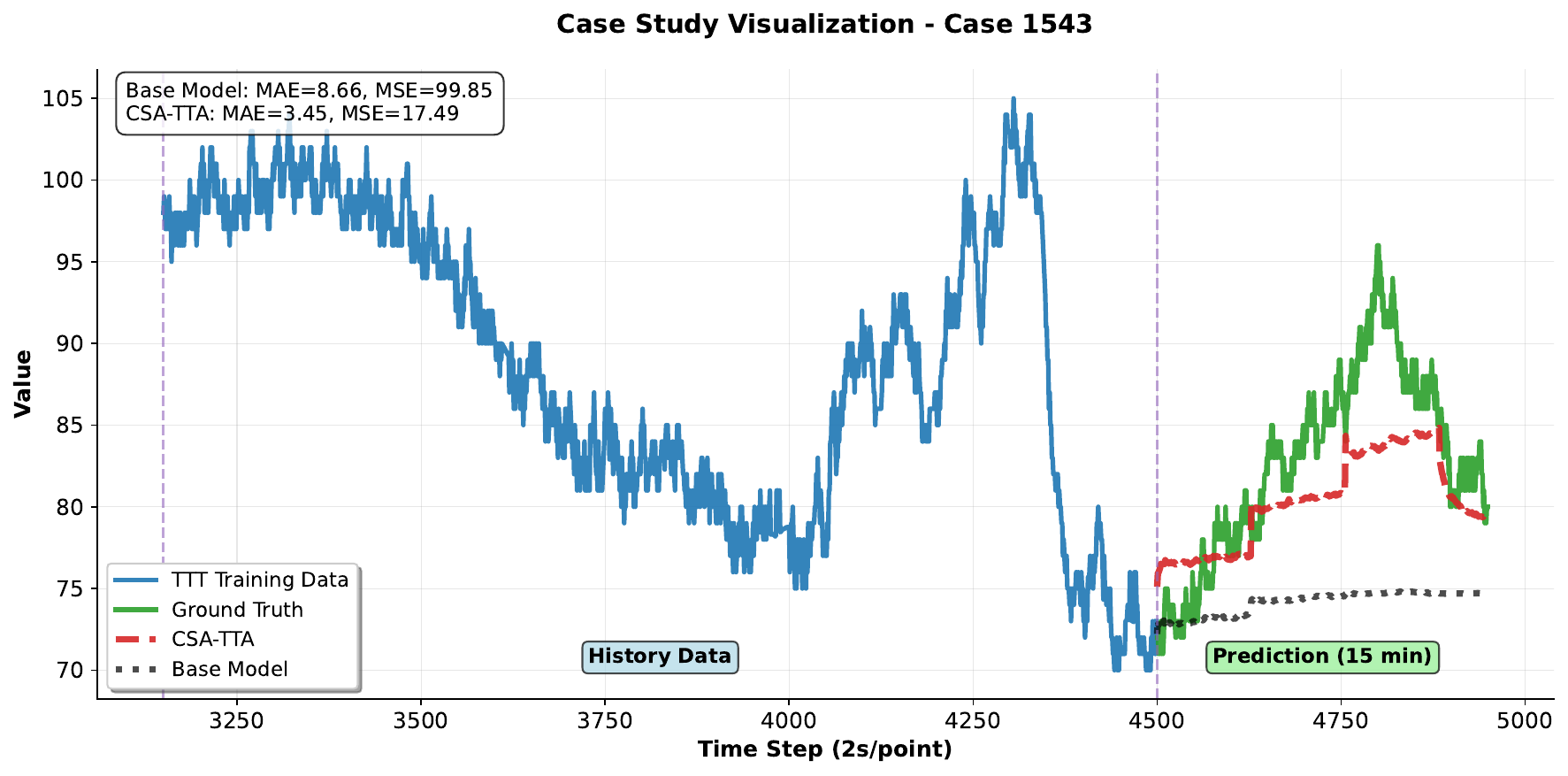}
        \caption{}\label{fig:case_study15_8}
    \end{subfigure}

    \caption{\textbf{Case study visualizations for the 15-minute prediction horizon.} Each subfigure shows an example case from the VitalDB dataset.}
    \label{fig:case_study_15}
\end{figure*}

\paragraph{Visual case studies.}
To illustrate the forecasting process, we present case studies from VitalDB in Fig.~\ref{fig:case_study_5}, Fig.~\ref{fig:case_study_10}, and Fig.~\ref{fig:case_study_15}, corresponding to prediction horizons of 5, 10, and 15 minutes respectively. Each model performs forecasting based on a fixed 15-minute input time series window. We compare the forecasting results of the proposed CSA-TTA method with those of baseline models. In each subplot, we simultaneously display the ground truth curve, the forecasts generated by the baseline model, and those from CSA-TTA. In addition, we report the corresponding MAE and MSE for each prediction relative to the ground truth to quantitatively assess performance. 

CSA-TTA more accurately captures complex personalized intraoperative dynamics across all horizons, especially rapid blood pressure drops or sudden rebounds that baseline models tend to smooth over or miss. 
Such visualizations highlight CSA-TTA’s ability to detect abrupt changes and subtle temporal patterns often overlooked by other models. This case-level analysis clearly demonstrates CSA-TTA’s enhanced adaptability and precision in personalized intraoperative hypotension prediction.

\definecolor{checkmark}{rgb}{0.1,0.8,0.1}
\definecolor{crossmark}{rgb}{1,0.1,0.1}
\definecolor{lightgray}{gray}{0.9}

\begin{table*}[ht]
\centering
\begin{subtable}[t]{\textwidth}
\centering
\caption{\textbf{Zero-shot setting}}
\begin{tabular}{l|l|rrrrrr|rrrrrr}
\toprule
\multicolumn{2}{c|}{\textbf{Setting}} & \multicolumn{6}{c|}{\textbf{VitalDB (2S)}} & \multicolumn{6}{c}{\textbf{VitalDB (30S)}} \\
\cmidrule(lr){1-2} \cmidrule(lr){3-8} \cmidrule(lr){9-14}
\textbf{Horizon} & \textbf{Model} & \textbf{F1$\uparrow$} & \textbf{Rec$\uparrow$} & \textbf{Prec$\uparrow$} & \textbf{Acc$\uparrow$} & \textbf{MAE$\downarrow$} & \textbf{MSE$\downarrow$}
   & \textbf{F1$\uparrow$} & \textbf{Rec$\uparrow$} & \textbf{Prec$\uparrow$} & \textbf{Acc$\uparrow$} & \textbf{MAE$\downarrow$} & \textbf{MSE$\downarrow$} \\
\midrule
\multirow{2}{*}{5min} & Test\cite{timesfm} & 67.70 & 76.50 & 60.70 & 91.90 & 4.94 & 57.54 & 68.90 & 66.70 & 71.10 & 91.70 & 5.20 & 63.79 \\
 & \cellcolor{lightgray}CSA-TTA(Ours) & \cellcolor{lightgray}\textbf{68.10} & \cellcolor{lightgray}\textbf{77.07} & \cellcolor{lightgray}\textbf{61.00} & \cellcolor{lightgray}\textbf{92.00} & \cellcolor{lightgray}\textbf{4.82} & \cellcolor{lightgray}\textbf{55.31} & \cellcolor{lightgray}\textbf{69.80} & \cellcolor{lightgray}\textbf{67.40} & \cellcolor{lightgray}\textbf{72.50} & \cellcolor{lightgray}\textbf{92.00} & \cellcolor{lightgray}\textbf{5.07} & \textbf{59.32} \\
\midrule
\multirow{2}{*}{10min} & Test\cite{timesfm} & 61.30 & 62.70 & 60.00 & 87.70 & 6.46 & 89.39 & 63.30 & 57.90 & 69.90 & 87.10 & 6.79 & 99.15 \\
 & \cellcolor{lightgray}CSA-TTA(Ours) & \cellcolor{lightgray}\textbf{63.30} & \cellcolor{lightgray}\textbf{63.80} & \cellcolor{lightgray}\textbf{62.81} & \cellcolor{lightgray}\textbf{88.20} & \cellcolor{lightgray}\textbf{6.41} & \cellcolor{lightgray}\textbf{87.13 }& \cellcolor{lightgray}\textbf{63.70} & \cellcolor{lightgray}\textbf{58.10} & \cellcolor{lightgray}\textbf{70.50} & \cellcolor{lightgray}\textbf{87.60} & \cellcolor{lightgray}\textbf{6.54} & \cellcolor{lightgray}\textbf{91.01} \\
\midrule
\multirow{2}{*}{15min} & Test\cite{timesfm} & 58.20 & 55.70 & 60.90 & 84.60 & 7.16 & 103.90 & 60.30 & 52.00 & 71.60 & 84.00 & 7.49 & 115.37 \\
 & \cellcolor{lightgray}CSA-TTA(Ours) & \cellcolor{lightgray}\textbf{60.90} & \cellcolor{lightgray}\textbf{57.41} & \cellcolor{lightgray}\textbf{64.84} & \cellcolor{lightgray}\textbf{85.00} & \cellcolor{lightgray}\textbf{6.98} & \cellcolor{lightgray}\textbf{99.21} & \cellcolor{lightgray}\textbf{61.20} & \cellcolor{lightgray}\textbf{52.30} & \cellcolor{lightgray}\textbf{73.60 }& \cellcolor{lightgray}\textbf{84.20} & \cellcolor{lightgray}\textbf{7.23} & \cellcolor{lightgray}\textbf{105.52} \\
\bottomrule
\end{tabular}
\end{subtable}

\vspace{0.5cm}

\begin{subtable}[t]{\textwidth}
\centering
\caption{\textbf{Fine-tuned setting}}
\begin{tabular}{l|l|rrrrrr|rrrrrr}
\toprule
\multicolumn{2}{c|}{\textbf{Setting}} & \multicolumn{6}{c|}{\textbf{VitalDB (2S)}} & \multicolumn{6}{c}{\textbf{VitalDB (30S)}} \\
\cmidrule(lr){1-2} \cmidrule(lr){3-8} \cmidrule(lr){9-14}
\textbf{Horizon} & \textbf{Model} & \textbf{F1$\uparrow$} & \textbf{Rec$\uparrow$} & \textbf{Prec$\uparrow$} & \textbf{Acc$\uparrow$} & \textbf{MAE$\downarrow$} & \textbf{MSE$\downarrow$}
   & \textbf{F1$\uparrow$} & \textbf{Rec$\uparrow$} & \textbf{Prec$\uparrow$} & \textbf{Acc$\uparrow$} & \textbf{MAE$\downarrow$} & \textbf{MSE$\downarrow$} \\
\midrule
\multirow{4}{*}{5min} & Test\cite{timesfm} & \underline{68.50} & \underline{77.40} & 61.40 & 92.10 & 4.79 & 53.86 & 70.00 & \textbf{72.00} & 68.10 & 91.50 & 4.87 & \underline{56.00} \\
 & TTT\cite{ttt} & 68.40 & 77.10 & \underline{61.46} & 92.10 & 4.80 & 53.72 & 70.00 & \textbf{72.00} & 68.10 & 91.50 & 4.87 & 56.11 \\
 & TTT++\cite{ttt++} & 68.40 & 77.30 & 61.34 & 92.10 & \underline{4.77} & \underline{53.53} & 70.00 & \textbf{72.00} & 68.10 & 91.50 & \underline{4.86} & 56.01 \\
 & \cellcolor{lightgray}CSA-TTA(Ours) & \cellcolor{lightgray}\textbf{69.20} & \cellcolor{lightgray}\textbf{78.61} & \cellcolor{lightgray}\textbf{61.80} & \cellcolor{lightgray}\textbf{92.40} & \cellcolor{lightgray}\textbf{4.70} & \cellcolor{lightgray}\textbf{52.42} & \cellcolor{lightgray}\textbf{70.60} & \cellcolor{lightgray}71.60 & \cellcolor{lightgray}\textbf{69.70} & \cellcolor{lightgray}\textbf{91.80} & \cellcolor{lightgray}\textbf{4.77} & \cellcolor{lightgray}\textbf{53.17} \\
\midrule
\multirow{4}{*}{10min} & Test\cite{timesfm} & \underline{62.80} & \underline{64.50} & \underline{61.10} & 88.10 & 6.57 & 90.14 & \underline{64.60} & \underline{60.70} & 69.10 & 87.20 & \underline{6.22} & 82.56 \\
 & TTT\cite{ttt} & 62.60 & 64.20 & 61.08 & 88.10 & \underline{6.56} & 90.21 & 64.50 & 60.50 & 69.10 & 87.20 & 6.21 & 82.43 \\
 & TTT++\cite{ttt++} & 62.70 & 64.40 & 61.09 & 88.10 & \underline{6.56} & \underline{90.08} & \underline{64.60} & 60.50 & \textbf{69.20 }& \underline{87.30} & \underline{6.22} & \underline{82.39} \\
 & \cellcolor{lightgray}CSA-TTA(Ours) & \cellcolor{lightgray}\textbf{63.60} & \cellcolor{lightgray}\textbf{65.96} & \cellcolor{lightgray}\textbf{61.40} & \cellcolor{lightgray}\textbf{88.40} & \cellcolor{lightgray}\textbf{6.50} & \cellcolor{lightgray}\textbf{89.11} & \cellcolor{lightgray}\textbf{64.70} & \cellcolor{lightgray}\textbf{60.80} & \cellcolor{lightgray}\textbf{69.20} & \cellcolor{lightgray}\textbf{87.50} & \cellcolor{lightgray}\textbf{6.05} & \cellcolor{lightgray}\textbf{77.60} \\
\midrule 
\multirow{4}{*}{15min} & Test\cite{timesfm} & \underline{61.30} & 52.90 & 72.90 & \underline{87.20} &\underline{6.72} & 89.62 & 62.80 & \underline{55.30} & \underline{72.70} & 84.60 & 6.72 & 90.24 \\
 & TTT\cite{ttt} & 61.00 & \underline{53.00} & 71.84 & 87.00 & 6.70 & \underline{89.16} & 62.80 & \textbf{55.40} & 72.60 & 84.60 & 6.73 & 90.29 \\
 & TTT++\cite{ttt++} & 61.20 & 52.70 & \underline{72.97} & \underline{87.20} & \underline{6.72} & 89.43 & 62.80 & \underline{55.30} & \underline{72.70} & 84.60 & \underline{6.71} & \underline{90.18} \\
 & \cellcolor{lightgray}CSA-TTA(Ours) & \cellcolor{lightgray}\textbf{61.70} & \cellcolor{lightgray}\textbf{53.40} & \cellcolor{lightgray}\textbf{73.05} & \cellcolor{lightgray}\textbf{87.40} & \cellcolor{lightgray}\textbf{6.62} & \cellcolor{lightgray}\textbf{87.03} & \cellcolor{lightgray}\textbf{62.90} & \cellcolor{lightgray}54.60 & \cellcolor{lightgray}\textbf{74.00} & \cellcolor{lightgray}\textbf{84.80} & \cellcolor{lightgray}\textbf{6.64} & \cellcolor{lightgray}\textbf{88.02} \\
\bottomrule
\end{tabular}
\end{subtable}
\caption{\textbf{Full results of the TimesFM backbone on VitalDB (2S/30S) under different adaptation settings: (a) zero-shot setting; (b) fine-tuning setting.} Results are shown for three prediction horizons (5, 10, and 15 minutes) with 15-minute lookback windows. The best performance is highlighted in \textbf{bold}, and the second-best is \underline{underlined}. Metrics with the $\uparrow$ symbol (e.g., F1, Recall) indicate higher is better, while those with $\downarrow$ (e.g., MAE, MSE) indicate lower is better.}
\label{tab:timesfm_full_results}
\end{table*}

\begin{table*}[ht]
\centering

\begin{subtable}[t]{\textwidth}
\centering
\caption{\textbf{Zero-shot setting}}
\begin{tabular}{l|l|rrrrrr|rrrrrr}
\toprule
\multicolumn{2}{c|}{\textbf{Setting}} & \multicolumn{6}{c|}{\textbf{VitalDB (2S)}} & \multicolumn{6}{c}{\textbf{VitalDB (30S)}} \\
\cmidrule(lr){1-2} \cmidrule(lr){3-8} \cmidrule(lr){9-14}
\textbf{Horizon} & \textbf{Model} & \textbf{F1$\uparrow$} & \textbf{Rec$\uparrow$} & \textbf{Prec$\uparrow$} & \textbf{Acc$\uparrow$} & \textbf{MAE$\downarrow$} & \textbf{MSE$\downarrow$}
   & \textbf{F1$\uparrow$} & \textbf{Rec$\uparrow$} & \textbf{Prec$\uparrow$} & \textbf{Acc$\uparrow$} & \textbf{MAE$\downarrow$} & \textbf{MSE$\downarrow$} \\
\midrule
\multirow{2}{*}{5min} & Test\cite{units} & 53.50 & 55.40 & \textbf{51.73 }& 90.50 & 7.08 & 93.26 & 56.90 & 49.27 & 67.33 & 89.50 & 6.94 & 89.95 \\
 & \cellcolor{lightgray}CSA-TTA(Ours) & \cellcolor{lightgray}\textbf{57.40} & \cellcolor{lightgray}\textbf{64.50} & \cellcolor{lightgray}51.71 & \cellcolor{lightgray}\textbf{90.80} & \cellcolor{lightgray}\textbf{6.94} & \cellcolor{lightgray}\textbf{91.63} & \cellcolor{lightgray}\textbf{62.80} & \cellcolor{lightgray}\textbf{57.60} & \cellcolor{lightgray}\textbf{69.03} & \cellcolor{lightgray}\textbf{89.80} & \cellcolor{lightgray}\textbf{6.77} & \cellcolor{lightgray}\textbf{87.87} \\
\midrule
\multirow{2}{*}{10min} & Test\cite{units} & 48.70 & 43.50 & 55.31 & 88.00 & 7.42 & 100.24 & 52.50 & 42.26 & \textbf{69.30} & 84.80 & 7.44 & 105.46 \\
 & \cellcolor{lightgray}CSA-TTA(Ours) & \cellcolor{lightgray}\textbf{53.60} & \cellcolor{lightgray}\textbf{50.60} & \cellcolor{lightgray}\textbf{56.98} & \cellcolor{lightgray}\textbf{88.30} & \cellcolor{lightgray}\textbf{7.27} & \cellcolor{lightgray}\textbf{98.30} & \cellcolor{lightgray}\textbf{57.50} & \cellcolor{lightgray}\textbf{52.70} & \cellcolor{lightgray}63.26 & \cellcolor{lightgray}\textbf{85.30} & \cellcolor{lightgray}\textbf{7.38} & \cellcolor{lightgray}\textbf{100.97} \\
\midrule
\multirow{2}{*}{15min} & Test\cite{units} & 46.80 & 35.60 & 68.28 & 84.10 & 7.59 & 103.15 & 47.30 & 38.20 & \textbf{76.80} & 82.10 & 7.58 & 104.48 \\
 & \cellcolor{lightgray}CSA-TTA(Ours) & \cellcolor{lightgray}\textbf{52.60} & \cellcolor{lightgray}\textbf{42.40} & \cellcolor{lightgray}\textbf{69.26} & \cellcolor{lightgray}\textbf{85.00} & \cellcolor{lightgray}\textbf{7.45} & \cellcolor{lightgray}\textbf{99.92} & \cellcolor{lightgray}\textbf{51.60} & \cellcolor{lightgray}\textbf{41.80} & \cellcolor{lightgray}67.40 & \cellcolor{lightgray}\textbf{82.40} & \cellcolor{lightgray}\textbf{7.43} & \cellcolor{lightgray}\textbf{98.67} \\
\bottomrule
\end{tabular}
\end{subtable}

\vspace{0.4cm}

\begin{subtable}[t]{\textwidth}
\centering
\caption{\textbf{Fune-tuned setting}}
\begin{tabular}{l|l|rrrrrr|rrrrrr}
\toprule
\multicolumn{2}{c|}{\textbf{Setting}} & \multicolumn{6}{c|}{\textbf{VitalDB (2S)}} & \multicolumn{6}{c}{\textbf{VitalDB (30S)}} \\
\cmidrule(lr){1-2} \cmidrule(lr){3-8} \cmidrule(lr){9-14}
\textbf{Horizon} & \textbf{Model} & \textbf{F1$\uparrow$} & \textbf{Rec$\uparrow$} & \textbf{Prec$\uparrow$} & \textbf{Acc$\uparrow$} & \textbf{MAE$\downarrow$} & \textbf{MSE$\downarrow$}
   & \textbf{F1$\uparrow$} & \textbf{Rec$\uparrow$} & \textbf{Prec$\uparrow$} & \textbf{Acc$\uparrow$} & \textbf{MAE$\downarrow$} & \textbf{MSE$\downarrow$} \\
\midrule
\multirow{4}{*}{5min} & Test\cite{units} & 68.20 & \underline{77.20} & 61.00 & \textbf{92.00} & \underline{4.75} & 52.65 & 67.50 & \underline{67.20} & 67.80 & 91.20 & 6.25 & 79.89 \\
 & TTT\cite{ttt} & 68.20 & 77.00 & 61.21 & 92.00 & 4.77 & \underline{52.43} & \underline{68.40} & 65.50 & 71.57 & \underline{91.50} & 6.21 & 79.25 \\
 & TTT++\cite{ttt++} & \underline{68.30} & \underline{77.20} & \underline{61.24} & 92.00 & 4.75 & 52.45 & 67.40 & 66.91 & \underline{67.90} & 91.30 & \underline{6.17} & \underline{79.18} \\
 & \cellcolor{lightgray}CSA-TTA(Ours) & \cellcolor{lightgray}\textbf{69.10} & \cellcolor{lightgray}\textbf{78.84} & \cellcolor{lightgray}\textbf{61.50} & \cellcolor{lightgray}92.20 & \cellcolor{lightgray}\textbf{4.71} & \cellcolor{lightgray}\textbf{51.47} & \cellcolor{lightgray}\textbf{69.40} & \cellcolor{lightgray}\textbf{67.50} & \cellcolor{lightgray}\textbf{71.41} & \cellcolor{lightgray}\textbf{91.90} & \cellcolor{lightgray}\textbf{6.10} & \cellcolor{lightgray}\textbf{77.28} \\
\midrule
\multirow{4}{*}{10min} & Test\cite{units} & 64.10 & \underline{61.90} & 66.60 & 89.20 & \underline{6.04} & 76.04 & \underline{63.50} & \underline{57.30} & 71.20 & 87.90 & 6.13 & 77.57 \\
 & TTT\cite{ttt} & \underline{64.50} & 61.50 & \underline{67.81} & \underline{89.40} & 6.06 & 75.91 & 63.00 & 53.40 & \textbf{76.70} & \underline{88.00} & 6.15 & 77.53 \\
 & TTT++\cite{ttt++} & 64.20 & 61.70 & 66.91 & 89.20 & 6.05 & \underline{75.88} & 63.10 & 56.28 & 71.80 & \underline{88.00} & \underline{6.12} & \underline{77.45} \\
 & \cellcolor{lightgray}CSA-TTA(Ours) & \cellcolor{lightgray}\textbf{65.80} & \cellcolor{lightgray}\textbf{63.46} & \cellcolor{lightgray}\textbf{68.32} & \cellcolor{lightgray}\textbf{89.40} & \cellcolor{lightgray}\textbf{5.91} & \cellcolor{lightgray}\textbf{74.61} & \cellcolor{lightgray}\textbf{64.70} & \cellcolor{lightgray}\textbf{58.50} & \cellcolor{lightgray}\underline{72.40} & \cellcolor{lightgray}\textbf{88.20} & \cellcolor{lightgray}\textbf{6.00} & \cellcolor{lightgray}\textbf{75.56} \\
\midrule
\multirow{4}{*}{15min} & Test\cite{units} & 61.50 & \underline{52.70} & 73.80 & 87.30 & \underline{6.61} & 86.27 & \underline{60.50} & 52.70 & 71.01 & \underline{84.80} & 6.72 & 87.86 \\
 & TTT\cite{ttt} & 61.22 & 52.50 & 73.41 & 87.30 & 6.63 & 86.41 & 59.00 & 46.80 & \textbf{80.00} & 84.70 & 6.62 & \underline{86.82} \\
 & TTT++\cite{ttt++} & \underline{61.63} & \underline{52.80} & \underline{74.01} & \underline{87.40} & 6.62 & \underline{86.01} & 59.00 & 46.80 & \textbf{80.00} & 84.70 & \underline{6.60} & 86.98 \\
 & \cellcolor{lightgray}CSA-TTA(Ours) & \cellcolor{lightgray}\textbf{62.30} & \cellcolor{lightgray}\textbf{53.50} & \cellcolor{lightgray}\textbf{74.56} & \cellcolor{lightgray}\textbf{87.50} & \cellcolor{lightgray}\textbf{6.55} & \cellcolor{lightgray}\textbf{84.12} & \cellcolor{lightgray}\textbf{62.60} & \cellcolor{lightgray}\textbf{53.80} & \cellcolor{lightgray}\underline{74.84} & \cellcolor{lightgray}\textbf{85.30} & \cellcolor{lightgray}\textbf{6.60} & \cellcolor{lightgray}\textbf{85.10} \\
\bottomrule
\end{tabular}
\end{subtable}
\caption{\textbf{Full results of the Units backbone on VitalDB (2S/30S) under different adaptation settings: (a) zero-shot setting; (b) fine-tuning setting.} Results are shown for three prediction horizons (5, 10, and 15 minutes) with 15-minute lookback windows. The best performance is highlighted in \textbf{bold}, and the second-best is \underline{underlined}. Metrics with the $\uparrow$ symbol (e.g., F1, Recall) indicate higher is better, while those with $\downarrow$ (e.g., MAE, MSE) indicate lower is better.}
\label{tab:units_full_results}
\end{table*}

\begin{table*}[ht]
\centering
\begin{tabular}{l|l|rrrrrr|rrrrrr}
\toprule
\multicolumn{2}{c|}{\textbf{Setting}} & \multicolumn{6}{c|}{\textbf{TimesFM\cite{timesfm}}} & \multicolumn{6}{c}{\textbf{Units\cite{units}}} \\
\cmidrule(lr){1-2} \cmidrule(lr){3-8} \cmidrule(lr){9-14}
\textbf{Horizon} & \textbf{Model} & \textbf{F1$\uparrow$} & \textbf{Rec$\uparrow$} & \textbf{Prec$\uparrow$} & \textbf{Acc$\uparrow$} & \textbf{MAE$\downarrow$} & \textbf{MSE$\downarrow$} 
   & \textbf{F1$\uparrow$} & \textbf{Rec$\uparrow$} & \textbf{Prec$\uparrow$} & \textbf{Acc$\uparrow$} & \textbf{MAE$\downarrow$} & \textbf{MSE$\downarrow$} \\
\midrule
\multirow{2}{*}{5min} & Test\cite{timesfm} & 72.30 & 63.50 & 83.93 & 89.90 & 5.14 & 67.82 & 57.30 & 49.30 & 68.40 & 86.60 & 6.13 & 85.23 \\
 & \cellcolor{lightgray}CSA-TTA(Ours) & \cellcolor{lightgray}\textbf{74.60} & \cellcolor{lightgray}\textbf{66.71} & \cellcolor{lightgray}\textbf{84.60} & \cellcolor{lightgray}\textbf{90.60} & \cellcolor{lightgray}\textbf{4.96} & \cellcolor{lightgray}\textbf{66.02} & \cellcolor{lightgray}\textbf{65.80} & \cellcolor{lightgray}\textbf{59.80} & \cellcolor{lightgray}\textbf{73.00} & \cellcolor{lightgray}\textbf{87.20} & \cellcolor{lightgray}\textbf{6.01} & \cellcolor{lightgray}\textbf{82.66} \\
\midrule
\multirow{2}{*}{10min} & Test\cite{timesfm} & 68.50 & 57.83 & 84.00 & 86.00 & 6.15 & 92.14 & 55.30 & 40.50 & \textbf{87.20} & 82.70 & 6.49 & 92.77 \\
 & \cellcolor{lightgray}CSA-TTA(Ours) & \cellcolor{lightgray}\textbf{69.50} & \cellcolor{lightgray}\textbf{59.02} & \cellcolor{lightgray}\textbf{84.50} & \cellcolor{lightgray}\textbf{86.50} & \cellcolor{lightgray}\textbf{5.97} & \cellcolor{lightgray}\textbf{87.84} & \cellcolor{lightgray}\textbf{62.50} & \cellcolor{lightgray}\textbf{50.10} & \cellcolor{lightgray}82.90 & \cellcolor{lightgray}\textbf{84.10} & \cellcolor{lightgray}\textbf{6.33} & \cellcolor{lightgray}\textbf{88.73} \\
\midrule
\multirow{2}{*}{15min} & Test\cite{timesfm} & 69.90 & 60.97 & 81.90 & 84.00 & 6.72 & 104.10 & 55.70 & 41.50 & 84.67 & 79.50 & 6.72 & 97.92 \\
 & \cellcolor{lightgray}CSA-TTA(Ours) & \cellcolor{lightgray}\textbf{70.70} & \cellcolor{lightgray}\textbf{61.58} & \cellcolor{lightgray}\textbf{83.00} & \cellcolor{lightgray}\textbf{84.50} & \cellcolor{lightgray}\textbf{6.31} & \cellcolor{lightgray}\textbf{96.24} & \cellcolor{lightgray}\textbf{63.10} & \cellcolor{lightgray}\textbf{50.10} & \cellcolor{lightgray}\textbf{85.40} & \cellcolor{lightgray}\textbf{82.10} & \cellcolor{lightgray}\textbf{6.55} & \cellcolor{lightgray}\textbf{94.68} \\
\bottomrule
\end{tabular}
\caption{\textbf{Full results of the TimesFM and Units backbone on in-Hospital test set in zero-shot setting.} Results are shown for three prediction horizons (5, 10, and 15 minutes) with 15-minute lookback windows. The best performance is highlighted in \textbf{bold}, and the second-best is \underline{underlined}. Metrics with the $\uparrow$ symbol (e.g., F1, Recall) indicate higher is better, while those with $\downarrow$ (e.g., MAE, MSE) indicate lower is better.}
\label{tab:inhospital_results}
\end{table*}

\end{document}